\PassOptionsToPackage{table,svgnames}{xcolor}
\documentclass[pmlr,twocolumn,10pt,noamsthm]{jmlr} 
\hypersetup{
    citecolor=NavyBlue,
    linkcolor=NavyBlue,
    urlcolor=NavyBlue
}
\usepackage{booktabs}
\usepackage{siunitx}

\usepackage[switch]{lineno}

\usepackage{graphicx}
\usepackage{wrapfig}
\usepackage{subcaption}
\allowdisplaybreaks

\makeatletter

\let\c@theorem\relax

\let\c@proposition\relax

\let\c@lemma\relax

\let\c@corollary\relax

\let\c@definition\relax

\let\c@remark\relax

\makeatother

\usepackage{amsthm}
\usepackage{thmtools}
\usepackage{thm-restate}

\newtheorem{theorem}{Theorem}
\newtheorem*{theorem*}{Theorem}
\newtheorem{proposition}[theorem]{Proposition}
\newtheorem*{proposition*}{Proposition}
\newtheorem{corollary}[theorem]{Corollary}
\newtheorem{lemma}[theorem]{Lemma}
\theoremstyle{definition}
\newtheorem{definition}{Definition}
\newtheoremstyle{exampstyle}
  {3pt} 
  {0pt} 
  {} 
  {} 
  {\bfseries} 
  {.} 
  {.5em} 
  {} 
\theoremstyle{exampstyle}
\newtheorem{assumption}{Assumption}
\theoremstyle{remark}
\newtheorem*{remark}{Remark}
\newcommand{\proofcomment}[1]{\text{\color{gray}#1}}

\usepackage[nameinlink]{cleveref}
\Crefname{assumption}{Assumption}{Assumptions}
\Crefname{equation}{Eqn}{Eqn}

\newcommand{\IS}{\textup{IS}}
\newcommand{\DM}{\textup{DM}}
\newcommand{\DR}{\textup{DR}}
\newcommand{\CIS}{\textup{IS\textsuperscript{+}}}
\newcommand{\CDM}{\textup{DM\textsuperscript{+}}}
\newcommand{\DMIS}{\textup{DR}}
\newcommand{\CDMIS}{\textup{DM\textsuperscript{+}-IS}}
\newcommand{\DMCIS}{\textup{DM-IS\textsuperscript{+}}}
\newcommand{\CDMCIS}{\textup{DM\textsuperscript{+}-IS\textsuperscript{+}}}
\usepackage{cleveref}
\usepackage{bbm}
\usepackage{makecell}

 \jmlrvolume{333}
\jmlryear{2026}
\jmlrsubmitted{LEAVE UNSET}
\jmlrpublished{LEAVE UNSET}
\jmlrworkshop{Conference on Health, Inference, and Learning (CHIL) 2026} 

\title[CANDOR]{CANDOR: Counterfactual ANnotated DOubly Robust Off-Policy Evaluation}

\author{%
\Name{Aishwarya Mandyam}\ \Email{am2@stanford.edu}\\
\addr Stanford University
\AND
\Name{Shengpu Tang} \Email{shengpu.tang@emory.edu}\\
\addr Emory University
\AND
\Name{Jiayu Yao} \Email{jy3491@columbia.edu}\\
\addr Columbia University
\AND
\Name{Jenna Wiens} \Email{wiensj@umich.edu}\\
\addr University of Michigan
\AND
\Name{Barbara E. Engelhardt} \Email{barbarae@stanford.edu}\\
\addr Stanford University, The Gladstone Institutes
}

\begin{document}

\maketitle
\begin{abstract}
Off-policy evaluation (OPE) is critical for applying contextual bandit algorithms to high-stakes decision-making settings such as healthcare, where new treatment policies must be evaluated prior to deployment. Unfortunately, OPE techniques are inherently limited by the breadth of the available data, which may not be sufficient to evaluate the performance of a new policy. Recent work attempts to improve dataset coverage by adding expert-annotated counterfactual samples. However, such annotations are often imperfect and can lead to worse estimator performance than using no annotations at all. To better leverage imperfect annotations, we propose a family of OPE estimators grounded in the doubly robust (DR) framework, which combines importance sampling (IS) with a reward model (direct method, DM) for better statistical guarantees. We study three ways of incorporating counterfactual annotations. Under mild assumptions, we prove that using annotations within just the DM component yields the most desirable theoretical results. Experiments on multiple healthcare tasks, including real-world electronic health records (EHR) data, show that this strategy is most robust under misspecified reward models and inaccurate annotations. By addressing the challenges posed by imperfect annotations, this work broadens the applicability of OPE methods and facilitates safer deployment of decision-making policies in healthcare.
\end{abstract}
\paragraph*{Data and Code Availability}
We consider three synthetic or semi-synthetic domains (Multi-Armed Bandit~\citep{tang2023counterfactualaugmented},  HeartSteps~\citep{mandyam2024adaptive}, Sepsis~\citep{oberst2019counterfactualoffpolicyevaluationgumbelmax}) and data from the MIMIC-IV dataset~\citep{mimicivdataset}. All data sources are publicly available. We include the code for our experiments on \href{https://github.com/bee-hive/candor.git}{Github}.

\paragraph*{Institutional Review Board (IRB)}
This research does not require IRB approval. 

\section{Introduction}
\label{sec:intro}
Contextual bandit methods have been widely applied to learn optimal sequential decision-making policies across several domains. In healthcare, they have been used for high-stakes tasks that directly affect patient outcomes, such as personalized treatment design and adaptive clinical decision support~\citep{yao2021power}. Given the nature of these applications, it is critical for practitioners to assess the performance of a new policy prior to deployment. 

A common approach to assess policy performance is off-policy evaluation (OPE) (\citet{Sutton_Barto_2018}, Chapter 5), which estimates the value of a new policy (the target policy) using a behavior dataset retrospectively collected from a different policy. For example, a hospital may wish to evaluate a new potassium repletion policy that adjusts intravenous potassium dosing, using only retrospective data from patients treated under the hospital's existing treatment protocol. OPE methods estimate how the new policy would perform (e.g., its treatment effect on patient outcomes) without ever deploying it on real patients. This makes these methods crucial for safe policy deployment. 

However, OPE is inherently limited by the quality and coverage of the behavior dataset. If the behavior dataset is collected from a hospital that has never prescribed a recently developed drug, then no OPE method can reliably evaluate a policy that recommends the new drug since there are no observed outcomes for that treatment.

To address this issue, \citet{tang2023counterfactualaugmented} proposed an importance sampling (IS)-based OPE estimator called C-IS (referred to in this work as $\CIS$). $\CIS$ improves dataset coverage by augmenting the behavior dataset with expert-sourced annotations (i.e., predicted rewards) of counterfactual actions. However, $\CIS$ relies on the strong assumption that counterfactual annotations are \textbf{free of errors}. In practice, even expert-generated annotations are prone to errors, and their accuracy is typically unknown a priori. For example, clinicians may systematically over- or underestimate the effects of an uncommon treatment, or disagree due to limited clinical experience. Determining the optimal way to incorporate annotations of unknown quality into an OPE estimator remains an open challenge. 

In light of this shortcoming, we propose a family of OPE estimators based on the standard doubly robust (DR) estimator~\citep{Dud_k_2014}. Compared to IS estimators, DR estimators offer provable reductions in variance while remaining unbiased. However, incorporating potentially imperfect counterfactual annotations into DR estimators without sacrificing these desirable theoretical properties is nontrivial.

In this work, we study three ways of modifying DR estimators to include counterfactual annotations, each of which impacts the estimator performance in a different way. 
We perform a theoretical analysis of the bias-variance trade-off and an empirical robustness analysis of the estimator performance. Under \textbf{imperfect annotations}, we identify one estimator that effectively leverages counterfactual annotations to improve coverage without compounding error from the annotations. In contrast, the other two estimators accumulate error in proportion to the magnitude of error in the annotations, resulting in policy value estimates that are worse than simply ignoring the annotations altogether. In summary, our contributions are the following:
{
\setlength{\leftmargini}{1.5em}
\begin{itemize}
\setlength{\itemsep}{0pt}
    \item \textbf{We propose a family of DR-inspired OPE estimators} that can leverage counterfactual annotations. We theoretically analyze our proposed estimators, showing that the manner in which annotations are incorporated into the estimator substantially affects estimator performance (\Cref{sec:methods}). Among these estimators, one is robust to both reward model misspecification and inaccurate annotations.
    \item \textbf{We evaluate our estimators using one synthetic and two semi-synthetic healthcare environments, as well as a real-world EHR dataset.} The synthetic and semi-synthetic settings are used to empirically validate our theoretical insights, while the EHR dataset demonstrates the practical utility of our method in a high-stakes clinical decision-making setting (\Cref{sec:experiments}). 
    \item \textbf{We conduct robustness analyses} of the proposed OPE estimators under realistic clinical conditions, where both annotation quality and reward model accuracy are unknown. We find that one of the proposed estimators is more robust than the others, making it the most suitable choice for reliable policy evaluation in healthcare settings (\Cref{sec:experiments}).
\end{itemize}
}

\section{Background}
\label{sec:background}
We consider a contextual bandit setting defined by $(\mathcal{S}, \mathcal{A}, R, d_0)$, where $\mathcal{S}$ is the discrete context space, $\mathcal{A}$ is the discrete action space, $R: \mathcal{S}\times\mathcal{A}\rightarrow \Delta(\mathbb{R})$ is the reward function, and $d_0$ is the context distribution. We observe a behavior dataset $D=\{(s_i,a_i,r_i)\}_{i=1}^N$ collected under a behavior policy $\pi_b$, with samples drawn i.i.d. from an underlying data-generating distribution $\mathcal{D}$.
We aim to estimate the value of a target policy $\pi_e$, defined as $v(\pi_e) = \mathbb{E}_{s \sim d_0, a \sim \pi_e(\cdot|s), r \sim R(s, a)}[r]$. 
\subsection{Off-Policy Evaluation}
We briefly review three classes of OPE methods for contextual bandits. Importance sampling (IS)~\citep{IPS_horvitz,eligibility_traces} assigns an inverse propensity score (IPS), $\rho_{s}(a) = \frac{\pi_e(a|s)}{\pi_b(a|s)}$, to each sample $(s_i, a_i, r_i)$ in the behavior dataset and estimates the target policy value as, 
$$\hat{V}^{\IS} = \frac{1}{N} \sum_{i=1}^{N} \rho_{s_i}(a_i) r_i.$$ 
Following prior work, we assume that the IPS $\rho$ is known~\citep{farajtabar2018robust,thomas2016dataefficient}. 
\begin{assumption}[Common support] \label{asm:common-support}
$\pi_e(a|s) > 0 \rightarrow \pi_b(a|s) > 0$.
\end{assumption}
Under a standard support assumption (\Cref{asm:common-support}),
IS yields an unbiased estimate of the target policy value, $v(\pi_e)$ \citep{eligibility_traces}. Under the same assumption, the variance of the estimator is
\begin{align}
      N \cdot \mathbb{V}[\hat{V}^{\IS}] &= \mathbb{V}_{s \sim d_0}[v^{\pi_e}(s)] \nonumber\\
   &+ \mathbb{E}_{s \sim d_0}[\mathbb{V}_{a \sim \pi_b(\cdot|s)}[\rho_{s}(a) \bar{R}(s,a)]] \label{eqn:standard_is_varaince}\\
   &+ \mathbb{E}_{s \sim d_0}[\mathbb{E}_{a \sim \pi_b(\cdot|s)}[\rho_{s}(a)^2 \sigma_R(s,a)^2]]\nonumber,
\end{align}
where $\bar{R}(s,a)=\mathbb{E}[R(s,a)]$ and $\sigma_R(s,a)^2=\mathbb{V}[R(s,a)]$ are the mean and variance of the reward distribution, respectively \citep{tang2023counterfactualaugmented}. 

Another approach to OPE is the direct method ($\DM$)~\citep{Li_2010, beygelzimer_2009,Seijen2009ATA, harutyunyan2016qlambdaoffpolicycorrections,le2019batchpolicylearningconstraints,voloshin2021empirical}.
$\DM$ first uses the behavior dataset to fit a reward model, $\hat{R}: \mathcal{S} \times \mathcal{A} \to \Delta(\mathbb{R})$, to predict the conditional mean reward. It then uses $\hat{R}$ to directly compute the target policy value as 
$$\hat{V}^{\DM} = \sum_{s} d_0(s) \sum_a \pi_e(a|s) \hat{R}(s, a).$$ 
The reward model $\hat{R}$ may range in complexity from simple regression models to neural networks. When the reward model is realizable and the behavior dataset provides full coverage, the $\DM$ estimator has zero bias and favorable variance. 
In practice, $\DM$ estimators often have lower variance than $\IS$ \citep{dudik2011doubly} when the reward model is accurate. 

The last category of OPE approaches consists of doubly robust (DR) methods~\citep{dudik2011doubly, Dud_k_2014, farajtabar2018robust, jiang2016doubly}. These methods are termed ``doubly robust'' because they maintain strong theoretical guarantees when either the IPS ratio $\rho$, or the estimated reward function $\hat{R}$, is inaccurate, thereby providing robustness to both sources of error. In this work, we aim to design an OPE estimator that is robust to misspecified reward models and inaccurate annotations. 

The standard DR estimator is defined as
\begin{equation}
\label{eqn:standard_dr}
\hat{V}^{\DR} = \frac{1}{N} \sum_{i=1}^N \underbrace{\hat{R}(s_i, \pi_e)}_{\text{DM part}} + \underbrace{\rho_{s_i}(a_i) (r_i - \hat{R}(s_i, a_i))}_{\text{IS part}},
\end{equation}
where $\hat{R}(s, \pi_e) = \sum_{a \in \mathcal{A}} \pi_e(a|s) \hat{R}(s, a)$ is the estimated value of state $s$ under the target policy $\pi_e$ using the reward model $\hat{R}$.
We refer to the first and second term in \Cref{eqn:standard_dr} as the \emph{DM part} and the \emph{IS part}, respectively.

Under the common support assumption (\Cref{asm:common-support}), the DR estimator produces an unbiased estimate of $v(\pi_e)$ (if the reward model and OPE estimate are learned using independent splits of the data). DR methods often exhibit lower variance than IS-based methods; the variance can be written as 
\begin{align*}
    N \cdot &\mathbb{V}[\hat{V}^{\DR}] = \mathbb{V}_{s \sim d_0}[v^{\pi_e}(s)] \\
    &+ \mathbb{E}_{s \sim d_0}\left[ \mathbb{V}_{a \sim \pi_b(\cdot|s)}\left[\rho_{s}(a)(\bar{R}(s, a) - \hat{R}(s, a)) \right]\right] \\
    &+ \mathbb{E}_{s \sim d_0}\left[\mathbb{E}_{a \sim \pi_b(\cdot|s)}\left[\rho_{s}(a)^2  \sigma_R(s, a)^2 \right] \right].
\end{align*}
Comparing to \Cref{eqn:standard_is_varaince}, the variance reduction relative to the IS estimator rests in the second term: the IPS $\rho$ is scaled by the residual $\bar{R}(s,a) - \hat{R}(s,a)$, while the standard $\IS$ estimator scales $\rho$ using $\bar{R}(s,a)$ alone. When the reward model $\hat{R}$ is well-specified, this residual approaches $0$, shrinking the impact of the IPS term and reducing overall variance.
\begin{figure}[htbp]
\floatconts
  {fig:dataset}
  {\caption{\textbf{A counterfactual-annotated dataset with two contexts and two actions.} Two factual samples, $(s_1, a_1, r_1)$ (left) and $(s_2, a_1, r_2)$ (right), are observed. The left sample has one  counterfactual annotation, $(s_1, a_2, g_1^{a_2})$,  while the right one has none.}}
  {\includegraphics[width=0.8\linewidth]{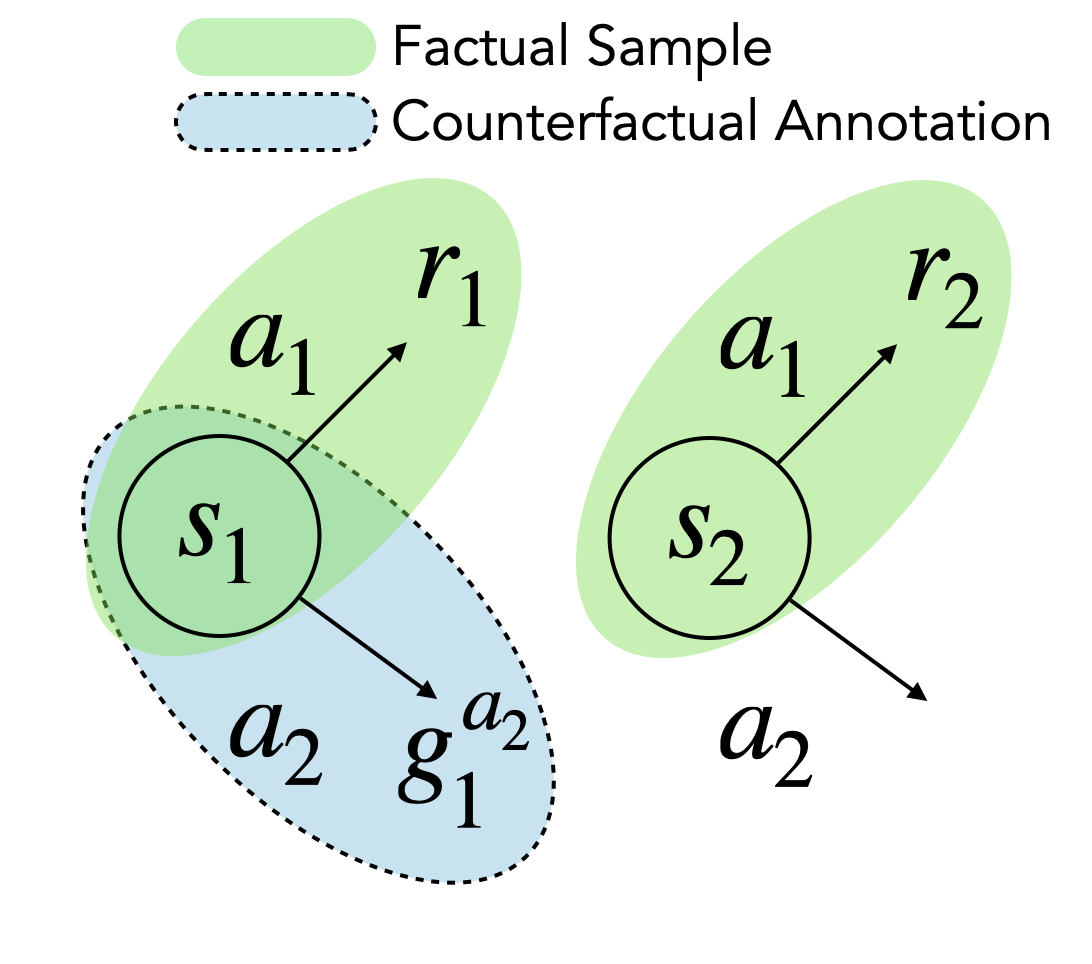}}
\end{figure}
\subsection{Counterfactual Annotations}
In our work, we consider incorporating counterfactual annotations to increase data coverage. Specifically, we assume that each factual sample  $(s_i,a_i)$ in the behavior dataset $D$ is associated with a set of counterfactual annotations $\mathbf{g}_i=\{g_i^{\tilde{a}} \mid \tilde{a}\in \mathcal{A} \setminus \{a_i\}\}$. Note that $\mathbf{g}_i$ may be empty. We assume that the annotation of the counterfactual action $\tilde{a}$ is drawn from a distribution $G: \mathcal{S} \times \mathcal{A} \to \Delta(\mathbb{R})$,  $g_i^{\tilde{a}}\sim G(s_i, \tilde{a})$.

Because counterfactual annotations are often expensive to obtain, we do not assume full annotation coverage. We refer to the dataset that augments factual samples and counterfactual annotations as the \textit{counterfactual-annotated} dataset, denoted by $D^+ \sim \mathcal{D^+}$, where $\mathcal{D^+}$ is the distribution over such datasets. 
\Cref{fig:dataset} illustrates a simple example of a counterfactual-annotated dataset with partial coverage, where two factual samples are observed and only one has an associated counterfactual annotation.

In \Cref{sec:theory}, we study three scenarios for the annotation distribution $G$ (perfect, biased, or noisy annotations). Additionally, for ease of notation, we use $c_i^a$ to refer to either the reward or the counterfactual annotation of the factual sample $(s_i, a_i)$, i.e., $c_i^a = r_i$ when $a = a_i$ and $c_i^a = g_i^a$ when $a \neq a_i$. 

\subsection{The \texorpdfstring{\CIS}{IS⁺} Estimator}
Naively incorporating counterfactual annotations into $\IS$ estimators introduces bias, as the context distribution in $D^+$ may differ from that of the behavior dataset $D$. To address this, \citet{tang2023counterfactualaugmented} introduced $\CIS$, which reweights factual and counterfactual samples to align these two distributions. 

Specifically,  let $\{w_i^a\}$ be a set of user-defined weights for the $i$-th factual sample $(s_i,a_i)$ and its associated counterfactual annotations $\{(s_i, g_i^a)\}$, with the constraint $\sum_{a \in \mathcal{A}} w_i^a = 1$ for each sample $i$ (i.e., the weights for all actions associated with a factual sample $s_i$ must add up to 1). We set $w_i^{\tilde{a}}=0$ if the annotation of the counterfactual action $\tilde{a}$ for the $i$-th sample's state is unavailable. Define the augmented IPS ratio as $\rho_{s}^+(a) = \frac{\pi_e(a|s)}{\pi_b^+(a|s)}$ and the augmented behavior policy as, 
\begin{align*}
    \pi_b^+(a|s) &= \bar{W}(a|s,a) \pi_b(a|s) \\
     &+ \sum_{\check{a} \in \mathcal{A} \setminus \{a\}} \bar{W}(a | s, \check{a}) \pi_b(\check{a} | s),
\end{align*}
where $\bar{W}(\tilde{a}|s,a)=\mathbb{E}[w^{\tilde{a}}]$ is the average weight of action $\tilde{a}$ for the factual context-action pair $(s,a)$.
The $\CIS$ estimator is defined as,
\[\hat{V}^{\CIS} =\frac{1}{N} \sum_{i=1}^N\sum_{a \in \mathcal{A}} w_i^a \rho_{s_i}^+(a) c_i^a. \]

\section{Methods}
\label{sec:methods}
When the behavior dataset has limited coverage, IS estimators are known to have high variance~\citep{jiang2016doubly}. In contrast, DM estimators have high bias when the reward model is misspecified. 
We therefore explore how to introduce counterfactual annotations into a DR estimator, which retains beneficial theoretical properties even when the reward model is misspecified, a common challenge in  healthcare settings. While DR estimators are well-understood in a setting with only factual samples, we seek to incorporate counterfactual annotations such that we can improve behavior dataset coverage. 

A naive approach is to apply a standard DR estimator directly to the counterfactual-annotated dataset $D^+$, treating the counterfactual annotations as additional samples. However, as we discuss in \Cref{apd:naive_dr}, this approach can produce arbitrarily biased estimates of $v(\pi_e)$ because it alters the context distribution of the behavior dataset, an issue that arises regardless of annotation quality. As such, we explore three new estimators that incorporate counterfactual annotations into the DR framework. 

\subsection{Proposed DR Estimators with Counterfactual Annotations}

The standard DR estimator (\Cref{eqn:standard_dr}) can be broken down into two components: the direct method (DM) part and the importance sampling (IS) part. We observe that counterfactual annotations can be incorporated independently into either 
component, or into both. Based on this insight, we propose three new DR-inspired estimators that can leverage counterfactual annotations (\Cref{eqn:cdmis,eqn:dmcis,eqn:cdmcis}). 
\begin{align}
\label{eqn:cdmis}
&\scalebox{0.85}{
$\displaystyle \hat{V}^{\CDMIS} = \frac{1}{N} \sum_{i=1}^N \Big(\textcolor{red}{\hat{R}^+}(s_i, \pi_e) + \rho_{s_i}(a_i) (r_i - \textcolor{red}{\hat{R}^+}(s_i,a_i))\Big)$} \\
\label{eqn:dmcis}
&\scalebox{0.85}{
$\displaystyle \hat{V}^{\DMCIS} = \frac{1}{N} \sum_{i=1}^N \Big(\hat{R}(s_i,\pi_e) + \textcolor{blue}{\sum_{a \in A} w_{i}^a \rho_{s_i}^+(a)(c_i^{a}} - \hat{R}(s_i,a))\Big)$} \\
\label{eqn:cdmcis}
&\scalebox{0.85}{
$\displaystyle \hat{V}^{\CDMCIS} = \frac{1}{N} \sum_{i=1}^N \Big(\textcolor{red}{\hat{R}^+}(s_i,\pi_e) + \textcolor{blue}{\sum_{a \in A} w_{i}^a \rho_{s_i}^+(a) (c_i^{a}} - \textcolor{red}{\hat{R}^+}(s_i,a))\Big)$}. \raisetag{5pt}
\end{align}

Here, $\hat{R}^+(s,a)$ is the reward function estimate learned using the counterfactual-annotated dataset $D^+$ (further discussion in \Cref{apd:reward_fn}), and $\hat{R}^+(s, \pi_e) = \sum_{a \in \mathcal{A}} \pi_e(a|s) \hat{R}^+(s, a)$. 
The first estimator, \textbf{$\CDMIS$} (\Cref{eqn:cdmis}), uses the counterfactual-annotated dataset to estimate the reward model and combines it with standard IS. The second estimator, \textbf{$\DMCIS$} (\Cref{eqn:dmcis}), instead uses counterfactual annotations to augment the IS part (as in $\CIS$) and combines it with a standard DM estimator. The third estimator, \textbf{$\CDMCIS$} (\Cref{eqn:cdmcis}) uses counterfactual annotations in both the DM and IS parts. 

\subsection{Theoretical Analyses under Imperfect Annotations}
\label{sec:theory}
In our problem setting, there are three possible sources of error: incorrect estimates of the behavior policy $\pi_b$, a misspecified reward model, and imperfect (biased or noisy) annotations. Much like prior work~\citep{farajtabar2018robust,thomas2016dataefficient}, we assume that the IPS ratio $\rho$ is known, and instead focus on identifying an OPE estimator that is robust to the last two sources of error. Accordingly, we analyze our proposed estimators under reward model misspecification and imperfect annotations, characterizing how these errors affect their performance. 
This analysis also provides insight into the robustness of the estimators, which we study empirically in \Cref{sec:experiments}. 

The novelty of our theoretical results is two-fold. 
First, prior work on DR estimators typically treats the reward model error as fixed. For example, \citet{dudik2011doubly} assumed that $\hat{R}(s,a) = \bar{R}(s,a) + \epsilon(s,a)$ without modeling how $\epsilon(s,a)$ depends on the data used to estimate $\hat{R}$. In contrast, we treat $\hat{R}$ as a random function whose distribution is induced by the dataset used to fit the reward model. Specifically, the reward model, $\hat{R}$, is estimated from a separate dataset, which we refer to as $D_{\hat{R}} \sim \mathcal{D}$ or $D_{\hat{R}^+} \sim \mathcal{D^+}$, depending on if counterfactual annotations are included. Importantly, we do not assume that the reward model class is well specified.

Second, we derive the bias and variance of our proposed DR estimators under imperfect annotations, a setting that more closely reflects practical applications. We formalize the quality of counterfactual annotations through the following three assumptions.
\begin{assumption}[Perfect annotations] 
\label{asm:perfect-annot} \, \\
$\mathbb{E}_{g^a \sim G(s,a)}[g^a]\! =\! \bar{R}(s,a)$, $\mathbb{V}_{g^a \sim G(s,a)}[g^a]\! =\! \sigma_R^2(s,a)$. 
\end{assumption}
\begin{assumption}[Biased annotations]
\label{asm:biased_annot} \, \\
$\mathbb{E}_{g^a \sim G(s,a)}[g^a] = \bar{R}(s,a) + \epsilon_G(s,a), \; \epsilon_G(s,a) \neq 0$. 
\end{assumption}
\begin{assumption}[Noisy annotations]
\label{asm:variance_annot} \, \\
$\mathbb{V}_{g^a \sim G(s,a)}[g^a] = \sigma_R(s,a)^2 + \Delta_G(s,a), \; \Delta_G(s,a) > 0$. 
\end{assumption}
\vspace{0.6em}
In \Cref{asm:biased_annot,asm:variance_annot}, annotation bias and variance, quantified by the terms $\epsilon_G(s,a)$ and $\Delta_G(s,a)$ respectively, are used to study the effect of biased and noisy (i.e., higher variance) annotations. \Cref{asm:perfect-annot} represents the idealized setting of perfect annotations and is mutually exclusive with \Cref{asm:biased_annot,asm:variance_annot}.
Finally, following \citet{tang2023counterfactualaugmented}, we adopt a relaxed data coverage assumption (\Cref{asm:common-support}) to analyze the properties of $\DMCIS$ and $\CDMCIS$. 

\begin{assumption}[{Common support with counterfactual annotations}] 
\label{asm:common-support-cf}
$\pi_e(a|s) > 0 \rightarrow \pi_{b}^{+}(a|s) > 0$.
\end{assumption}

With these assumptions in place, we now summarize the main theoretical implications for the bias and variance of the proposed estimators under both idealized and realistic annotation regimes. First, we consider the setting with perfect annotations (\Cref{asm:perfect-annot}). Under appropriate coverage assumptions (\Cref{asm:common-support} or \labelcref{asm:common-support-cf}), all three proposed estimators are unbiased (\Cref{thm:CDM-IS_unbiasedness,thm:C-DM+C-IS_unbiasedness,thm:DM+C-IS_unbiasedness} in \Cref{apd:bias_variance_dr}). Additionally, when all counterfactual actions are annotated and uniform weights are used (i.e., $w^a=1/|\mathcal{A}|$), both $\DMCIS$ and $\CDMCIS$ are equivalent to the $\CIS$ estimator (\Cref{corr:equiv_equal_weights,corr:equiv_variance_equal_weights} in \Cref{apd:equivalence}). These results suggest that in the absence of annotation error, counterfactual annotations can be viewed as an additional high-quality dataset and yield unbiased OPE estimates.

We next analyze the more realistic setting in which the annotations are imperfect (i.e., \Cref{asm:perfect-annot} is violated). We begin by characterizing the bias of our proposed estimators. Note that the bias derivations only rely on \Cref{asm:biased_annot}, which captures annotation bias and do not require \Cref{asm:variance_annot}, which governs annotation noise.

\begin{proposition}[Unbiasedness of $\CDMIS$ under imperfect annotations]
\label{thm:CDM-IS-biased}
Under common support (\Cref{asm:common-support}) and biased annotations (\Cref{asm:biased_annot}):  $\mathbb{E}[\hat{V}^{\CDMIS}] = v(\pi_e)$.
\end{proposition}

\begin{theorem}[Bias of $\DMCIS$ and $\CDMCIS$ under imperfect annotations]
\label{thm:DR_aug_biased_annot}
Under common support (\Cref{asm:common-support-cf}) and biased annotations (\Cref{asm:biased_annot}), $\DMCIS$ and $\CDMCIS$ have the same expectation: 
\begin{align}
\label{eqn:thm2}
    &\mathbb{E}[\hat{V}^{\DMCIS}] = \mathbb{E}[\hat{V}^{\CDMCIS}] \\
    &= v(\pi_e) + \mathbb{E}_{\substack{s \sim d_0 \\ a \sim \pi_e(s)}}\left[\left(1- \frac{\bar{W}(a|s,a) \pi_b(a|s)}{\pi_b^+(a|s)}\right) \epsilon_G(s, a) \right].
    \nonumber
\end{align}
\end{theorem}

\begin{remark}
\Cref{thm:CDM-IS-biased} establishes that when $\rho$ is known, $\CDMIS$ is an unbiased estimator of the target policy value $v(\pi_e)$, even with biased annotations. In contrast, \Cref{thm:DR_aug_biased_annot} shows that both $\DMCIS$ and $\CDMCIS$ will produce biased estimates of $v(\pi_e)$. Note that the last term in \Cref{eqn:thm2} is identical to the expectation derivation for $\CIS$~\citep{tang2023counterfactualaugmented}.
\end{remark}

Having characterized the bias, we now study the variance of our proposed estimators under imperfect annotations. We begin with $\CDMIS$. 

\begin{theorem}[Variance of $\CDMIS$ under imperfect annotations]
\label{thm:CDM-IS-imperfect_variance}
Under \Cref{asm:biased_annot,asm:variance_annot,,asm:common-support},  
\begin{align*}
    N \cdot &\mathbb{V}[\hat{V}^{\CDMIS}] = \mathbb{V}_{s \sim d_0}[v^{\pi_e}(s)]  \\
    &+ \mathbb{E}_{s \sim d_0}\mathbb{E}_{a \sim \pi_b(s)}[\rho_{s}(a)^2 \sigma_R^2(s,a)] \\
    &+ \textcolor{Purple}{\mathbb{E}_{s \sim d_0} \mathbb{E}_{a \sim \pi_b} \Big[ \big(\rho_{s}(a)^2 - \tfrac{1}{\pi_b(a|s)} \big) \mathbb{V}_{D_{\hat{R}^+}}[\hat{R}^+(s,a)] \Big]} \\
    &+ \textcolor{Green}{{\mathbb{E}_{s \sim d_0} \Big[ \mathbb{E}_{a \sim \pi_b}[\rho_{s}(a)^2 \varepsilon_{\hat{R}^+}(s,a)^2] - \varepsilon_{\hat{R}^+}^{\pi_e}(s)^2 \Big]}},
\end{align*}
\text{where} $\varepsilon_{\hat{R}^+}(s,a) = \mathbb{E}_{D_{\hat{R}^+}}[\hat{R}^+(s,a)] - \bar{R}(s,a)$,  \text{and} $\varepsilon_{\hat{R}^+}^{\pi_e}(s) = \mathbb{E}_{a \sim \pi_e}[\varepsilon_{\hat{R}^+}(s,a)]$. 
\end{theorem}

\begin{remark}
\Cref{thm:CDM-IS-imperfect_variance} characterizes the variance of $\CDMIS$ under imperfect counterfactual annotations, showing that it depends on both the noise and the bias of the counterfactual annotations as reflected in the last two terms. The \textcolor{Purple}{third term}, which depends on $\hat{R}^+$, can dominate the overall variance when annotations are highly noisy. The \textcolor{Green}{last term} emerges from the systematic estimation error of the reward model due to biased annotations.  
\end{remark}

We also derive the variance of $\DMCIS$ and $\CDMCIS$ under imperfect annotations (\Cref{prop:dmcis_imperfect_variance} and \Cref{cdmcis_variance_imperfect_annot}). 
Despite their closed-form analytical expressions, two key questions remain difficult to answer analytically: (1) whether the proposed estimators reduce variance relative to the standard $\DMIS$ estimator, and (2) how the three proposed estimators compare to one another in terms of variance.  Answering these questions depends critically on how the annotation noise ($\Delta_G(s,a)$) and bias ($\epsilon_G(s,a)$) vary across the state-action space; these quantities are highly application-dependent and resistant to general theoretical analysis. We therefore turn to empirical studies in \Cref{sec:experiments}.

We summarize our theorems in Appendix \Cref{tab:theory}, with full proofs provided in \Cref{apd:bias_variance_dr}. In short, under perfect annotations, all of our proposed DR estimators are unbiased. Under imperfect annotations, $\CDMCIS$ and $\DMCIS$ share the same bias, while $\CDMIS$ remains unbiased. We expect imperfect annotations to increase the variance of all three proposed estimators due to the increased bias and variance of the estimated reward-model. Now, we turn to empirical analysis to understand how these estimators perform in practice.

\section{Experiments}
\label{sec:experiments}
Our experiments seek to answer the following questions:
\textbf{1)} How do imperfect annotations alone empirically affect the performance of the proposed methods? \textbf{2)} How do the proposed methods perform under compounding errors from imperfect annotations and misspecified reward models? \textbf{(3)} Which estimator is most robust when the quality of the annotations and reward model is unknown?

\subsection{Experimental Domains}
To answer these questions, we investigate three healthcare-inspired contextual bandit domains with progressively increasing state and action space sizes. The first two domains are semi-synthetic, thus enabling direct control over the data-generating process. This allows us to validate hypotheses derived from our theoretical analyses under controlled conditions. The last domain uses real-world clinical data, where counterfactual annotations are generated using a large language model (LLM), reflecting a more realistic and noisy deployment setting. We additionally verify our proofs in a fully synthetic environment under idealized conditions, with results deferred to \Cref{apd:ws_reward} and \Cref{apd:ms_reward}.

\noindent \textbf{Heartsteps}~\citep{mandyam2024adaptive}: This semi-synthetic mobile health simulator models the user's physical activities given mobile interventions based on the Heartsteps study~\citep{heartsteps}. The context is a three-dimensional vector that includes a treatment effect term and the step count of the previous day. At each decision time, the agent selects between two actions (either \emph{send an intervention} or \emph{do nothing}), and the reward is drawn from a normal distribution with the mean being the square root of the user's observed step count. 

\noindent \textbf{Sepsis}~\citep{oberst2019counterfactualoffpolicyevaluationgumbelmax}: We adapt the semi-synthetic sepsis simulator used in prior work (originally built for the sequential Markov decision process (MDP) setting) to a contextual bandit setting by interacting with the environment for only one step. The patient context is an 8-dimensional vector that contains information about vitals and ongoing treatments. There are 8 treatment options, and the reward is an indicator function of whether the patient is under treatment and has stable vitals. 

\noindent 
\textbf{MIMIC-IV}~\citep{mimicivdataset,physionet}:
MIMIC-IV contains electronic health records for over $65,000$ admitted patients. We study a subset of patients who received intravenous (IV) potassium repletion. Potassium repletion is a common task in critical care settings; imbalanced potassium levels can have severe side effects including cardiac arrest~\citep{prasadpotassiumrepletion}.

We created two splits of the dataset based on whether a patient has renal disease (we refer to these splits as ``renal'' and ``non-renal''). The behavior policy $\pi_b$ and the target policy $\pi_e$ are defined by the clinician treatment policies for non-renal and renal patients, respectively.
In Appendix \Cref{apd:potassium_dosages}, we see that the treatment policies for these subgroups are different. 
Specifically, patients with renal disease are administered lower dosages to account for their impaired kidney function~\citep{Shrimanker2025Potassium}. 
Our goal is to estimate the value of the target policy using data generated under the behavior policy.

In this setting, the patient context is a 20-dimensional vector that contains information about vitals, administered medications, and static covariates. The actions are five discretized potassium dosage levels. The reward is an indicator function of whether the patient's lab potassium value is within the reference range 2 hours after administering a given dosage. We used linear regression to fit our estimated reward model. Distinct from the prior settings, neither $\pi_b$ nor $\pi_e$ is known and both are instead estimated using behavior cloning. While alternatives such as inverse reinforcement learning exist, they are generally more computationally demanding and require additional assumptions, making behavior cloning the more practical choice. 

\subsection{Generating Counterfactual Annotations and Misspecified Reward Models}
For the semi-synthetic domains, to produce perfect counterfactual annotations of state $s$ and counterfactual action $\tilde{a}$, we sample from the true reward model, i.e., $G(s,\tilde{a}) = \mathcal{N}(\bar{R}(s, \tilde{a}), \sigma_R(s, \tilde{a}))$.
To produce biased and noisy counterfactual annotation, we sample from $\mathcal{N}(\bar{R}(s, \tilde{a}) + \epsilon_G(s,a), \sigma_R(s, \tilde{a}) + \Delta_G(s,a))$, where $\epsilon_G(s,a)$ and $\Delta_G(s,a)$ refer to the additional bias and variance that compromise the quality of the annotations. 

For the MIMIC-IV data, we randomly selected a subset of state-action pairs in the behavior dataset and generated counterfactual annotations for those samples using OpenAI's ``o1'' model~\citep{openai2024openaio1card} following prior work~\citep{mandyam2024adaptive}. Specifically, ``o1'' is prompted to predict a patient's blood potassium level after administering a counterfactual dosage of IV potassium. This procedure mimics a setting where counterfactual annotations may be imperfect. Further details regarding the dataset and annotation construction are in \Cref{apd:simulators}.

In addition to imperfect annotations, we study the compounding error of misspecified reward models in the semi-synthetic settings. In our experiments, we create misspecified reward models by either partially observing the state or altering the state representation (\Cref{tab:environments} and \Cref{apd:simulator_details}).

\subsection{Baselines and Metrics}
We compare our proposed estimators against two classes of baselines. The first consists of common OPE estimators that do not use counterfactual annotations ($\IS$, $\DM$, and $\DMIS$). The second includes estimators that leverage counterfactual annotations ($\CIS$ and $\CDM$). $\CDM$ is a direct method estimator that estimates the reward model using the counterfactual-annotated dataset, defined as $$\hat{V}^{\CDM} = \sum_{s} d_0(s) \sum_a \pi_e(a|s) \hat{R}^+(s, a)$$.

For the semi-synthetic domains, we consider various combinations of stochastic behavior and target policies (details in \Cref{apd:simulators}). Specifically, the behavior policies vary in their coverage of the action space. We present results averaged across these combinations. We estimate the ground truth target policy value using Monte Carlo simulation in the synthetic and semi-synthetic domains, and by averaging observed rewards in the MIMIC-IV dataset. For all settings, we report the root mean squared error (RMSE) of estimated policy values. 

\subsection{Results}
\label{sec:results}
\subsubsection{Imperfect annotations with a well-specified reward model}
\label{sec:exp_imperfect_annot_ws_reward}
\begin{figure}[htbp]
\floatconts
  {fig:bias_variance}
  {\caption{\textbf{In Heartsteps, the bias of the counterfactual annotations has a larger impact on RMSE than the variance}. The $x,y$-axis represents the variance ($\Delta_G$) and the bias ($\epsilon_G$) of the annotations respectively. We report mean RMSE, and error metrics \Cref{apd:ws_reward}. The RMSE remains nearly constant as annotation variance increases, and instead increases proportional to the magnitude of the absolute annotation bias. This trend is particularly noticeable in $\DMCIS$ and $\CDMCIS$. The RMSE of $\CDMIS$ is far more consistent regardless of the annotation bias and variance.}}
  {\includegraphics[width=\linewidth]{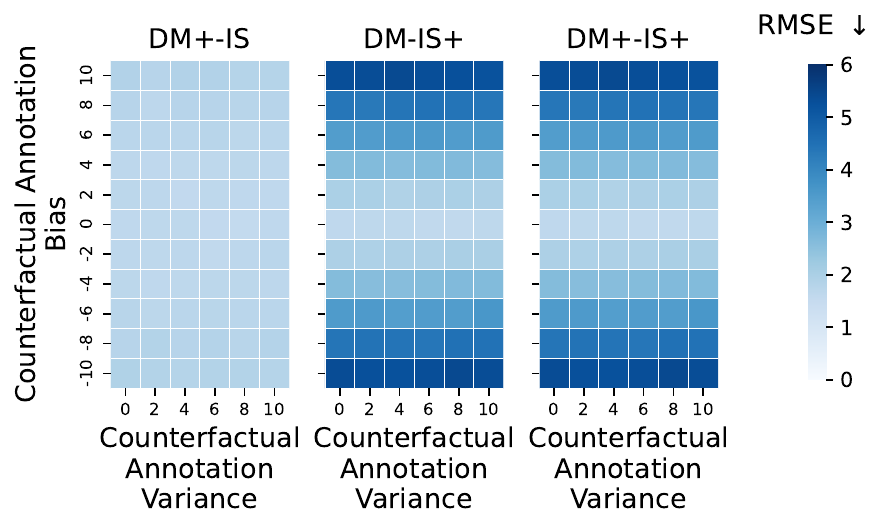}}
\end{figure}
  
First, we isolate the impact of imperfect annotations by studying a setting with well-specified reward models, thereby eliminating confounding effects due to reward model misspecification. In the Heartsteps domain, we observe that \textbf{annotation bias has a greater effect on the RMSE of the proposed estimators than annotation noise (\Cref{fig:bias_variance})}. In particular, the RMSE remains nearly unchanged as the annotation variance increases. From observing the results in \Cref{fig:bias_variance} and \Cref{fig:ws_reward}, we find that this trend holds across all OPE methods that leverage counterfactual annotations and across all synthetic and semi-synthetic domains. 

We also note that the effect of imperfect annotations is especially pronounced in methods that incorporate annotations into the ``IS'' component (e.g., $\CIS$, $\DMCIS$, $\CDMCIS$).
This observation is consistent with our theoretical results (\Cref{{thm:CDM-IS-biased}}, \Cref{thm:DR_aug_biased_annot}), which demonstrate that these estimators are biased when the counterfactual annotations are biased. 

Furthermore, \Cref{fig:bias_variance} and \Cref{fig:ws_reward} also demonstrate that $\CDMIS$ exhibits greater robustness to annotation quality than baseline approaches. The RMSE of the estimator varies little across different magnitudes of annotation bias and variance. Although our theoretical analysis in \Cref{thm:CDM-IS-imperfect_variance} indicates that noisier annotations can, in principle, increase the variance of $\CDMIS$, we empirically observe that this increase is not substantial. 

\subsubsection{Imperfect annotations with a misspecified reward model}
\label{sec:exp_imperfect_annot_mis_reward}
\begin{figure}[htbp]
\floatconts
  {fig:kitchen}
  {\caption{\textbf{$\CDMIS$ is robust to the joint effects of annotation bias and reward model misspecification.} 
We report the mean RMSE across the Heartsteps (\Cref{fig:hs_kitchen}) and Sepsis environments (\Cref{fig:sepsis_kitchen}) as a function of annotation bias ($x$-axis, $\epsilon_G$). 
Across all settings, $\CDMIS$ consistently matches or exceeds the performance of the best baselines (further results in \Cref{apd:empirical_results}). 
Notably, among methods that use counterfactual annotations, $\CDMIS$ exhibits the lowest RMSE across all degrees of annotation bias. 
When compared to OPE baselines without annotations, $\CDMIS$ achieves a lower or comparable RMSE. 
Error metrics are provided in \Cref{apd:ms_reward}.
  }}
  {%
    \subfigure[Heartsteps setting.]{\label{fig:hs_kitchen}%
      \includegraphics[width=\linewidth]{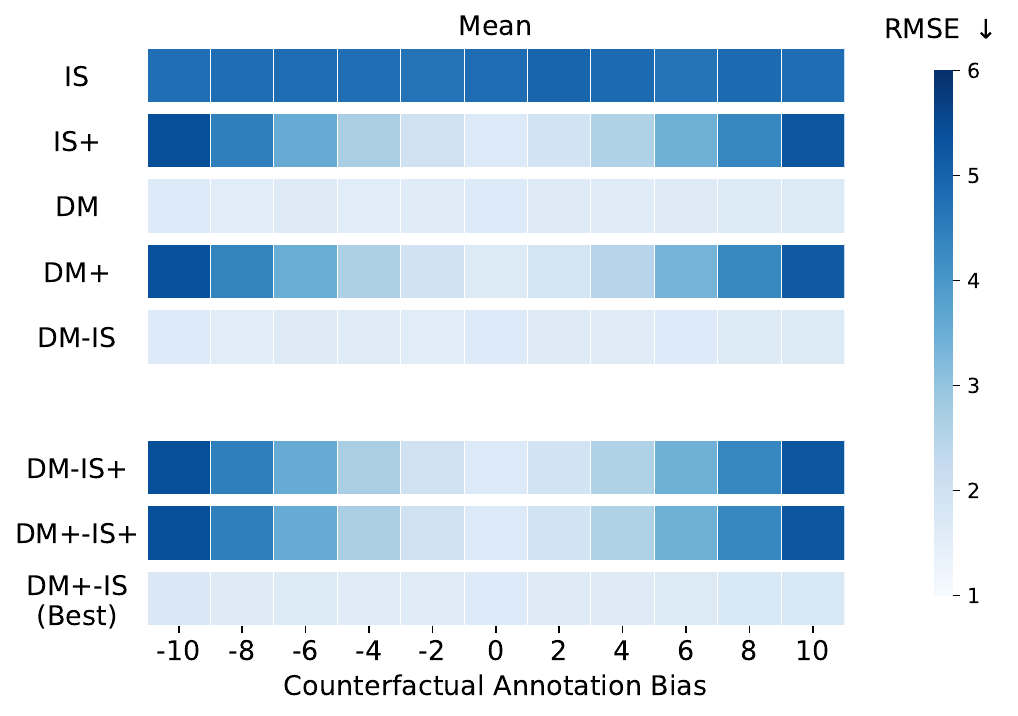}}%
    \qquad
    \subfigure[Sepsis treatment setting.]{\label{fig:sepsis_kitchen}%
      \includegraphics[width=\linewidth]{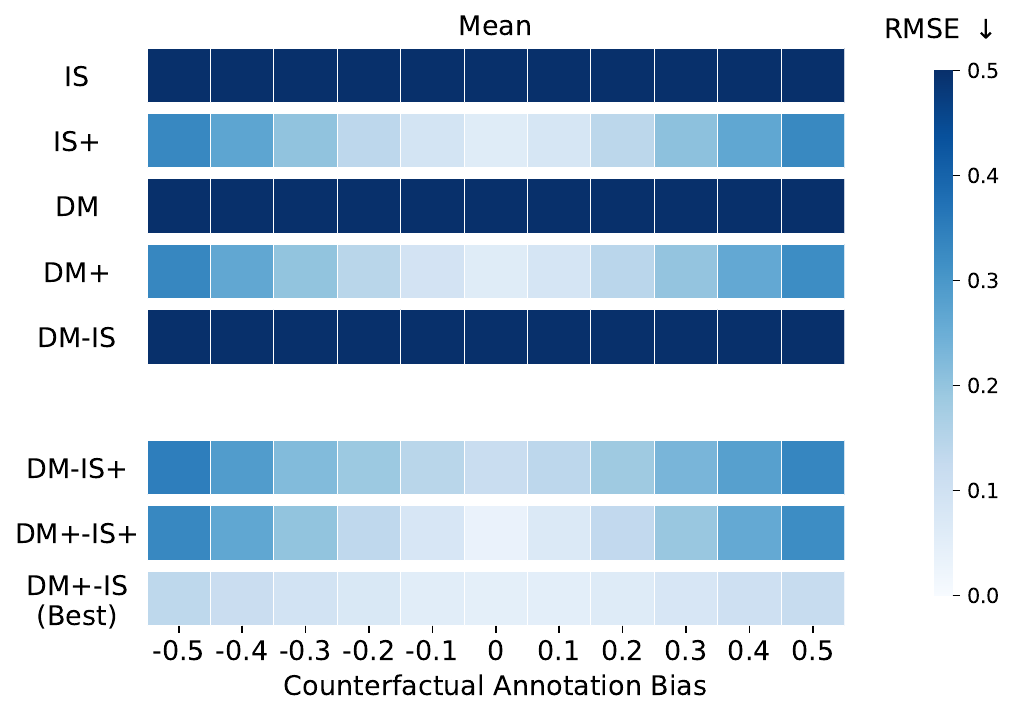}}
  }
\end{figure}
In \Cref{sec:exp_imperfect_annot_ws_reward}, we isolated the effect of imperfect annotations by assuming a well-specified reward model. 
However, many realistic healthcare settings often involve both reward model misspecification and imperfect annotations. In this section, we  demonstrate that \textbf{$\CDMIS$ is most robust to these combined sources of error.} 
\begin{figure}[t!]
  \centering
\includegraphics[width=\linewidth]{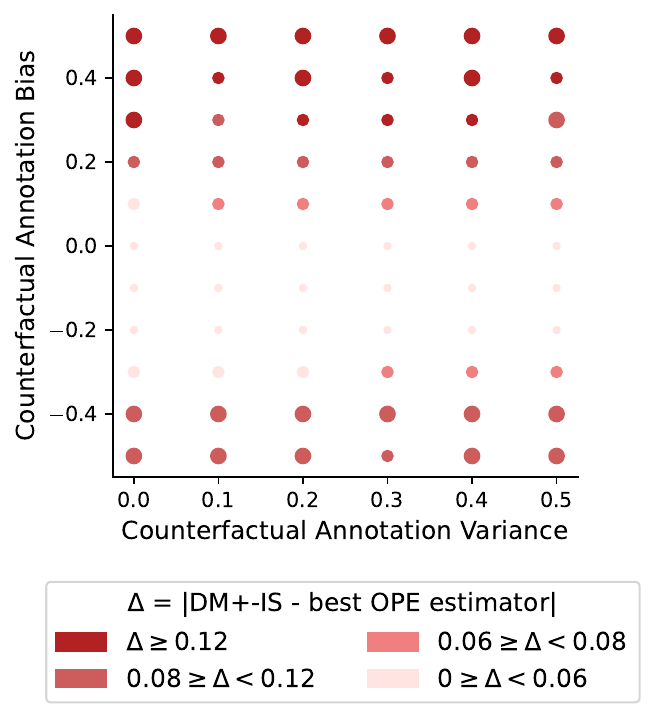}
  \caption{\textbf{$\CDMIS$ is a robust estimator across annotations of various qualities.} Here, we study the performance of $\CDMIS$ with a misspecified reward model in comparison to the best-performing OPE method in the sepsis domain. The $x,y$-axis represents the variance ($\Delta_G$) and the bias ($\epsilon_G$) of the annotations, respectively. We report $\Delta = v^\CDMIS - v^{\text{best estimate}}|$ where the best estimate is identified by observing prediction error from all baseline OPE estimators. The color of the dots represents the magnitude of mean $\Delta$, and the size of the dots is proportional to the variance of $\Delta$ across 100 iterations. We find that regardless of how imperfect the annotations are, $\CDMIS$ produces estimates with low mean $\Delta$ relative to the range of reward in the sepsis environment (5.2). $\Delta$ increases in magnitude proportional to the magnitude of the annotation bias, though even in extreme cases, $\Delta$ is relatively small.}
  \label{fig:practical_considerations}
\end{figure}

In \Cref{sec:exp_imperfect_annot_ws_reward}, we observed that the annotation noise has a limited effect on the RMSE of the proposed estimators. We therefore shift our focus to annotation bias. We first examine the Heartsteps and sepsis treatment domains, where the magnitude of bias and noise can be controlled. \Cref{fig:kitchen} shows that $\CDMIS$ is most resilient to the compounded errors arising from biased annotations and misspecified reward models. Across both semi-synthetic domains, we see that $\CDMIS$ consistently achieves the lowest RMSE or performs comparably to the best performing baseline. We attribute this robustness to the fact that
$\CDMIS$ is the only proposed estimator that is unbiased in the presence of imperfect annotations, and retains beneficial theoretical properties even with a misspecified reward model. 

To further demonstrate the robustness of $\CDMIS$, we compare it to the best-performing OPE estimator under varying degrees of annotation imperfection in the sepsis environment (\Cref{fig:practical_considerations}). Here, the best-performing OPE estimator is given access to a well-specified reward model and oracle knowledge of annotation quality; in particular, it uses counterfactual annotations only when they are perfect.

Our results indicate that \textbf{$\CDMIS$ remains robust, achieving RMSE within a small margin of the best achievable OPE performance despite these informational disadvantages}. 
In particular, $\Delta$ (the performance gap between $\CDMIS$ and the best-performing OPE method) is small relative to the reward range in the sepsis environment, indicating that $\CDMIS$ produces estimates of $v(\pi_e)$ close to those of the best-performing OPE method. 

\begin{figure}[htbp]
\floatconts
  {fig:mimic-iv_results}
  {\caption{\textbf{$\CDMIS$ outperforms baselines on MIMIC-IV dataset.} 
  We first study how the estimator performance varies as more annotations are added (\Cref{fig:mimic-iv_perf_dm+is}) and study the RMSE of all estimators given a small number of counterfactual annotations (\Cref{fig:mimic-iv_perf_m=100}). }}
  {%
  \subfigure[\textbf{As the number of counterfactual annotations increases, the performance of $\CDMIS$ initially improves and then plateaus.} Error bars represent 95\% confidence intervals. In contrast, other estimators either have higher RMSE or have worse performance as the number of annotations increase.]{%
      \label{fig:mimic-iv_perf_dm+is}%
      \includegraphics[width=\linewidth]{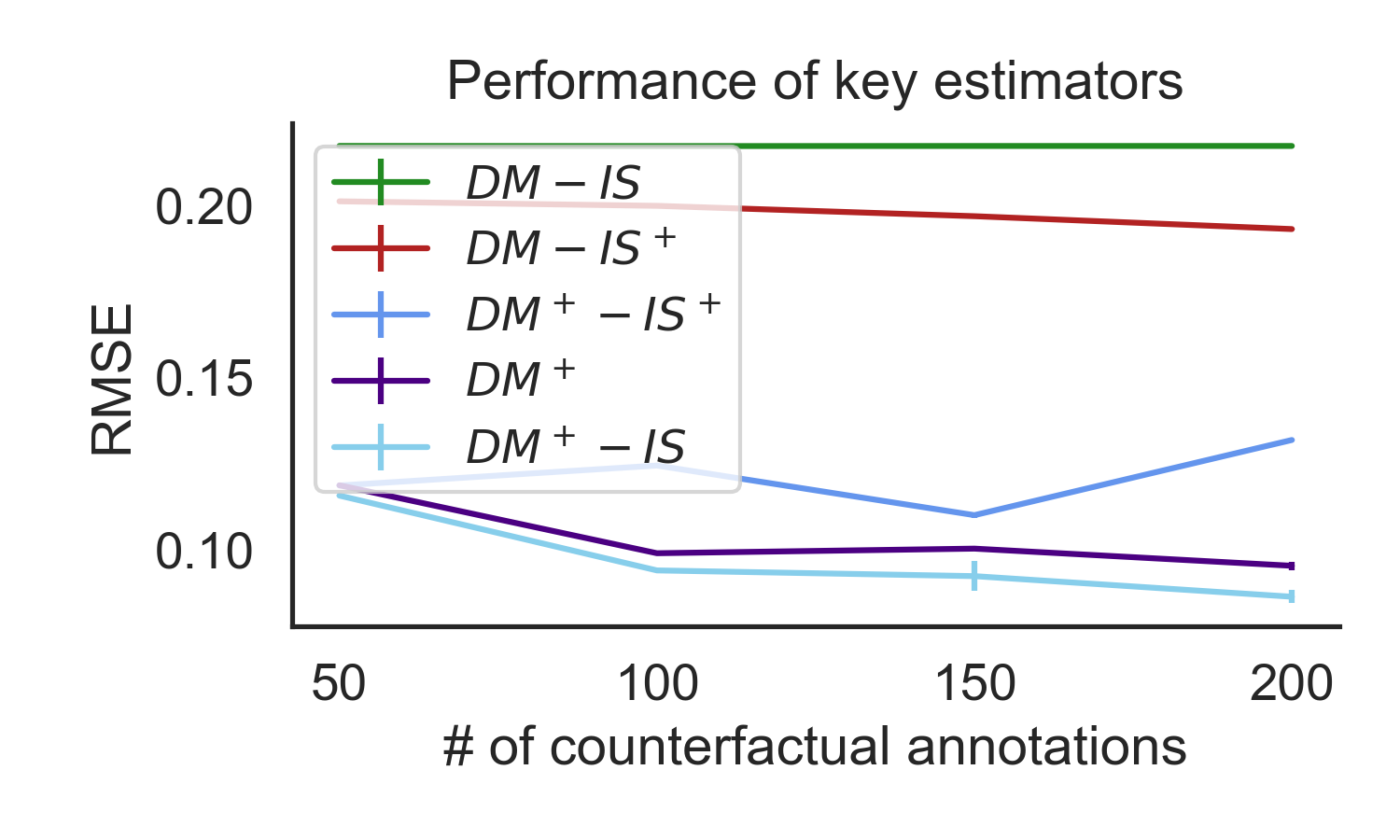}
    }\hfill
    \qquad
    \subfigure[\textbf{$\CDMIS$ outperforms all estimators with 100 counterfactual annotations.} Error bars represent 95\% confidence intervals, and $\CDMIS$ outperforms baselines with no overlapping intervals.]{%
      \label{fig:mimic-iv_perf_m=100}%
      \includegraphics[width=\linewidth]{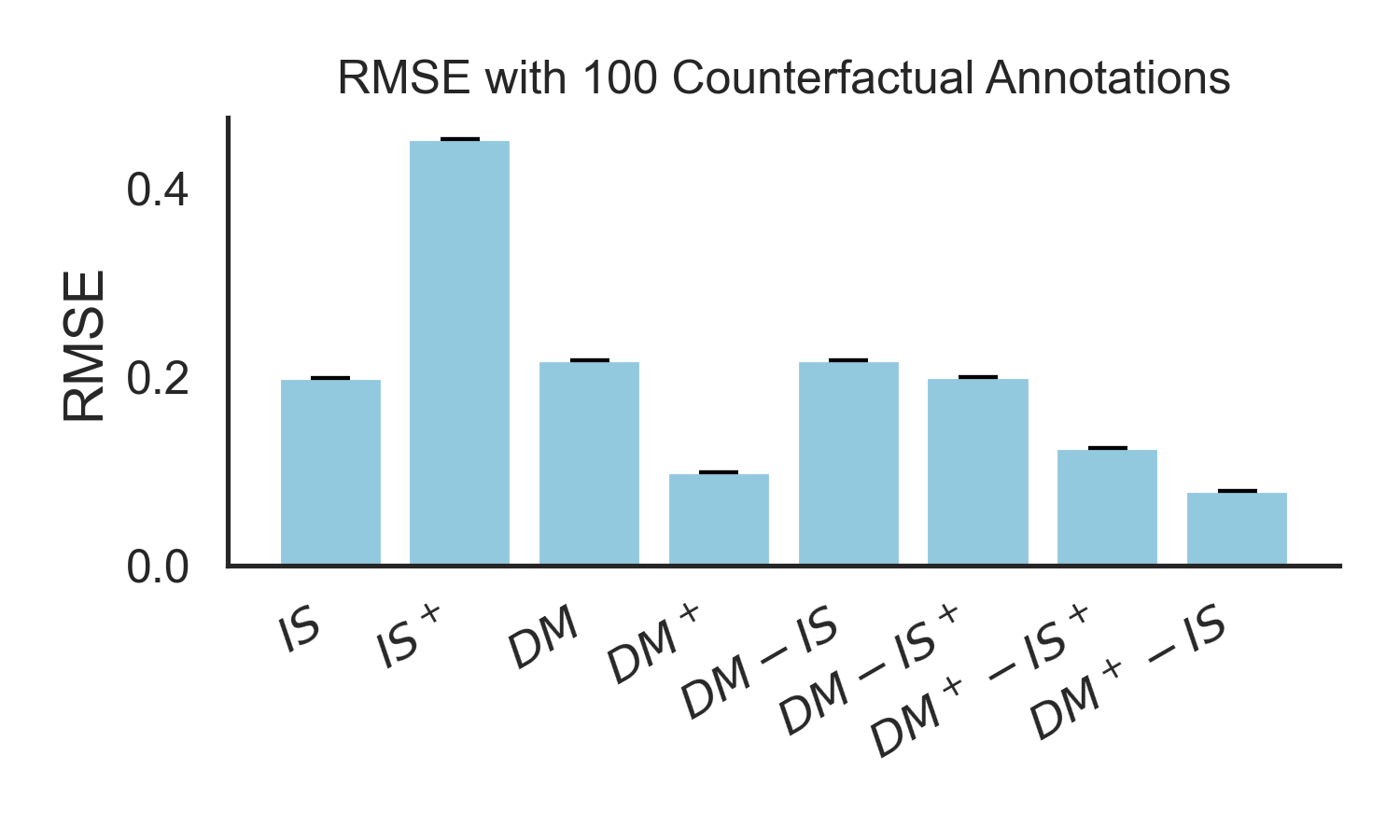}
    }
  }
\end{figure}

\subsubsection{Robustness analysis of $\CDMIS$ under realistic conditions}
In \Cref{sec:exp_imperfect_annot_mis_reward}, we examined the robustness of the proposed estimators under controlled conditions. However, in many real-world settings, the reward model and annotation quality are typically unknown. To evaluate performance under realistic conditions, we apply our proposed OPE estimators to potassium administration using the MIMIC-IV database. 

We first examine how the performance of key OPE estimators varies as the number of counterfactual annotations increases (\Cref{fig:mimic-iv_perf_dm+is}). While the RMSE of $\DMCIS$ and $\CDMCIS$ remains high as annotation count increases, \textbf{the RMSE of $\CDMIS$ decreases initially and then plateaus with no evidence of degradation.} This pattern suggests that $\CDMIS$ leverages the counterfactual annotations more effectively than baseline estimators. These trends are consistent with those observed in the synthetic or semi-synthetic experiments (\Cref{fig:kitchen,fig:bias_variance}).

We next examine all OPE estimators under a limited annotation budget (100 counterfactual annotations). We find that \textbf{$\CDMIS$ achieves the lowest RMSE across all baselines (\Cref{fig:mimic-iv_perf_m=100})}. Notably, $\CIS$ exhibits the highest error, which suggests that the counterfactual annotations are imperfect. The relatively small performance gap between $\CDM$ and $\CDMIS$ implies that the estimated reward model is reasonably accurate in this setting. In summary, these results demonstrate that $\CDMIS$ is well suited to realistic conditions in which the reward model and annotation quality are unknown. 

\section{Conclusion}
\label{sec:conclusion}
In this work, we address the open problem of incorporating imperfect counterfactual annotations into an OPE estimator. This is a research question of growing importance: as LLMs become increasingly capable of generating realistic synthetic samples, a broader challenge emerges of how to integrate such samples into statistical estimators without inheriting data biases.
We systematically study a range of design choices for integrating annotations into a DR-style estimators. We find that imperfect counterfactual annotations are most beneficial when incorporated into just the DM part of such an estimator. Through theoretical and empirical analyses, we find that the performance of an OPE estimator relies most on two critical factors: (1) whether the reward model is well-specified, and (2) the annotation quality. We conclude that under the most realistic conditions, where the reward model and annotation quality are unknown, our $\CDMIS$ estimator is most robust in comparison to baseline approaches. Overall, our approach relaxes restrictive assumptions about annotation quality and facilitates more reliable use of bandit algorithms in high-stakes applications through improved off-policy evaluation.

\textbf{Limitations and Future Work.} This work focuses on the contextual bandit setting, with future directions including extensions to the MDP setting. Additionally, this work considers a subset of possible reward function parameterizations. A promising avenue for future work includes optimizing the use of a limited budget of counterfactual annotations, and identifying which counterfactual actions to label, in the spirit of ~\citet{zrnic2026activestatisticalinference}. Furthermore, future work can consider identifying practical ways to pre-process a set of counterfactual annotations to mitigate the effect of bias and noise. 

\acks{The authors would like to thank Emma Brunskill, Matthew J\"orke, Yash Chandak, Allen Nie, Alex Nam, and Jared Moore for feedback on versions of this manuscript. AM was funded in part by a Stanford Data Science Fellowship. ST was funded in part by the Emory URC Research Award. JY was funded by Columbia University Data Science Institute. BEE was funded in part by grants from the Parker Institute for Cancer Immunology (PICI), the Chan-Zuckerberg Institute (CZI), the Biswas Family Foundation, NIH NHGRI R01 HG012967, and NIH NHGRI R01 HG013736.
BEE is a CIFAR Fellow in the Multiscale Human Program. BEE is on the Scientific Advisory Board for ArrePath Inc, GSK AI for Cancer, and Freenome; she consults for Neumora.}
\bibliography{ref}

\begin{thebibliography}{25}
\providecommand{\natexlab}[1]{#1}
\providecommand{\url}[1]{\texttt{#1}}
\expandafter\ifx\csname urlstyle\endcsname\relax
  \providecommand{\doi}[1]{doi: #1}\else
  \providecommand{\doi}{doi: \begingroup \urlstyle{rm}\Url}\fi

\bibitem[Beygelzimer and Langford(2009)]{beygelzimer_2009}
Alina Beygelzimer and John Langford.
\newblock The offset tree for learning with partial labels.
\newblock In \emph{Proceedings of the 15th ACM SIGKDD International Conference on Knowledge Discovery and Data Mining}, KDD '09, page 129–138, New York, NY, USA, 2009. Association for Computing Machinery.
\newblock ISBN 9781605584959.
\newblock \doi{10.1145/1557019.1557040}.
\newblock URL \url{https://doi.org/10.1145/1557019.1557040}.

\bibitem[Dudik et~al.(2011)Dudik, Langford, and Li]{dudik2011doubly}
Miroslav Dudik, John Langford, and Lihong Li.
\newblock Doubly robust policy evaluation and learning, 2011.

\bibitem[Dudík et~al.(2014)Dudík, Erhan, Langford, and Li]{Dud_k_2014}
Miroslav Dudík, Dumitru Erhan, John Langford, and Lihong Li.
\newblock Doubly robust policy evaluation and optimization.
\newblock \emph{Statistical Science}, 29\penalty0 (4), November 2014.
\newblock ISSN 0883-4237.
\newblock \doi{10.1214/14-sts500}.
\newblock URL \url{http://dx.doi.org/10.1214/14-STS500}.

\bibitem[Farajtabar et~al.(2018)Farajtabar, Chow, and Ghavamzadeh]{farajtabar2018robust}
Mehrdad Farajtabar, Yinlam Chow, and Mohammad Ghavamzadeh.
\newblock More robust doubly robust off-policy evaluation, 2018.

\bibitem[Goldberger et~al.(2000)Goldberger, Amaral, Glass, Hausdorff, Ivanov, Mark, Mietus, Moody, Peng, and Stanley]{physionet}
Ary~L Goldberger, Luis~AN Amaral, Leon Glass, Jeffrey~M Hausdorff, Plamen~Ch Ivanov, Roger~G Mark, Joseph~E Mietus, George~B Moody, Chung-Kang Peng, and H~Eugene Stanley.
\newblock Physiobank, physiotoolkit, and physionet: components of a new research resource for complex physiologic signals.
\newblock \emph{Circulation}, 2000.

\bibitem[Harutyunyan et~al.(2016)Harutyunyan, Bellemare, Stepleton, and Munos]{harutyunyan2016qlambdaoffpolicycorrections}
Anna Harutyunyan, Marc~G. Bellemare, Tom Stepleton, and Remi Munos.
\newblock Q($\lambda$) with off-policy corrections, 2016.
\newblock URL \url{https://arxiv.org/abs/1602.04951}.

\bibitem[Horvitz and Thompson(1952)]{IPS_horvitz}
D.~G. Horvitz and D.~J. Thompson.
\newblock A generalization of sampling without replacement from a finite universe.
\newblock \emph{Journal of the American Statistical Association}, 47\penalty0 (260):\penalty0 663--685, 1952.
\newblock ISSN 01621459.
\newblock URL \url{http://www.jstor.org/stable/2280784}.

\bibitem[Jiang and Li(2016)]{jiang2016doubly}
Nan Jiang and Lihong Li.
\newblock Doubly robust off-policy value evaluation for reinforcement learning, 2016.

\bibitem[Johnson et~al.(2020)Johnson, Bulgarelli, Pollard, Horng, Celi, and Mark~IV]{mimicivdataset}
Alistair Johnson, Lucas Bulgarelli, Tom Pollard, Steven Horng, Leo~Anthony Celi, and R~Mark~IV.
\newblock Mimic-iv (version 0.4).
\newblock \emph{PhysioNet}, 2020.

\bibitem[Klasnja et~al.(2019)Klasnja, Smith, Seewald, Lee, Hall, Luers, Hekler, and Murphy]{heartsteps}
Predrag Klasnja, Shawna Smith, Nicholas~J Seewald, Andy Lee, Kelly Hall, Brook Luers, Eric~B Hekler, and Susan~A Murphy.
\newblock Efficacy of contextually tailored suggestions for physical activity: a micro-randomized optimization trial of heartsteps.
\newblock \emph{Annals of Behavioral Medicine}, 53\penalty0 (6):\penalty0 573--582, 2019.

\bibitem[Le et~al.(2019)Le, Voloshin, and Yue]{le2019batchpolicylearningconstraints}
Hoang~M. Le, Cameron Voloshin, and Yisong Yue.
\newblock Batch policy learning under constraints, 2019.
\newblock URL \url{https://arxiv.org/abs/1903.08738}.

\bibitem[Li et~al.(2010)Li, Chu, Langford, and Schapire]{Li_2010}
Lihong Li, Wei Chu, John Langford, and Robert~E. Schapire.
\newblock A contextual-bandit approach to personalized news article recommendation.
\newblock In \emph{Proceedings of the 19th international conference on World wide web}, WWW ’10. ACM, April 2010.
\newblock \doi{10.1145/1772690.1772758}.
\newblock URL \url{http://dx.doi.org/10.1145/1772690.1772758}.

\bibitem[Mandyam et~al.(2024)Mandyam, Jörke, Denton, Engelhardt, and Brunskill]{mandyam2024adaptive}
Aishwarya Mandyam, Matthew Jörke, William Denton, Barbara~E. Engelhardt, and Emma Brunskill.
\newblock Adaptive interventions with user-defined goals for health behavior change, 2024.

\bibitem[Oberst and Sontag(2019)]{oberst2019counterfactualoffpolicyevaluationgumbelmax}
Michael Oberst and David Sontag.
\newblock Counterfactual off-policy evaluation with gumbel-max structural causal models, 2019.
\newblock URL \url{https://arxiv.org/abs/1905.05824}.

\bibitem[OpenAI et~al.(2024)OpenAI, :, Jaech, Kalai, Lerer, Richardson, El-Kishky, Low, Helyar, Madry, Beutel, Carney, Iftimie, Karpenko, Passos, Neitz, Prokofiev, Wei, Tam, Bennett, Kumar, Saraiva, Vallone, Duberstein, Kondrich, Mishchenko, Applebaum, Jiang, Nair, Zoph, Ghorbani, Rossen, Sokolowsky, Barak, McGrew, Minaiev, Hao, Baker, Houghton, McKinzie, Eastman, Lugaresi, Bassin, Hudson, Li, de~Bourcy, Voss, Shen, Zhang, Koch, Orsinger, Hesse, Fischer, Chan, Roberts, Kappler, Levy, Selsam, Dohan, Farhi, Mely, Robinson, Tsipras, Li, Oprica, Freeman, Zhang, Wong, Proehl, Cheung, Mitchell, Wallace, Ritter, Mays, Wang, Such, Raso, Leoni, Tsimpourlas, Song, von Lohmann, Sulit, Salmon, Parascandolo, Chabot, Zhao, Brockman, Leclerc, Salman, Bao, Sheng, Andrin, Bagherinezhad, Ren, Lightman, Chung, Kivlichan, O'Connell, Osband, Gilaberte, Akkaya, Kostrikov, Sutskever, Kofman, Pachocki, Lennon, Wei, Harb, Twore, Feng, Yu, Weng, Tang, Yu, Candela, Palermo, Parish, Heidecke, Hallman, Rizzo, Gordon, Uesato, Ward,
  Huizinga, Wang, Chen, Xiao, Singhal, Nguyen, Cobbe, Shi, Wood, Rimbach, Gu-Lemberg, Liu, Lu, Stone, Yu, Ahmad, Yang, Liu, Maksin, Ho, Fedus, Weng, Li, McCallum, Held, Kuhn, Kondraciuk, Kaiser, Metz, Boyd, Trebacz, Joglekar, Chen, Tintor, Meyer, Jones, Kaufer, Schwarzer, Shah, Yatbaz, Guan, Xu, Yan, Glaese, Chen, Lampe, Malek, Wang, Fradin, McClay, Pavlov, Wang, Wang, Murati, Bavarian, Rohaninejad, McAleese, Chowdhury, Chowdhury, Ryder, Tezak, Brown, Nachum, Boiko, Murk, Watkins, Chao, Ashbourne, Izmailov, Zhokhov, Dias, Arora, Lin, Lopes, Gaon, Miyara, Leike, Hwang, Garg, Brown, James, Shu, Cheu, Greene, Jain, Altman, Toizer, Toyer, Miserendino, Agarwal, Hernandez, Baker, McKinney, Yan, Zhao, Hu, Santurkar, Chaudhuri, Zhang, Fu, Papay, Lin, Balaji, Sanjeev, Sidor, Broda, Clark, Wang, Gordon, Sanders, Patwardhan, Sottiaux, Degry, Dimson, Zheng, Garipov, Stasi, Bansal, Creech, Peterson, Eloundou, Qi, Kosaraju, Monaco, Pong, Fomenko, Zheng, Zhou, McCabe, Zaremba, Dubois, Lu, Chen, Cha, Bai, He, Zhang, Wang,
  Shao, and Li]{openai2024openaio1card}
OpenAI, :, Aaron Jaech, Adam Kalai, Adam Lerer, Adam Richardson, Ahmed El-Kishky, Aiden Low, Alec Helyar, Aleksander Madry, Alex Beutel, Alex Carney, Alex Iftimie, Alex Karpenko, Alex~Tachard Passos, Alexander Neitz, Alexander Prokofiev, Alexander Wei, Allison Tam, Ally Bennett, Ananya Kumar, Andre Saraiva, Andrea Vallone, Andrew Duberstein, Andrew Kondrich, Andrey Mishchenko, Andy Applebaum, Angela Jiang, Ashvin Nair, Barret Zoph, Behrooz Ghorbani, Ben Rossen, Benjamin Sokolowsky, Boaz Barak, Bob McGrew, Borys Minaiev, Botao Hao, Bowen Baker, Brandon Houghton, Brandon McKinzie, Brydon Eastman, Camillo Lugaresi, Cary Bassin, Cary Hudson, Chak~Ming Li, Charles de~Bourcy, Chelsea Voss, Chen Shen, Chong Zhang, Chris Koch, Chris Orsinger, Christopher Hesse, Claudia Fischer, Clive Chan, Dan Roberts, Daniel Kappler, Daniel Levy, Daniel Selsam, David Dohan, David Farhi, David Mely, David Robinson, Dimitris Tsipras, Doug Li, Dragos Oprica, Eben Freeman, Eddie Zhang, Edmund Wong, Elizabeth Proehl, Enoch Cheung, Eric
  Mitchell, Eric Wallace, Erik Ritter, Evan Mays, Fan Wang, Felipe~Petroski Such, Filippo Raso, Florencia Leoni, Foivos Tsimpourlas, Francis Song, Fred von Lohmann, Freddie Sulit, Geoff Salmon, Giambattista Parascandolo, Gildas Chabot, Grace Zhao, Greg Brockman, Guillaume Leclerc, Hadi Salman, Haiming Bao, Hao Sheng, Hart Andrin, Hessam Bagherinezhad, Hongyu Ren, Hunter Lightman, Hyung~Won Chung, Ian Kivlichan, Ian O'Connell, Ian Osband, Ignasi~Clavera Gilaberte, Ilge Akkaya, Ilya Kostrikov, Ilya Sutskever, Irina Kofman, Jakub Pachocki, James Lennon, Jason Wei, Jean Harb, Jerry Twore, Jiacheng Feng, Jiahui Yu, Jiayi Weng, Jie Tang, Jieqi Yu, Joaquin~Quiñonero Candela, Joe Palermo, Joel Parish, Johannes Heidecke, John Hallman, John Rizzo, Jonathan Gordon, Jonathan Uesato, Jonathan Ward, Joost Huizinga, Julie Wang, Kai Chen, Kai Xiao, Karan Singhal, Karina Nguyen, Karl Cobbe, Katy Shi, Kayla Wood, Kendra Rimbach, Keren Gu-Lemberg, Kevin Liu, Kevin Lu, Kevin Stone, Kevin Yu, Lama Ahmad, Lauren Yang, Leo Liu,
  Leon Maksin, Leyton Ho, Liam Fedus, Lilian Weng, Linden Li, Lindsay McCallum, Lindsey Held, Lorenz Kuhn, Lukas Kondraciuk, Lukasz Kaiser, Luke Metz, Madelaine Boyd, Maja Trebacz, Manas Joglekar, Mark Chen, Marko Tintor, Mason Meyer, Matt Jones, Matt Kaufer, Max Schwarzer, Meghan Shah, Mehmet Yatbaz, Melody~Y. Guan, Mengyuan Xu, Mengyuan Yan, Mia Glaese, Mianna Chen, Michael Lampe, Michael Malek, Michele Wang, Michelle Fradin, Mike McClay, Mikhail Pavlov, Miles Wang, Mingxuan Wang, Mira Murati, Mo~Bavarian, Mostafa Rohaninejad, Nat McAleese, Neil Chowdhury, Neil Chowdhury, Nick Ryder, Nikolas Tezak, Noam Brown, Ofir Nachum, Oleg Boiko, Oleg Murk, Olivia Watkins, Patrick Chao, Paul Ashbourne, Pavel Izmailov, Peter Zhokhov, Rachel Dias, Rahul Arora, Randall Lin, Rapha~Gontijo Lopes, Raz Gaon, Reah Miyara, Reimar Leike, Renny Hwang, Rhythm Garg, Robin Brown, Roshan James, Rui Shu, Ryan Cheu, Ryan Greene, Saachi Jain, Sam Altman, Sam Toizer, Sam Toyer, Samuel Miserendino, Sandhini Agarwal, Santiago Hernandez,
  Sasha Baker, Scott McKinney, Scottie Yan, Shengjia Zhao, Shengli Hu, Shibani Santurkar, Shraman~Ray Chaudhuri, Shuyuan Zhang, Siyuan Fu, Spencer Papay, Steph Lin, Suchir Balaji, Suvansh Sanjeev, Szymon Sidor, Tal Broda, Aidan Clark, Tao Wang, Taylor Gordon, Ted Sanders, Tejal Patwardhan, Thibault Sottiaux, Thomas Degry, Thomas Dimson, Tianhao Zheng, Timur Garipov, Tom Stasi, Trapit Bansal, Trevor Creech, Troy Peterson, Tyna Eloundou, Valerie Qi, Vineet Kosaraju, Vinnie Monaco, Vitchyr Pong, Vlad Fomenko, Weiyi Zheng, Wenda Zhou, Wes McCabe, Wojciech Zaremba, Yann Dubois, Yinghai Lu, Yining Chen, Young Cha, Yu~Bai, Yuchen He, Yuchen Zhang, Yunyun Wang, Zheng Shao, and Zhuohan Li.
\newblock Openai o1 system card, 2024.
\newblock URL \url{https://arxiv.org/abs/2412.16720}.

\bibitem[Prasad et~al.(2022)Prasad, Mandyam, Chivers, Draugelis, Hanson, Engelhardt, and Laudanski]{prasadpotassiumrepletion}
Niranjani Prasad, Aishwarya Mandyam, Corey Chivers, Michael Draugelis, Clarence Hanson, Barbara Engelhardt, and Krzysztof Laudanski.
\newblock Guiding efficient, effective, and patient-oriented electrolyte replacement in critical care: An artificial intelligence reinforcement learning approach.
\newblock \emph{Journal of Personalized Medicine}, 12:\penalty0 661, 04 2022.
\newblock \doi{10.3390/jpm12050661}.

\bibitem[Precup et~al.(2000)Precup, Sutton, and Singh]{eligibility_traces}
Doina Precup, Richard Sutton, and Satinder Singh.
\newblock Eligibility traces for off-policy policy evaluation.
\newblock \emph{Computer Science Department Faculty Publication Series}, 06 2000.

\bibitem[Shrimanker and Bhattarai(2025)]{Shrimanker2025Potassium}
Ishaan Shrimanker and Suman Bhattarai.
\newblock Potassium.
\newblock \url{https://www.ncbi.nlm.nih.gov/books/NBK539791/}, 2025.
\newblock URL \url{https://www.ncbi.nlm.nih.gov/books/NBK539791/}.
\newblock Accessed: 2025-05-15.

\bibitem[Sutton and Barto(2018)]{Sutton_Barto_2018}
Richard~S. Sutton and Andrew~G. Barto.
\newblock \emph{Reinforcement learning: an introduction}.
\newblock Adaptive computation and machine learning series. The MIT Press, second edition edition, 2018.
\newblock ISBN 9780262039246.

\bibitem[Tang and Wiens(2023)]{tang2023counterfactualaugmented}
Shengpu Tang and Jenna Wiens.
\newblock Counterfactual-augmented importance sampling for semi-offline policy evaluation.
\newblock In \emph{Thirty-seventh Conference on Neural Information Processing Systems}, 2023.
\newblock URL \url{https://openreview.net/forum?id=dsH244r9fA}.

\bibitem[Thomas and Brunskill(2016)]{thomas2016dataefficient}
Philip~S. Thomas and Emma Brunskill.
\newblock Data-efficient off-policy policy evaluation for reinforcement learning, 2016.

\bibitem[van Seijen et~al.(2009)van Seijen, Hasselt, Whiteson, and Wiering]{Seijen2009ATA}
Harm van Seijen, H.~V. Hasselt, Shimon Whiteson, and Marco~A Wiering.
\newblock A theoretical and empirical analysis of expected sarsa.
\newblock \emph{2009 IEEE Symposium on Adaptive Dynamic Programming and Reinforcement Learning}, pages 177--184, 2009.
\newblock URL \url{https://api.semanticscholar.org/CorpusID:6230754}.

\bibitem[Voloshin et~al.(2021)Voloshin, Le, Jiang, and Yue]{voloshin2021empirical}
Cameron Voloshin, Hoang~M. Le, Nan Jiang, and Yisong Yue.
\newblock Empirical study of off-policy policy evaluation for reinforcement learning, 2021.

\bibitem[Yao et~al.(2021)Yao, Brunskill, Pan, Murphy, and Doshi-Velez]{yao2021power}
Jiayu Yao, Emma Brunskill, Weiwei Pan, Susan Murphy, and Finale Doshi-Velez.
\newblock Power constrained bandits.
\newblock In \emph{Proceedings of the 6th Machine Learning for Healthcare Conference}, pages 209--259, 2021.

\bibitem[Zrnic and Candès(2026)]{zrnic2026activestatisticalinference}
Tijana Zrnic and Emmanuel~J. Candès.
\newblock Active statistical inference, 2026.
\newblock URL \url{https://arxiv.org/abs/2403.03208}.

\end{thebibliography}

\newpage
\onecolumn
\appendix
\setcounter{figure}{0}
\renewcommand{\thefigure}{S\arabic{figure}}
\renewcommand{\theproposition}{S\arabic{proposition}}
\renewcommand{\thecorollary}{S\arabic{corollary}}
\section{Additional Empirical Results}
  \label{apd:empirical_results}
 In the main text, we primarily focus on the case when the annotations are imperfect and the reward model is misspecified. We characterize the performance of the baseline OPE methods and our proposed approaches in four settings: either a well-specified or misspecified reward model, and either perfect or imperfect annotations. Here, we report additional results for each setting including error metrics on plots reported in the main text. 

\subsection{Well-Specified Reward}
\label{apd:ws_reward}
In \Cref{fig:bias_variance}, we note that the bias of the counterfactual annotation plays a larger role in affecting the RMSE of the proposed methods than the variance of the counterfactual annotation. 
Here, we report the mean and standard deviation of the RMSE across all datasets (\Cref{fig:ws_reward}). Our results indicate that $\CDMIS$ has the least fluctuation in standard deviation across the range of counterfactual annotation bias and variance in comparison to all methods that use counterfactual annotations. 

\begin{figure}[htbp]
\floatconts
  {fig:ws_reward} 
  {\caption{\textbf{Heatmap of mean and standard deviation of RMSE with a well-specified reward model (lower mean/standard deviation is represented lighter)}: The mean and standard deviation are reported with respect to the different combinations of behavior and target policies we consider. The $x,y$-axis represents the variance ($\Delta_G$) and the bias ($\epsilon_G$) of the annotations, respectively. In a well-specified reward setting, $\DM$ and $\DMIS$ perform comparably and best. Of note, $\IS$ sometimes performs well in this setting largely due to high coverage in the behavior policy for most pairings. In terms of standard deviation of RMSE, all methods that use counterfactual annotations with the exception of $\DMIS$ experience high standard deviation for the most imperfect annotations.}}
  {%
    \subfigure[2-context bandit.]{%
      \label{fig:apd_ws_2state}%
      \includegraphics[width=0.5\linewidth]{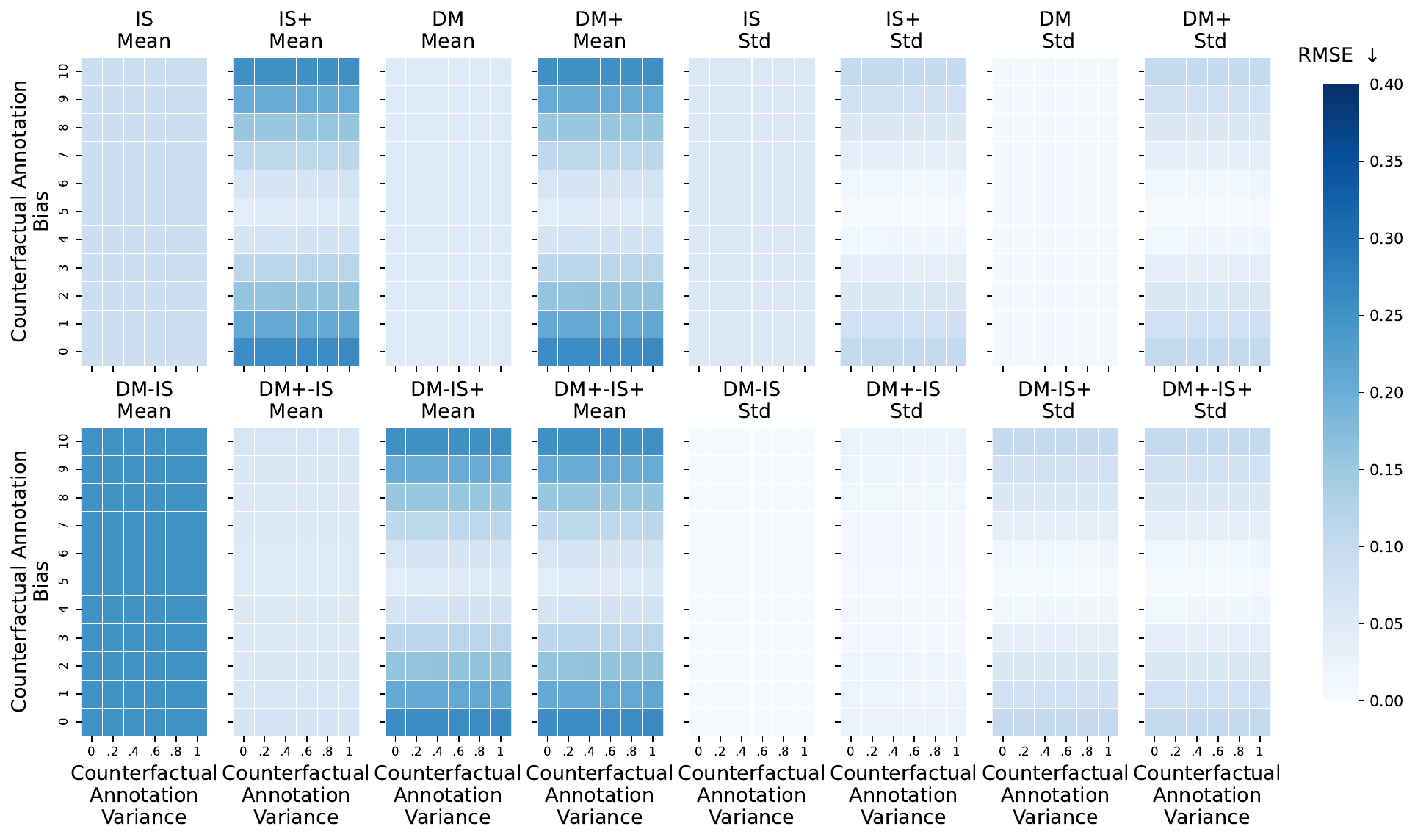}%
    }\\
    \subfigure[Heartsteps.]{%
      \label{fig:apd_ws_heartsteps}%
      \includegraphics[width=0.5\linewidth]{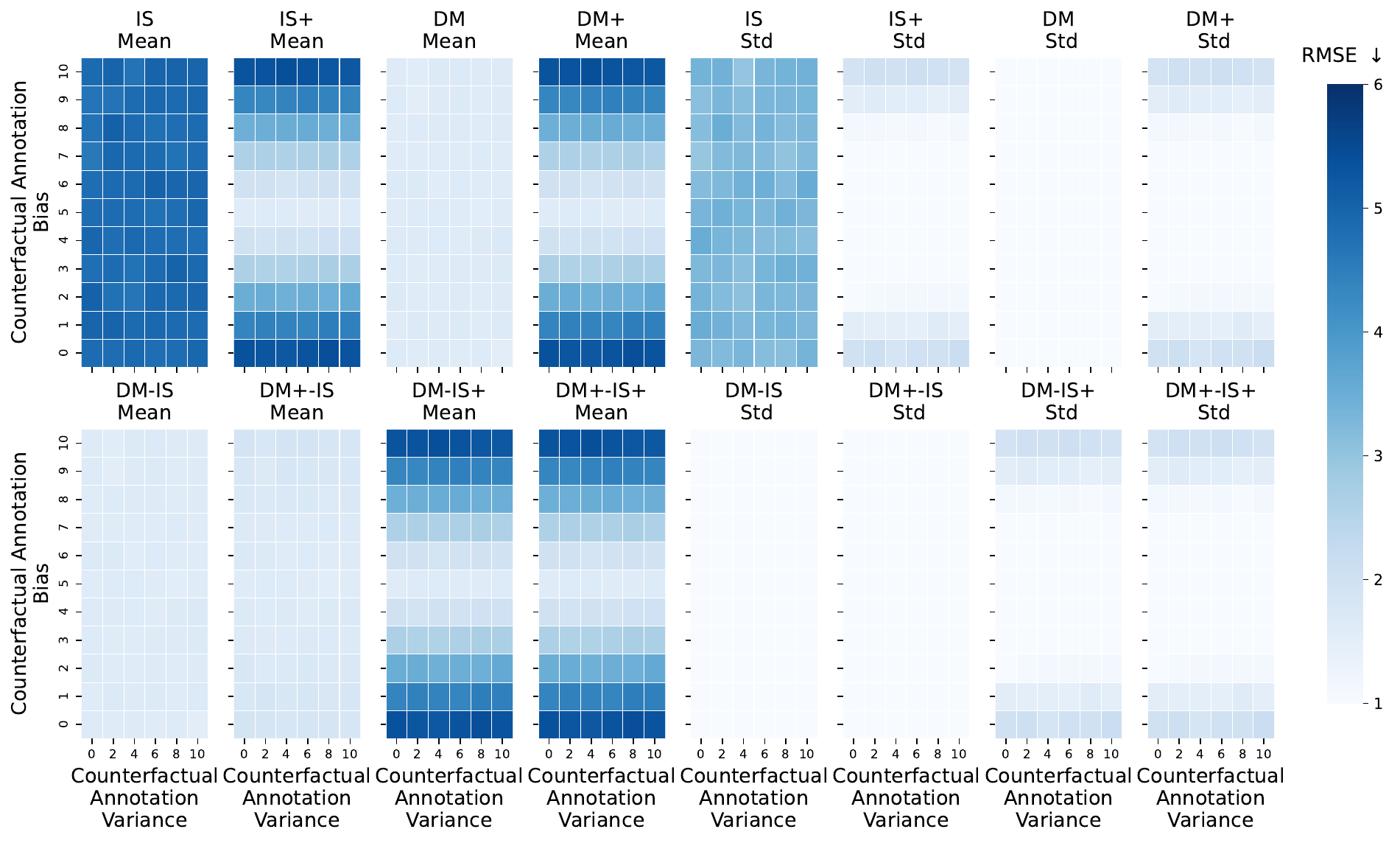}%
    }\\
    \subfigure[Sepsis.]{%
      \label{fig:apd_ws_sepsis}%
      \includegraphics[width=0.5\linewidth]{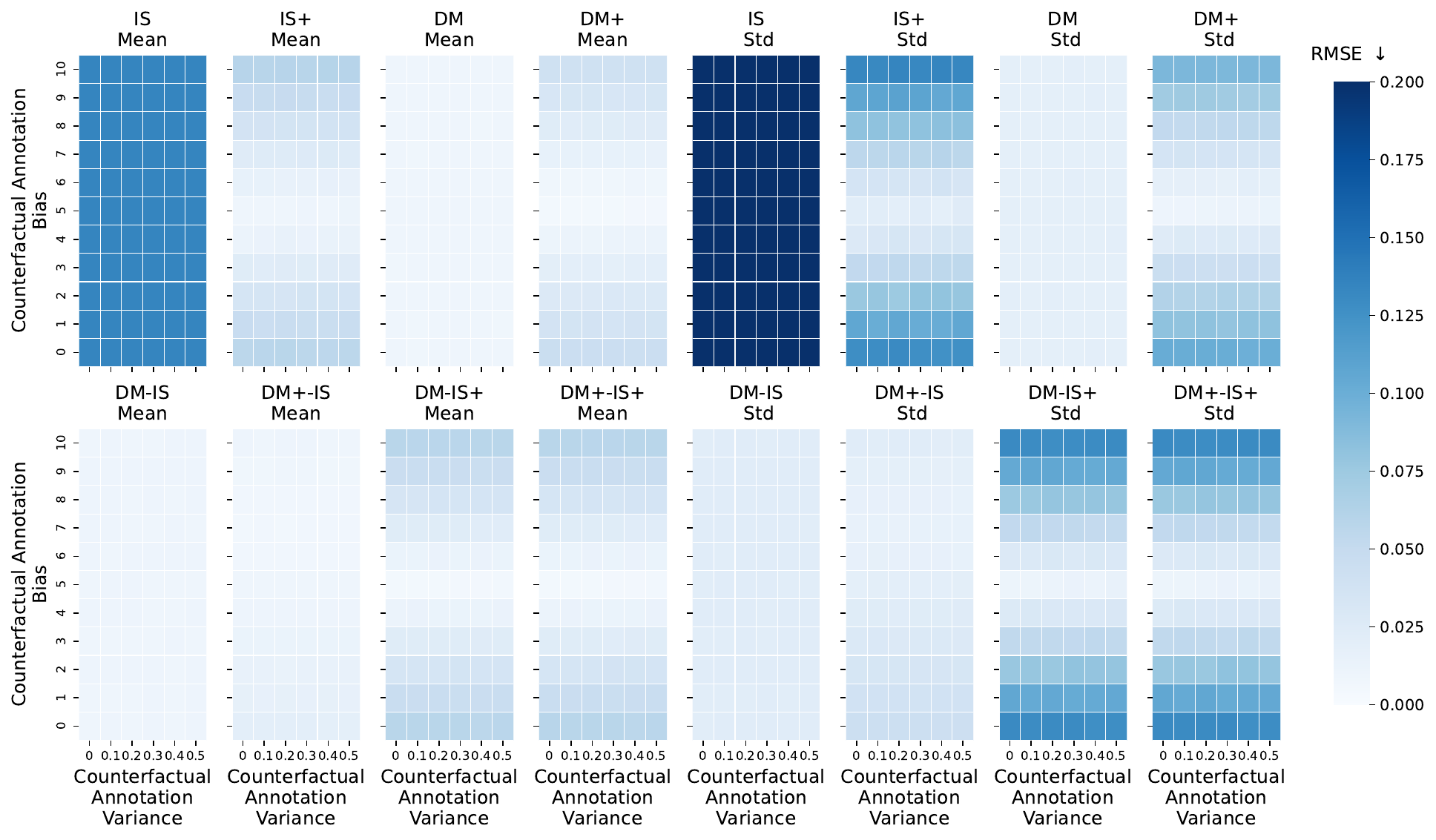}%
    }
  }
\end{figure}

\subsection{Misspecified Reward}
\label{apd:ms_reward}
In \Cref{fig:kitchen} we report mean RMSE across a variety of counterfactual annotation bias values in the Heartsteps and sepsis settings. In \Cref{fig:ms_reward}, we report the mean and standard deviation of RMSE across all three datasets.

\begin{figure}[htbp]
\floatconts
  {fig:ms_reward} 
  {\caption{\textbf{Heatmap of mean and standard deviation of RMSE with a misspecified reward model (lower mean/standard deviation is represented lighter)}:
    The $x,y$-axis represents the variance ($\Delta_G$) and the bias ($\epsilon_G$) of the annotations, respectively. In a misspecified reward setting, $\DM$ and $\CDM$ tend to suffer. In comparison, $\CDMIS$ outperforms all method or performs comparably to the best performing methods for both RMSE mean and standard deviation.}}
  {%
    \subfigure[2-context bandit.]{%
      \label{fig:apd_ms_2state}%
      \includegraphics[width=0.5\linewidth]{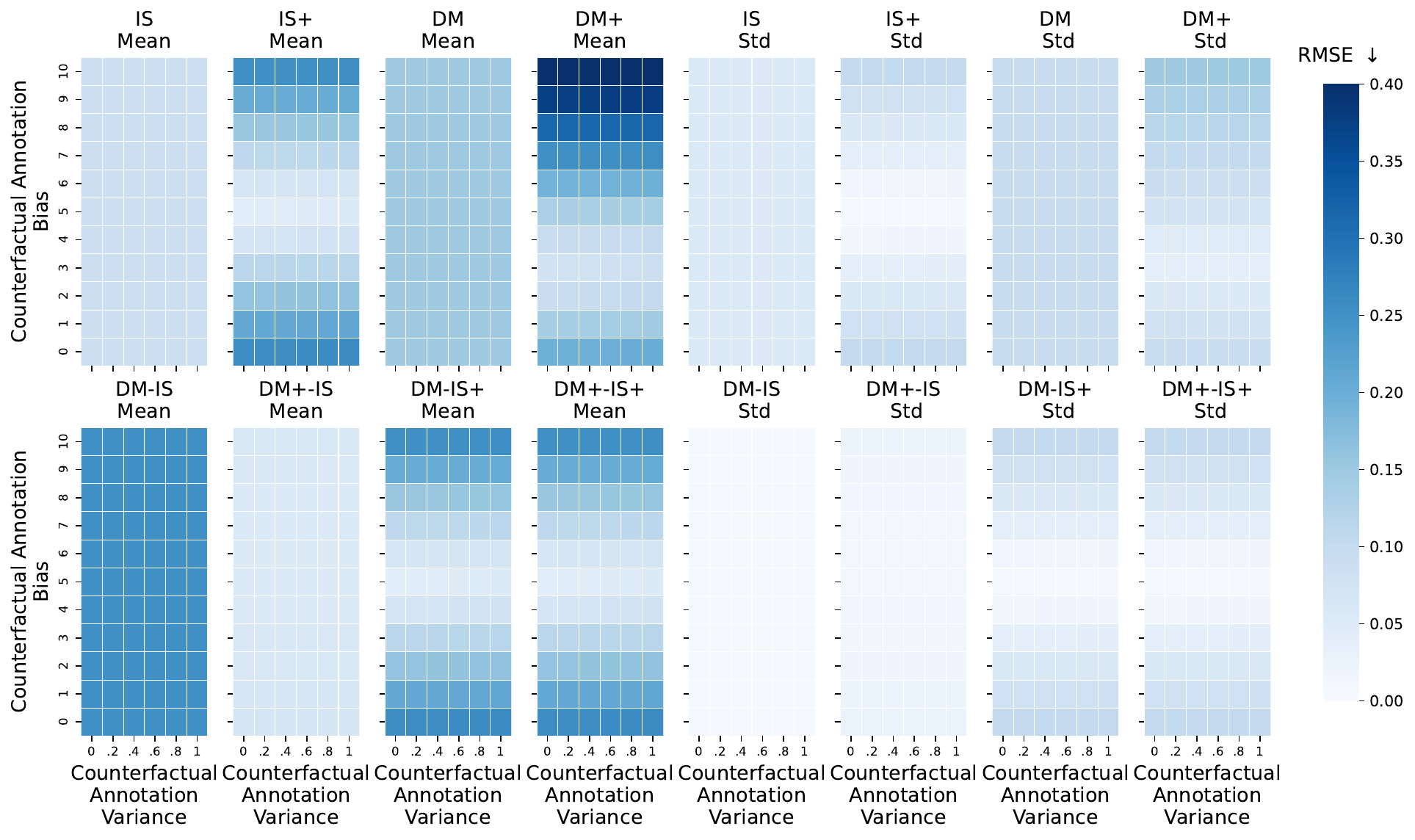}%
    }\\
    \subfigure[Heartsteps.]{%
      \label{fig:apd_ms_heartsteps}%
      \includegraphics[width=0.5\linewidth]{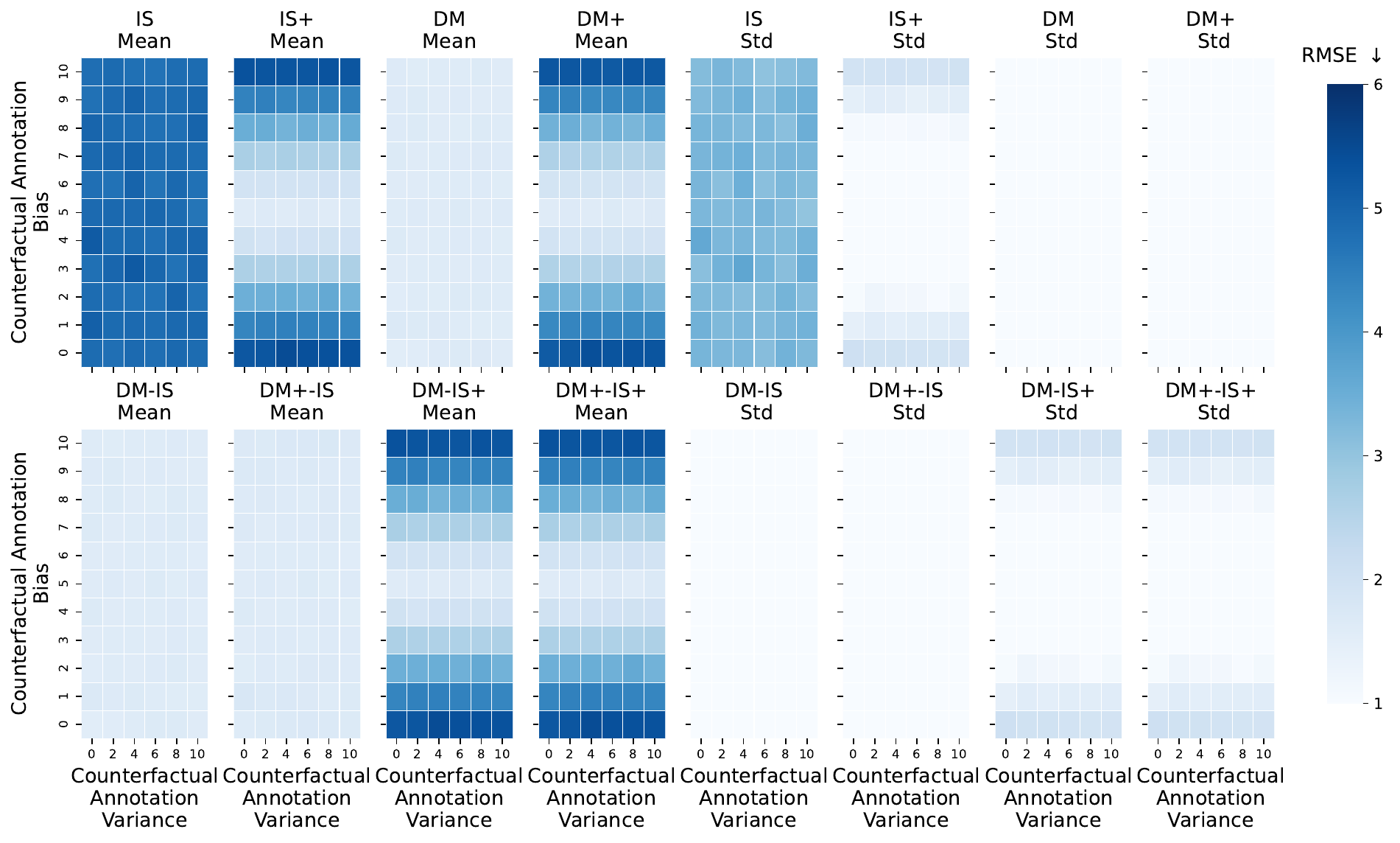}%
    }\\
    \subfigure[Sepsis.]{%
      \label{fig:apd_ms_sepsis}%
      \includegraphics[width=0.5\linewidth]{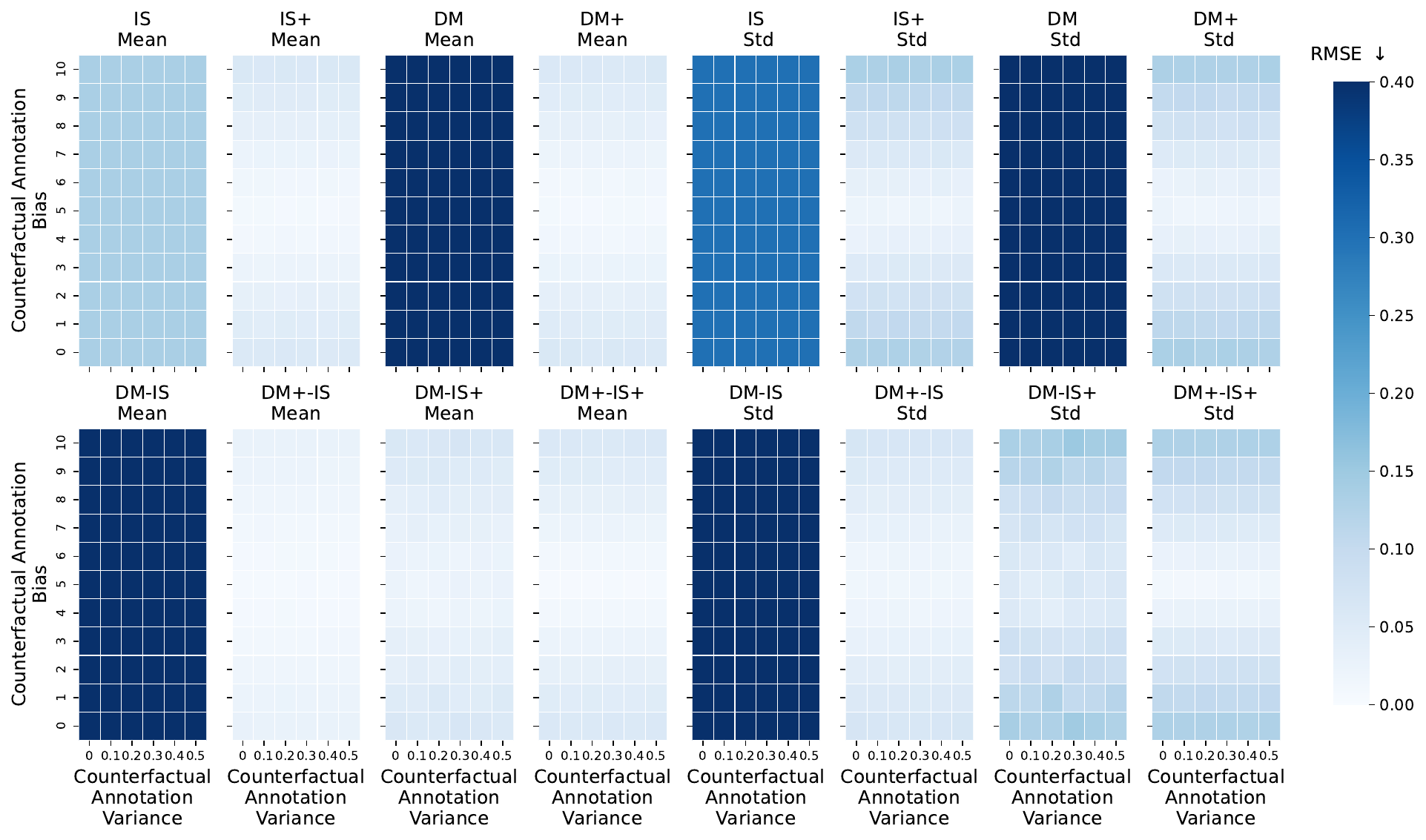}%
    }
  }
\end{figure}

In \Cref{fig:practical_considerations}, we find that $\CDMIS$ is robust to annotation quality and reward model misspecification. We report the same plot for $\DMCIS$ and $\CDMCIS$ in \Cref{fig:apd_practical_considerations}. 
Our conclusions remain identical to those discussed in \Cref{sec:experiments}. 


Finally, as discussed in \Cref{sec:methods}, our goal is to define an estimator that is ``doubly robust'' to two sources of error, namely the error of the annotation and the error of the reward model. As such, we do not account for imperfect IPS ratios and assume that the estimates of IPS ratios are fairly accurate. 
In the case that the IPS ratio is inaccurate, all proposed estimators will be biased. The estimation error of the IPS ratio will thus propagate through the bias and variance reductions introduced in our work. To further illustrate this, we report the performance of $\CDMIS$ with varying degrees of incorrectly estimated behavior policies $\epsilon$ (\Cref{fig:apd_unknown_ips}). We leave further evaluation of this setting to future work.

\begin{figure}[t!]
\floatconts
  {fig:apd_unknown_ips} 
  {\caption{\textbf{Analyzing the consequences of incorrect IPS ratios in $\CDMIS$ in the Two-context Bandit:} The $x,y$ axes correspond to the annotation variance and bias respectively. The heatmap color corresponds to the mean RMSE of the $\CDMIS$ estimator. We study three settings of $\epsilon$, where $\hat{\pi_b} = \pi_b + \epsilon$ is the estimated behavior policy. As $\epsilon$ increases, $\CDMIS$ becomes a more biased estimator.}} 
  {\includegraphics[width=0.4\linewidth]{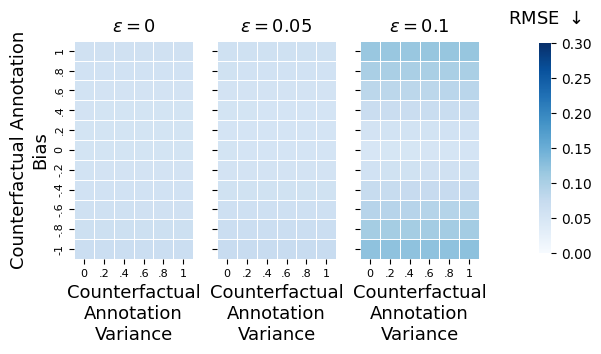}} 
\end{figure}

\subsection{Additional MIMIC-IV Results}
\begin{figure}[t!]
  \centering
  \includegraphics[width=\linewidth]{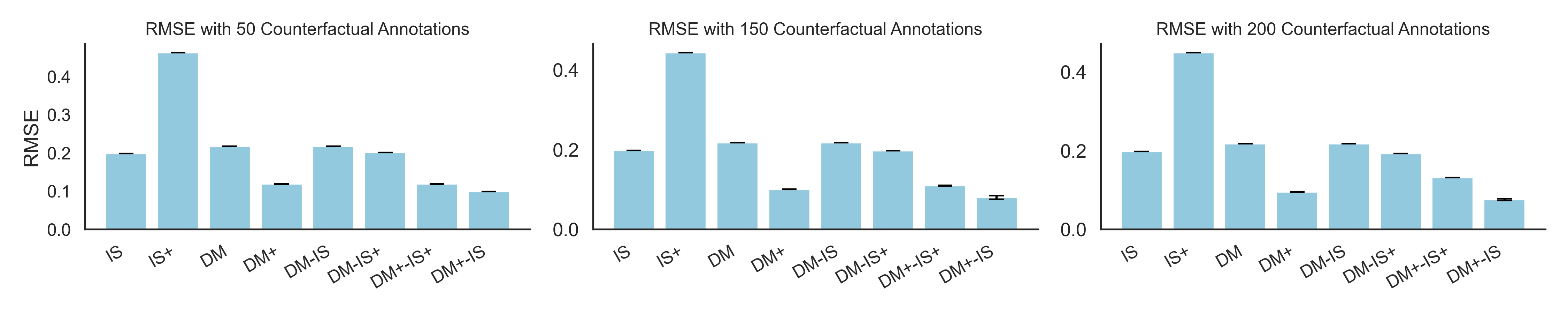}
  \caption{\textbf{Comparing the performance of all estimators in MIMIC-IV as we increase the number of counterfactual annotations.} The x-axis enumerates the possible estimators, and the y-axis measures the RMSE of the learned policy estimate. Error bars represent standard deviation. Each plot corresponds to the performance at different number of counterfactual annotations. We note that $\CDMIS$ has the most favorable performance across all settings. }
  \label{fig:mimic-iv_m}
\end{figure}

In \Cref{sec:results}, we compared the performance on MIMIC-IV data of all estimators with 100 counterfactual annotations. Now, we demonstrate that $\CDMIS$ has the most favorable performance across a wider range of $m$, the number of counterfactual annotations available (\Cref{fig:mimic-iv_m}). In particular, we note that the performance of $\CDMIS$ stays consistently better in comparison to the other estimators across all values of $m$ and that the performance either improves or remains consistent, but does not degrade if you increase the number of counterfactual annotations. 

\begin{figure}[htbp]
\floatconts
  {fig:apd_practical_considerations} 
  {\caption{Here we compare the consequences of choosing the other two proposed estimators ($\DMCIS$ and $\CDMCIS$). Note that across the range of possible annotation quality, $\DMCIS$ has a larger error (darker color) in comparison to $\CDMIS$ as displayed in \Cref{fig:practical_considerations}. While $\CDMCIS$ has smaller $\Delta$ when the counterfactual annotations have low variance, $\Delta$ more rapidly increases as the annotations become more imperfect.}}
  {%
    \subfigure[Exploring the consequences of choosing $\DMCIS$ when annotation and reward model quality are unknown.]{%
      \includegraphics[width=0.4\linewidth]{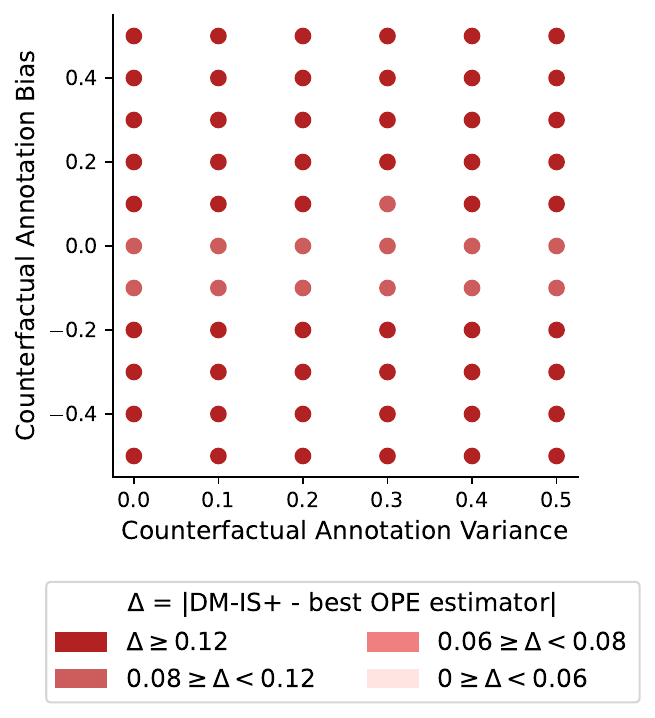}%
    }
    \subfigure[Exploring the consequences of choosing $\CDMCIS$ when annotation and reward model quality are unknown.]{%
      \includegraphics[width=0.4\linewidth]{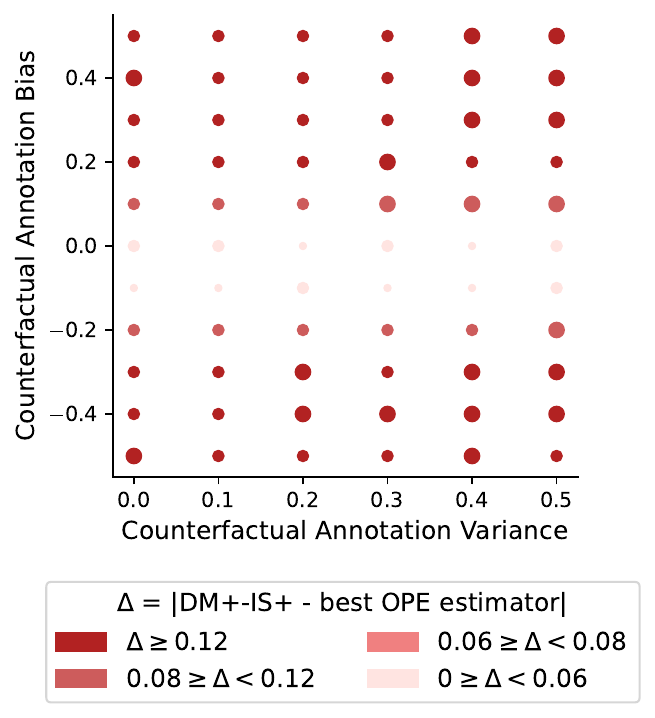}%
    }
  }
\end{figure}

\section{Simulator environments}
\label{apd:simulator_details}

In all of our experiments, we use a separate random seed to generate datasets. We report results across 50 runs, each of which is a sampled dataset. Our results across all experiments take approximately 100 hours of compute, which was run on a local university cluster. Our code is attached as a portion of the supplement material and will be made public upon acceptance via a Github link. Our code is covered under the MIT license. 
Here we discuss the implementation details for the three simulator settings we use in our empirical results. Key details of the settings can be observed in \Cref{tab:environments}.  
\label{apd:simulators}
\subsection{Two-context bandit}
The two-context bandit setting contains two contexts, each with two actions. The bandit receives reward for taking either action from the first context, and no reward for taking any action from the second context. In particular, the reward is sampled from the normal distribution $\mathcal{N}(1, 0.5)$ for the first action in the first context and the distribution $\mathcal{N}(2, 0.5)$ for the second action in the first context. In this domain, we sample a counterfactual annotation for every sample. Factual rewards are sampled from the reward distribution. We use a uniform initial context distribution, and report results over 100 runs for all estimators. We use a dataset of size $N=100$ to fit the OPE estimate and $M=100$ counterfactual annotations. The same dataset size and distribution is used to learn a reward function estimate if necessary depending on the OPE method. We equally weight all samples (0.5). The reward function is represented as a sample mean. A misspecified reward function uses a partially observed context. In particular, for 50\% of the samples, the context is randomly selected. There are no hyperparameters to be tuned. We report results averaged across 9 possible combinations of $\pi_b, \pi_e$. We use all combinatorial combinations of the policies $[0.1, 0.9], [0.5, 0.5], [0.9, 0.1]$. 
\subsection{Heartsteps}
The Heartsteps simulator is a step count simulator based off earlier work~\citep{mandyam2024adaptive}. All samples are sampled randomly and independently from an initial context distribution $d_0$ where each context is the square root of the prior day's step count. There are two possible actions: send a notification, or do not send a notification. The context and action are projected into $\mathbb{R}^3$ using a function $\phi(s,a)$, which outputs a vector that contains a scalar to represent the eventual decrease in step count over time, the previous day's step count, and a treatment effect term that is nonzero when the action is nonzero. Then, the step count is calculated as $\phi(s,a) \cdot \theta^T$ where $\theta= [-0.04,0.9999,0.3]$. The well-specified reward model is a linear function of $\phi(s,a)$. The misspecified reward model is a linear function of only the first two indices of the vector $\phi(s,a)$. We report results averaged across 9 possible combinations of $\pi_b, \pi_e$. We use all combinatorial combinations of the policies $[0.1, 0.9], [0.5, 0.5], [0.9, 0.1]$. 

\subsection{Sepsis Treatment}
The sepsis treatment simulator is based off prior work~\citep{oberst2019counterfactualoffpolicyevaluationgumbelmax}. While the original simulator assumes a Markov Decision Process (MDP) setting, we adapt the simulator to accommodate a contextual bandit setting. To do this, we sample one-step transitions rather than full trajectories. There are 1442 possible contexts and 8 possible actions in the environment. The initial context distribution $d_0$ samples uniformly across the possible contexts. Each context is represented as a vector of length 8, where the features describe patient heart rate, systolic blood pressure, blood glucose level, percentage oxygen, and the presence of three treatments including antibiotics, vasopressors, and ventilation. The 8 possible actions represent every combinatorial combination of three binary treatments: antibiotics, vasopressors, and ventilation. The well-specified reward function is a linear function of the number of abnormal vitals, and whether the patient is on treatment or not, with $\theta=[-1, -1]$. The misspecified reward model one-hot-encodes the context and action and projects into a vector of length 168. The reward function is then a linear function of the vector of length 168. In this setting, we use one target policy, $\pi_e = [0.3, 0.2, 0, 0, 0.2, 0.1, 0.1, 0.1]$. We report results averaged across the following behavior policies:
\begin{align*}
    \pi_{b1} = [0.1, 0.1, 0.4, 0.3, 0.1, 0.0, 0.0, 0.0] \\
    \pi_{b2} = [0.1, 0.1, 0.4, 0.2, 0.1, 0.1, 0.0, 0.0] \\
    \pi_{b3} = [0.1, 0.1, 0.4, 0.1, 0.1, 0.1, 0.0, 0.1]  \\
    \pi_{b4} = [0.1, 0.1, 0.3, 0.1, 0.1, 0.1, 0.1, 0.1]  \\
    \pi_{b5} = [0.2, 0.1, 0.2, 0.1, 0.1, 0.1, 0.1, 0.1] \\
    \pi_{b6} = [0.3, 0.1, 0.2, 0.0, 0.1, 0.1, 0.1, 0.1] \\
\end{align*}

\subsection{MIMIC-IV}
Here, we use data from MIMIC-IV~\citep{mimicivdataset}, an electronic health records dataset sourced from the Beth Israel Deaconess Medical Center in Boston, MA. We consider a subset of the patients who receive potassium repletion. That is, we include all patients who have received at least one instance of potassium administration through an IV. Furthermore, we treat this setting as a one-step contextual bandit setting. A patient context is represented as a 20-dimensional vector containing information about static covariates (e.g., age, gender), aggregated lab values observed in the previous four hour window, aggregated medicines administered in the previous four hour window, and any indication of procedures that were undertaken (e.g., the patient was placed on a ventilator). There are five possible actions, each corresponding to a dosage of potassium (units are mEq). After administering a dosage of potassium, we observe a reward that is a function of the patient's updated context. 

In our other empirical results, we report RMSE as the key metric; this hinges on the fact that we know the reward function for this setting. To emulate this in MIMIC-IV, a setting in which there is no reported reward signal, we construct a specific reward function. In particular, this reward function is a binary indicator of whether a patient's potassium lab value observed after potassium administration is within the potassium reference range (3.5-5 mmol/L). That is, the reward function $R(s_{t+1})$ is a function of the patient's next observed context. It is reasonable to assume that the reward function is shared across all patients because the potassium reference range is shared across all patients. 

To emulate a setting in which we have a behavior policy and a target policy, we further split the cohort of patients that receive potassium repletion into two sub-cohorts. In particular, one sub-cohort does not have renal disease, and one sub-cohort does have renal disease. The repletion policies are different between these cohorts because the policy must consider the inability for the renal disease patients' kidneys to properly filter potassium. We treat the non-renal population as the behavior cohort and the renal population as the target cohort. The value of the target policy can be calculated by treating the target samples as Monte-Carlo samples. We estimate the policies from the data. 

\begin{figure}[h]
    \centering
    \includegraphics[width=0.4\linewidth]{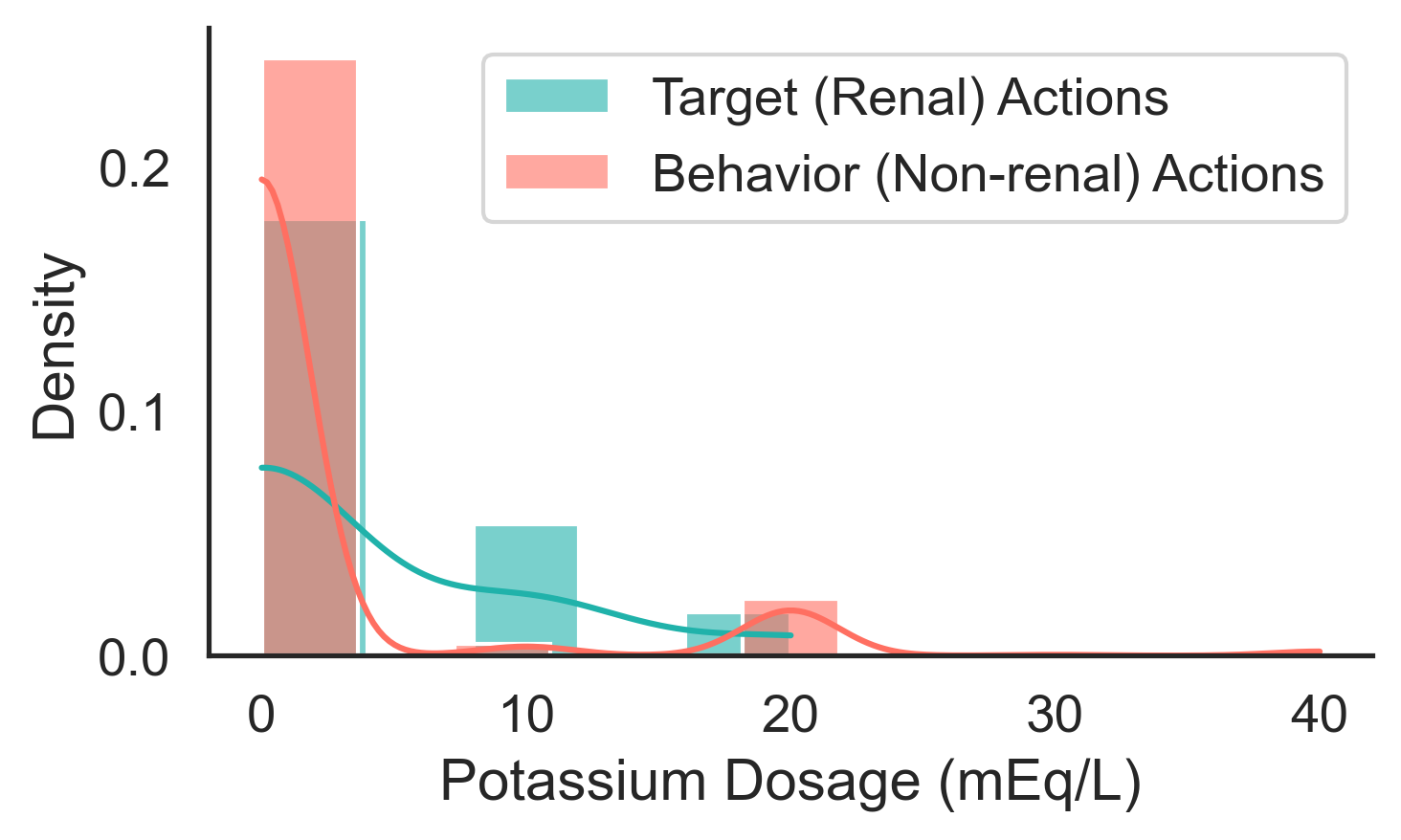}
    \caption{\textbf{Repletion policies differ between the behavior and target cohorts}, making this a setting in which OPE can be helpful. In particular, we note that the target dataset samples actions (e.g., 10 mEq of potassium) that are under-observed in the behavior dataset.}
    \label{apd:potassium_dosages}
\end{figure}

We expect the repletion policies between the two groups to differ since the policy for renal patients must account for the inability of their kidneys to efficiently absorb potassium. 
This setup allows us to calculate the ground-truth value of the target policy using the returns of the target trajectories. As shown in \Cref{apd:potassium_dosages}, patients with renal disease are administered a variety of dosages, including ones that are under-observed in the behavior dataset (e.g., 10 mEq); as a result, this is a setting in which OPE can be useful.
Finally, we construct annotations using LLMs for the behavior cohort. The annotations are sourced from a powerful LLM, OpenAI ‘o1’. LLMs have been successful in reasoning about medical domains, presenting a cheaper alternative to soliciting an expert (i.e., doctor). To obtain annotations, we ask the LLM to complete two tasks. The first is to summarize the patient context, highlighting covariates relevant for predicting blood potassium level. Then, we ask the LLM to predict the patient’s blood potassium if they were administered a given dosage. Note that we have ground truth lab values for the dosages administered in MIMIC-IV; in our experiments we found ‘o1’ predictions to largely align with ground truth lab values. Annotations are obtained by soliciting lab value predictions for unobserved dosages and feeding these values into the known reward function.

\begin{table*}[t]
  \small
  \setlength{\tabcolsep}{4pt} 
  \renewcommand{\arraystretch}{1.2} 
  \begin{tabular}{|m{1.4cm}|m{0.7cm}|m{0.7cm}|m{1.4cm}|m{0.7cm}|m{0.7cm}|m{1.2cm}|m{2.4cm}|m{3.2cm}|}
    \hline
    \rowcolor{Gainsboro!60}
    \textbf{Dataset} & $\textbf{N}$ & \textbf{M} & \makecell{\textbf{\%} \\ \textbf{annotated}} & \makecell{$|\mathcal{A}|$} & \makecell{$|\mathcal{S}|$} & \makecell{\# of \\ $\pi_b$, $\pi_e$} & \makecell{\textbf{Well-specified} \\ \textbf{reward}} & \makecell{\textbf{Misspecified} \\ \textbf{reward}} \\
    \hline
    2-context Bandit & 100 & 100 & 100 & 2 & 2 & 9 & Sample mean & A random state is observed 50\% of the time, creating partial observability like ModelFail in \citet{thomas2016dataefficient}. \\
    \hline
    Heartsteps & 200 & 200 & 100 & 2 & 80 & 9 & Linear regression (3 features) & Linear regression (2 features) \\
    \hline
    Sepsis & 700 & 700 & 12.5 & 8 & 1442 & 6 & Linear regression (top 2 features) & Linear regression (168 least relevant features) \\
    \hline
    MIMIC-IV & 652 & Varies & Varies & 5 & $>$ 13000 & 1 & N/A & Indicator function of whether the lab is within the reference range \\
    \hline
  \end{tabular}
  \caption{Key characteristics of contextual bandits settings used in empirical results.}
  \label{tab:environments}
\end{table*}

\section{Naive Doubly Robust Estimator}
\label{apd:naive_dr}
As discussed in \Cref{sec:intro}, a naive doubly robust estimator uses the dataset $D^+$ within the context of a standard doubly robust estimator. The definition of the naive doubly robust estimator is
$$
\hat{V}^{\text{Naive-DR}} = \sum_{i=1}^{N + M} \Big(\hat{R}^+(s_i, \pi_e) + \rho_{s_i}(a_i) (c_i^{a_i} - \hat{R}^+(s_i, a_i)\Big), 
$$
where $c_i^{a_i}$ is the factual reward or counterfactual annotation depending on if $(s_i, a_i)$ is a factual or counterfactual sample. We claim in the main text that this approach will always lead to an arbitrarily biased estimate of $v(\pi_e)$ because it does not preserve the context distribution in $D$. To demonstrate this, we report results (\Cref{tab:naive_dr}) for the naive doubly robust estimator in the 2-context bandit environment under a well-specified reward model and a perfect annotation setting (\Cref{asm:perfect-annot}). Note that the standard doubly robust estimator ($\DMIS$) is unbiased here, and in contrast, the naive doubly robust estimator is biased. This is because the naive doubly robust estimator alters the context distribution $d_0$, which is assumed to be constant between the behavior dataset and the dataset used to calculate the target policy value $v(\pi_e)$. 

\begin{table}[]
    \centering
    \caption{Naive doubly robust estimator performance. We report mean $\pm$ standard error for all values.}
    \begin{tabular}{|c|c|c|c|}
        \hline
       \textbf{Method}  &  \makecell{\textbf{RMSE}} & \makecell{\textbf{Bias}} & \makecell{\textbf{Std}}  \\ \hline
       Naive DR & 0.317 $\pm$ 0.0007 & 0.3038 $\pm$ 0.0922&0.092 $\pm$ 0.0001 \\ \hline
       $\DMIS$ & 0.108 $\pm$ 0.0002 & -0.016$\pm$ 0.1066 & 0.1067 $\pm$ 0.0001  \\ \hline
    \end{tabular}
    \label{tab:naive_dr}
\end{table}

\section{Table of theoretical results}
We provide a summary for our theoretical results to guide the reader to the appropriate proofs (\Cref{tab:theory}). Earlier work~\citep{tang2023counterfactualaugmented} introduced an IS-based estimator that incorporates counterfactual annotations, and found that re-weighting the samples was required to maintain the context-action distribution defined by the original factual dataset. In our work, we note that this re-weighting is not necessary when building a reward function, which we discuss in ~\Cref{apd:reward_fn}. 
On the journey to constructing a doubly robust estimator, we derive the bias and variance of the standard direct method OPE estimator in ~\Cref{apd:directmethod}. 
Our work identifies three opportunities to incorporate counterfactual annotations into a doubly robust estimator. In this work, we investigate the theoretical properties of these estimators under three annotation settings: perfect annotations, biased annotations, and higher variance annotations. We prove expectation and variance terms for each of our three estimators with and without assumptions on the annotation quality in ~\Cref{apd:bias_variance_dr}. Note that we avoid deriving the variance for $\DMCIS$ and $\CDMCIS$ because these terms are very complicated. Instead, we analyze these variance terms empirically in our simulated settings. Finally, we establish an equivalence between two of our doubly robust approaches and $\CIS$~\citep{tang2023counterfactualaugmented} under an equal weighting scheme in ~\Cref{apd:equivalence}. 
\begin{table}[]
    \centering
    \caption{Summary of theoretical results and associated proofs. Note that only \Cref{asm:biased_annot} is required for the corresponding bias proofs. }
    \begin{tabular}{|c|c|c|c|c|}
        \hline
       \textbf{Method}  &  \makecell{\textbf{Assumption}\\\textbf{Annotation}} & \makecell{\textbf{Assumption}\\\textbf{Coverage}} & \textbf{Bias} & \textbf{Std}  \\ \hline
       $\CDMIS$ & \ref{asm:perfect-annot} & \ref{asm:common-support} & \Cref{thm:CDM-IS_unbiasedness} & \Cref{thm:CDM-IS_variance} \\ \hline
       $\DMCIS$ & \ref{asm:perfect-annot} & \ref{asm:common-support-cf} &  \Cref{thm:DM+C-IS_unbiasedness}& \Cref{thm:DM+C-IS_variance} \\ \hline
       $\CDMCIS$ & \ref{asm:perfect-annot} & \ref{asm:common-support-cf} & \Cref{thm:C-DM+C-IS_unbiasedness} & \Cref{thm:C-DM-C-IS_variance} \\ \hline
       $\CDMIS$ & \ref{asm:biased_annot},\ref{asm:variance_annot} & \ref{asm:common-support} & \Cref{thm:CDM-IS-biased} & \Cref{thm:CDM-IS-imperfect_variance} \\ \hline
       $\DMCIS$ & \ref{asm:biased_annot},\ref{asm:variance_annot} & \ref{asm:common-support-cf} & \Cref{thm:DR_aug_biased_annot} & \textcolor{Brown}{\Cref{prop:dmcis_imperfect_variance}} \\ \hline
       $\CDMCIS$ & \ref{asm:biased_annot},\ref{asm:variance_annot} & \ref{asm:common-support-cf} & \Cref{thm:DR_aug_biased_annot} & \textcolor{Brown}{\Cref{cdmcis_variance_imperfect_annot}}  \\ \hline
    \end{tabular}
    \label{tab:theory}
\end{table}

\section{Weighted vs. unweighted reward function}
\label{apd:reward_fn}
In the main text, we claim that we do not need to re-weight samples when constructing a reward function to produce an unbiased estimate, like earlier work had to do with $\CIS$~\citep{tang2023counterfactualaugmented}. To demonstrate this, we first derive the expectation and variance of the weighted reward function, and identify that the unweighted (or equally weighted) reward function is a special case. We then compare the variance terms for the unweighted and weighted reward functions, and prove that the variance of the unweighted reward function is lower. This result suggests that not weighting the samples when constructing the reward function is a superior strategy in the case that the annotations are perfect. 
\subsection{Weighted Reward Function}
The weighted reward function is 
$$
\hat{R}^+(s,a) = \frac{\sum_{i=1}^N \mathbbm{1} (s_i = s, a_i = a) * w_i^{a} * r_i + \sum_{j=1}^M \mathbbm{1} (s_j = s, a_j = a) * w_j^{a} * g_j}{\sum_{i=1}^N \mathbbm{1} (s_i = s, a_i = a) * w_i^{a} + \sum_{j=1}^M \mathbbm{1} (s_j = s, a_j = a) * w_j^{a}}
$$
where $w_i^{a} \sim W(s_i, a_i)$ is a known weight associated with the sample $s_i, a_i$ that arises from some function $W$. 
\subsubsection{Bias}
\label{apd:weighted_reward_fn_bias}
\begin{proposition}[name=Unbiasedness of weighted reward function] \label{prop:unbiased_weighted_reward_fn}
Under \Cref{asm:perfect-annot}, the weighted reward function is unbiased. $\mathbb{E}[\hat{R}^+(s,a)] = \bar{R}(s,a)$. 
\end{proposition}
Proof: Let $N(s,a) = \sum_{i=1}^{M+N} \mathbf{1}(s_i=s, a_i=a)$ be the number of times the sample $s, a$ appears in the counterfactual-annotated dataset $D^+$. We can re-write the weighted reward function as a function of $N(s,a)$. Let $c_i=r_i$ if the sample is in the factual dataset, and $c_i=g_i$ if the sample is a counterfactual annotation. 
\begin{align*}
    \hat{R}^+(s,a) &= \frac{1}{\sum_{i=1}^{N(s,a)} w_i^a} \sum_{i=1}^{N(s,a)} w_i^a c_i 
\end{align*}
To calculate expectation, we first use the law of total expectation and the expectation of the scalar reward or annotation. 
\begin{align}
    &\mathbb{E}_{D^+ \sim \mathcal{D^+}} [\hat{R}^+(s,a)] = \mathbb{E}_{\substack{N_{s,a} \sim D^+, c_i \sim R(s_i,a_i),\\ w_i^a \sim W(s_i,a)}} \left[ \frac{1}{\sum_{i=1}^{N(s,a)} w_i^a} \sum_{i=1}^{N(s,a)} w_i^a c_i \right] \\
    &= \mathbb{E}_{\substack{N_{s,a} \sim D^+}} \left[ \mathbb{E}_{c_i \sim R(s_i,a_i), w_i^a \sim W(s_i,a)} \left[\frac{1}{\sum_{i=1}^{N(s,a)} w_i^a} \sum_{i=1}^{N(s,a)} w_i^a c_i \right]\right] \\
    &= \mathbb{E}_{\substack{N_{s,a} \sim D^+}} \left[ \mathbb{E}_{w_i^a \sim W(s_i,a)} \left[\frac{1}{\sum_{i=1}^{N(s,a)} w_i^a} \sum_{i=1}^{N(s,a)} w_i^a \mathbb{E}_{c_i \sim R(s_i,a_i)}[c_i] \right]\right] \\
    &= \mathbb{E}_{\substack{N_{s,a} \sim D^+}} \left[ \mathbb{E}_{w_i^a \sim W(s_i,a)} \left[\frac{1}{\sum_{i=1}^{N(s,a)} w_i^a} \sum_{i=1}^{N(s,a)} w_i^a \bar{R}(s,a)\right]\right] 
\end{align}
Let $W_i^{a} = \frac{w_i^a}{\sum_{i=1}^{N(s,a)} w_i^a}$. Because each $w_i^a$ is normalized by the sum of all the weights across all the samples, the sum of all weights is 1. Now, we can write the sum as a weighted mean:
\begin{align}
    &= \mathbb{E}_{\substack{N_{s,a} \sim D^+}} \left[ \mathbb{E}_{w_i^a \sim W(s_i,a)} \left[\sum_{i=1}^{N(s,a)} W_i^{a} \bar{R}(s,a)\right]\right] \\
\end{align}
Because $\bar{R}(s_i,a_i)$ is a constant, we can pull it out of the expectation. 
\begin{align}
    &= \mathbb{E}_{\substack{N_{s,a} \sim D^+}} \left[ \mathbb{E}_{w_i^a \sim W(s_i,a)} \left[\bar{R}(s,a) \sum_{i=1}^{N(s,a)} W_i^{a} \right]\right] \\
    &= \mathbb{E}_{\substack{N_{s,a} \sim D^+}} \left[ \mathbb{E}_{w_i^a \sim W(s_i,a)} \left[\bar{R}(s,a) \right]\right] \\
    &= \bar{R}(s,a) 
\end{align}
\subsubsection{Variance}
\label{apd:weighted_reward_fn_variance}
\begin{proposition}
\label{prop:variance_weighted_reward_fn}
Under \Cref{asm:perfect-annot}, the weighted reward function has variance, $\mathbb{V}[\hat{R}^+(s,a)] = \mathbb{E}_{\substack{N_{s,a} \sim D^+}}\left[\sum_{i=1}^{N(s,a)} (W_i^{a})^2 \sigma_R^2(s,a) \right]$. 
\end{proposition}
Proof: We use the law of total variance and the definition of $W_i^{a}$.
\begin{align}
    &\mathbb{V}_{D^+ \sim \mathcal{D^+}}[\hat{R}^+(s,a)] = \mathbb{V}_{\substack{N_{s,a} \sim D^+,\\ c_i \sim R(s_i,a_i)}}\left[\sum_{i=1}^{N(s,a)} \frac{w_i^a c_i}{\sum_{i=1}^{N(s,a)} w_i^a} \right] \\
    &=  \mathbb{E}_{\substack{N_{s,a} \sim D^+}}\left[\mathbb{V}_{c_i \sim R(s_i,a_i)}\left[\sum_{i=1}^{N(s,a)} \frac{w_i^a c_i}{\sum_{i=1}^{N(s,a)} w_i^a} \right]\right] \\
     &+ \mathbb{V}_{\substack{N_{s,a} \sim D^+}}\left[ \mathbb{E}_{c_i \sim R(s_i,a_i)}\left[\sum_{i=1}^{N(s,a)} \frac{w_i^a c_i}{\sum_{i=1}^{N(s,a)} w_i^a} \right]\right]\\
    &=  \mathbb{E}_{\substack{N_{s,a} \sim D^+}}\left[\mathbb{V}_{c_i \sim R(s_i,a_i)}\left[\sum_{i=1}^{N(s,a)} \frac{w_i^a c_i}{\sum_{i=1}^{N(s,a)} w_i^a} \right]\right] \\
    &=  \mathbb{E}_{\substack{N_{s,a} \sim D^+}}\left[\mathbb{V}_{c_i \sim R(s_i,a_i)}\left[\sum_{i=1}^{N(s,a)} W_i^{a} c_i \right]\right] \\
    &=  \mathbb{E}_{\substack{N_{s,a} \sim D^+}}\left[\sum_{i=1}^{N(s,a)} (W_i^{a})^2 \mathbb{V}_{c_i \sim R(s,a)}\left[c_i \right]\right] \\
    &=  \mathbb{E}_{\substack{N_{s,a} \sim D^+}}\left[\sum_{i=1}^{N(s,a)} (W_i^{a})^2 \sigma_R^2(s,a) \right] \\
\end{align}

\subsection{Unweighted Reward Function}
Here the unweighted reward function is identical to the weighted one, except without weights. This can also be seen as a special case of the weighted reward function with equal weights. The unweighted reward function is :  
\begin{align}
    \hat{R}^+(s,a) = \frac{\sum_{i=1}^{N} \mathbbm{1} (s_i = s, a_i = a) * r_i + \sum_{j=1}^M \mathbbm{1} (s_j = s, a_j = a) * g_j}{\sum_{i=1}^N \mathbbm{1} (s_i = s, a_i = a) + \sum_{j=1}^M \mathbbm{1} (s_j = s, a_j = a)}
\end{align}
\subsubsection{Bias}
\label{apd:biased_rewardfn}
\begin{proposition}
\label{biased_reward_fn_proof}
    Under \Cref{asm:biased_annot}, the unweighted reward function has expectation $\mathbb{E}[\hat{R}^+(s,a)] = \mathbb{E}_{N_{s,a}, M_{s,a}} \left[ \bar{R}(s_i, a_i) + \frac{M_{s,a} \epsilon_G}{N_{s,a} + M_{s,a}} \right]$ where $N_{s,a}, M_{s,a}$ are the number of factual and counterfactual samples with context-action equivalent to $s,a$ in the counterfactual augmented dataset. 
\end{proposition}
Proof: 
Define $N_{s,a} = \sum_{i=1}^N \mathbbm{1}(s_i=s, a_i=a)$ and $M_{s,a} = \sum_{j=1}^M \mathbbm{1}(s_j=s, a_j=a)$ where $N$ is the number of total factual samples and $M$ is the total number of counterfactual annotations. 

Now, we can re-write the reward function as $\hat{R}^+(s,a) = \frac{\sum_{i=1}^{N_{s,a}} r_i + \sum_{j=1}^{M_{s,a}} g_j}{N_{s,a} + M_{s,a}}$.

We calculate the expectation of this reward function under the dataset distribution, which is the joint distribution across $N_{s,a} \sim D^+, M_{s,a}\sim D^+, r \sim R(s,a), g \sim G(s,a)$. 
\begin{align}
        \mathbb{E}_{D^+ \sim \mathcal{D^+}}[\hat{R}^+(s,a)] &= \mathbb{E}_{N_{s,a}, M_{s,a} \sim D^+, r \sim R(s,a), g \sim G(s,a)} \left[ \frac{\sum_{i=1}^{N_{s,a}} r_i + \sum_{j=1}^{M_{s,a}} g_j}{N_{s,a} + M_{s,a}}\right]
\end{align}
We can use the law of total expectation to separate out the joint distribution:
\begin{align}
    &= \mathbb{E}_{\substack{N_{s,a}\\ \sim D^+}} \left[ \mathbb{E}_{M_{s,a} \sim D^+, r \sim R(s,a), g \sim G(s,a)}\left[\frac{\sum_{i=1}^{N_{s,a}} r_i + \sum_{j=1}^{M_{s,a}} g_j}{N_{s,a} + M_{s,a}}\right]\right] \\
    &= \mathbb{E}_{\substack{N_{s,a}\\ \sim D^+}} \left[ \mathbb{E}_{M_{s,a} \sim D^+}\left[\mathbb{E}_{r \sim R(s,a), g \sim G(s,a)}\left[\frac{\sum_{i=1}^{N_{s,a}} r_i + \sum_{j=1}^{M_{s,a}} g_j}{N_{s,a} + M_{s,a}} \right]\right]\right] \\
    &= \mathbb{E}_{\substack{N_{s,a}\\ \sim D^+}} \left[ \mathbb{E}_{M_{s,a} \sim D^+}\left[\frac{1}{N_{s,a} + M_{s,a}}\sum_{i=1}^{N_{s,a}} \bar{R}(s_i, a_i) + \sum_{j=1}^{M_{s,a}} \bar{R}(s_j, a_j) + \epsilon_G \right]\right] \\
    &= \mathbb{E}_{N_{s,a} \sim D^+} \left[ \mathbb{E}_{M_{s,a} \sim D^+}\left[\frac{1}{N_{s,a} + M_{s,a}}\sum_{i=1}^{N_{s,a}} \bar{R}(s_i, a_i) + \sum_{j=1}^{M_{s,a}} \bar{R}(s_j, a_j) + \epsilon_G \right]\right] \\
    &= \mathbb{E}_{N_{s,a} \sim D^+} \left[ \mathbb{E}_{M_{s,a} \sim D^+}\left[\frac{1}{N_{s,a} + M_{s,a}} N_{s,a} \times \bar{R}(s_i, a_i) + M_{s,a} \times ( \bar{R}(s_j, a_j) + \epsilon_G) \right]\right] \\
    &= \mathbb{E}_{N_{s,a} \sim D^+} \left[ \mathbb{E}_{M_{s,a} \sim D^+}\left[\bar{R}(s_i, a_i) + \frac{M_{s,a} \epsilon_G}{N_{s,a} + M_{s,a}} \right]\right] \\
    &= \mathbb{E}_{N_{s,a}, M_{s,a} \sim D^+} \left[ \bar{R}(s_i, a_i) + \frac{M_{s,a} \epsilon_G}{N_{s,a} + M_{s,a}} \right] \\
\end{align}
\begin{proposition}
\label{prop:unbiased_unweighted_reward_fn}
Under \Cref{asm:perfect-annot},
the unweighted reward function is unbiased, $\mathbb{E}[\hat{R}^+(s,a)] = \bar{R}(s,a)$. 
\end{proposition}
Proof: If we set $\epsilon_G=0$ in \Cref{biased_reward_fn_proof}, the expected value of the unweighted reward function is $\bar{R}(s,a)$, which means that it is an unbiased estimator. 
\subsubsection{Variance}
\begin{proposition}
\label{prop:unbiased_unweighted_reward_fn_var}
Under \Cref{asm:perfect-annot},
the unweighted reward function has variance, $\mathbb{V}[\hat{R}^+(s,a)] = \sigma_R^2(s,a) \mathbb{E}_{N_{s,a}}\left[\frac{1}{N_{s,a}}\right]$. 
\end{proposition}
Proof: 
Let $N_{s,a}$ denote how many samples in the total dataset have the same state-action as $s,a$. We now use the law of total variance, the earlier bias result, and the expectation of $c_i$. 
\begin{align}
    \mathbb{V}_{D^+ \sim \mathcal{D^+}}[\hat{R}^+(s,a)] &= \mathbb{V}_{\substack{N_{s,a}\\ c_i \sim R(s,a)}}\left[
       \frac{1}{N_{s,a}} \sum_{i=1}^{N_{s,a}} c_i \right] \\
    &=  \mathbb{E}_{\substack{N_{s,a}}}\left[\mathbb{V}_{c \sim R}\left[\frac{1}{N_{s,a}} \sum_{i=1}^{N_{s,a}} c_i \right]\right] \\
     &+ \mathbb{V}_{\substack{N_{s,a}}}\left[ \mathbb{E}_{c
    _i \sim R(s,a)}\left[\frac{1}{N_{s,a}}  \sum_{i=1}^{N_{s,a}} c_i \right] \right]\\
     &= \mathbb{E}_{N_{s,a}} \left[\mathbb{V}_{c_i \sim R(s,a)} \left[\frac{1}{N_{s,a}} \sum_{i=1}^{N_{s,a}} c_i \right] \right]\\
     &= \mathbb{E}_{\substack{N_{s,a}}} \left[ \frac{1}{{N_{s,a}}^2} \sum_{i=1}^{N_{s,a}}\mathbb{V}_{c_i \sim R(s,a)} \left[ c_i \right] \right] \\
        &= \mathbb{E}_{\substack{N_{s,a}}} \left[ \frac{1}{{N_{s,a}}^2} \sum_{i=1}^{N_{s,a}} \sigma^2_R(s, a) \right] \\
        &= \mathbb{E}_{\substack{N_{s,a}}} \left[ \frac{1}{{N_{s,a}}^2} \times N_{s,a} \times \sigma^2_R(s, a) \right] \\
        &= \mathbb{E}_{\substack{N_{s,a}}} \left[ \frac{\sigma^2_R(s,a)}{N_{s,a}} \right] \\
        &= \sigma^2_R(s,a) \mathbb{E}_{\substack{N_{s,a}}} \left[ \frac{1}{N_{s,a}} \right] \\
\end{align}
\subsection{Comparing variance terms}
\begin{proposition}
\label{prop:comparing_var_rewardfn}
The variance of the unweighted reward function is less than or equal to the variance of the weighted reward function. 
\end{proposition}
Proof: We can prove that the variance of the weighted reward function is higher than that of the unweighted reward function using Jensen's inequality. Since both the variance terms for the weighted and unweighted reward function have a $\sigma_R^2(s,a)$, we consider only the coefficient. We start with the coefficient for the weighted reward function on the LHS of the first line. We know by Jensen's inequality that:
\begin{align}
    \frac{1}{N_{s,a}} \sum_{i=1}^{N_{s,a}} (W_i^{a})^2 &\geq \frac{1}{N_{s,a}^2} \left( \sum_{i=1}^{N_{s,a}} W_i^{a} \right)^2 \\
    \frac{1}{N_{s,a}} \sum_{i=1}^{N_{s,a}} (W_i^{a})^2 &\geq \frac{1}{N_{s,a}^2} \times 1 \\
     \sum_{i=1}^{N_{s,a}} (W_i^{a})^2 &\geq \frac{1}{N_{s,a}}
\end{align}
Note that the coefficient of the variance term for the unweighted reward function is $\frac{1}{N_{s,a}}$. This result suggests that the variance of the randomly weighted reward function is always at least as high as the variance of the uniformly weighted reward function. As a result, we recommend using a uniformly weighted (or unweighted) reward function estimator when we have perfect annotations.

In the case that we know the annotations are imperfect, we recommend using a weighted reward function that down-weights the imperfect annotation (and consequently up-weights the corresponding factual sample). We hypothesize that this procedure will result in an improved estimator in both the misspecified and well-specified reward function scenarios because it limits the effect of the imperfect annotation. 

We include a baseline which augments the standard DM estimator using counterfactual annotations. This estimator is $\hat{V}^{DM^+} = \sum_{s} d_0(s) \sum_{a} \pi(a|s) \hat{R}^+(s,a)$.

\section{Bias and variance of the standard direct method (DM) OPE estimator}
\label{apd:directmethod}
In the main text, we reference the variance term for the standard direct method (DM) estimator. Here, we derive the expectation and variance of the DM estimator, which is defined as:
\begin{align}
    \hat{V}^{\textrm{DM}} = \frac{1}{N}\sum_{s_i \in D_{\hat{R}}} \sum_{a \in A} \pi_e(a|s_i) \hat{R}(s_i,a)
\end{align}
For the purposes of the expectation and variance derivations below, we assume that the reward model is fully realizable according to \Cref{asm:realizability}.
\begin{assumption}[Realizability] \label{asm:realizability}
$R^* \in \mathcal{F}$. 
\end{assumption}
We derive the expectation and variance with respect to two datasets. $D_0$ is used to estimate the OPE value, and $D_{\hat{R}}$ is used to estimate the reward function.

\begin{proposition}[Unbiasedness of $\hat{V}^{\text{DM}}$] \label{thm:Vhat-exp}
Under \Cref{asm:realizability}, $\mathbb{E}_{D_{0} \sim \mathcal{D}, D_{\hat{R}} \sim \mathcal{D}}[\hat{V}^{\text{DM}}] = v(\pi_e)$ if $N_{s,a} > 0$ for all $(s,a)$ where $d(s) > 0$ and $\pi_e(a|s) > 0$ in the dataset $D_{\hat{R}}$. 
\end{proposition}
Proof:
\begin{align}
\mathbb{E}_{D_0 \sim\mathcal{D}, D_{\hat{R}} \sim \mathcal{D}}[\hat{V}^{\text{DM}}] &= \mathbb{E}_{D_0, D_{\hat{R}} \sim\mathcal{D}} \left[ \frac{1}{N}\sum_{s_i\in D_{0}} \sum_{a \in A} \pi_e(a|s_i) \hat{R}(s_i,a) \right] \\
&= \mathbb{E}_{D_{\hat{R}} \sim \mathcal{D}} \left[ \frac{1}{N} \sum_{i=1}^{N} \sum_{a \in A} \pi_e(a|s_i) \mathbb{E}_{D_{\hat{R}} \sim \mathcal{D}} [\hat{R}(s_i,a) ] \right] \\
&= \frac{1}{N} \sum_{i=1}^{N} \mathbb{E}_{D_{\hat{R}} \sim\mathcal{D}} \left[ \sum_{a \in A} \pi_e(a|s_i) \mathbb{E}_{D_{\hat{R}} \sim \mathcal{D}} [\hat{R}(s_i,a) ] \right] \\
\end{align}
Now, because we consider every sample as independent, we consider just the term inside the summation without $\frac{1}{N}$.
\begin{align}
&= \sum_{i=1}^{N} \sum_{s \in \mathcal{S}} d(s) \left( \sum_{a \in A} \pi_e(a|s) \mathbb{E}_{D_{\hat{R}} \sim \mathcal{D}} [\hat{R}(s,a) ] \right) \\
&= \sum_{i=1}^{N} \sum_{s \in \mathcal{S}} d(s) \left( \sum_{a \in A} \pi_e(a|s) \bar{R}(s,a) \right) \\
&= \sum_{i=1}^{N} \sum_{s \in \mathcal{S}} d(s) v^{\pi_e}(s_i) \\
&= \sum_{i=1}^{N} v(\pi_e) \\
&= v(\pi_e)
\end{align}
where we apply linearity of expectation, definition of expectation over $D_{\hat{R}}$, substitute the expectation of $\hat{R}(s,a)$ where $d(s) > 0$ and $\pi_e(a|s) > 0$, definition of value function and policy value. 

\begin{proposition}[Variance of $\hat{V}^{\text{DM}}$] \label{thm:Vhat-var}
Under \Cref{asm:realizability}, 
$\mathbb{V}_{D \sim \mathcal{D}, D_{\hat{R}} \sim \mathcal{D}}[\hat{V}^{\text{DM}}] = \frac{1}{N}\mathbb{V}_{s \sim d_0} \left[ {V^{\pi_e}(s)} \right] + \mathbb{E}_{s \sim d_0}\mathbb{E}_{a \sim \pi_e} \left[\Big(\frac{1}{N} + \big(1-\frac{1}{N}\big)d(s) \Big) \, \pi_e(a|s) \, \sigma_R^2(s,a) \, \mathbb{E}_{D_{\hat{R}} \sim \mathcal{D}}\Big[\frac{1}{N_{s,a}(D_{\hat{R}})}\Big] \right]$ . 
\end{proposition}
Proof:
By law of total variance, we can decompose the variance into two terms:
\begin{align}
& \mathbb{V}_{D_{\hat{R}} \sim\mathcal{D}, D_{\hat{R}} \sim \mathcal{D}}[\hat{V}^{\text{DM}}] = \underbrace{\mathbb{V}_{D_{\hat{R}} \sim\mathcal{D}} \mathbb{E}_{D_{\hat{R}} \sim \mathcal{D}} \left[\hat{V}^{\text{DM}}\right]}_{(1)} + \underbrace{\mathbb{E}_{D_{\hat{R}} \sim\mathcal{D}} \mathbb{V}_{D_{\hat{R}} \sim \mathcal{D}} \left[ \hat{V}^{\text{DM}} \right]}_{(1')}
\end{align}

We can substitute the intermediate result from the proof of \Cref{thm:Vhat-exp} into $(1)$:
\begin{align}
(1) &= \mathbb{V}_{D_{\hat{R}} \sim\mathcal{D}} \left[ \mathbb{E}_{D_{\hat{R}} \sim \mathcal{D}}\big[ \hat{V}^{\text{DM}} \big] \right] = \mathbb{V}_{D_{\hat{R}} \sim\mathcal{D}} \left[ \hat{R}(d,\pi_e) \right]
\end{align}

For $(1')$, we first consider the inner variance with respect to $D_{\hat{R}}$ assuming $\hat{R}$ is given:
\begin{align}
\mathbb{V}_{D_{\hat{R}} \sim \mathcal{D}} \left[ \hat{V}^{\text{DM}} \right] &= \mathbb{V}_{D_{\hat{R}} \sim \mathcal{D}} \left[ \frac{1}{N} \sum_{i=1}^{N} \sum_{a \in A} \pi_e(a|s_i) \hat{R}(s_i,a) \right] \\
&= \frac{1}{{N}^2} \sum_{i=1}^{N} \mathbb{V}_{s_i \sim d_0} \left[ \hat{R}(s_i,\pi_e) \right] && \proofcomment{var of sum of iid} \\
&= \frac{1}{N} \mathbb{V}_{s \sim d_0} \left[ \hat{R}(s,\pi_e) \right] && \proofcomment{iid sample average} 
\end{align}

Substituting this into $(1')$:
\begin{align}
& (1') = \mathbb{E}_{D_{\hat{R}} \sim\mathcal{D}} \mathbb{V}_{D_{\hat{R}} \sim \mathcal{D}} \left[ \hat{V}^{\text{DM}} \right] \\
&= \mathbb{E}_{D_{\hat{R}} \sim\mathcal{D}} \left[ \frac{1}{N} \mathbb{V}_{s \sim d_0} \left[ \hat{R}(s,\pi_e) \right] \right] \\
&= \frac{1}{N} \mathbb{E}_{D_{\hat{R}} \sim\mathcal{D}} \left[ \mathbb{E}_{s \sim d_0} \left[ \hat{R}(s,\pi_e)^2 \right] - \mathbb{E}_{s \sim d_0} \left[ \hat{R}(s,\pi_e) \right]^2 \right]  && \proofcomment{definition of var} \\
&= \frac{1}{N} \Bigg( \underbrace{\mathbb{E}_{D_{\hat{R}} \sim\mathcal{D}} \mathbb{E}_{s \sim d_0} \left[ \hat{R}(s,\pi_e)^2 \right]}_{(2)} - \underbrace{\mathbb{E}_{D_{\hat{R}} \sim\mathcal{D}}\left[\mathbb{E}_{s \sim d_0} \left[ \hat{R}(s,\pi_e) \right]^2 \right]}_{(2')} \Bigg)
\end{align}

\begin{align}
(2) &= \mathbb{E}_{D_{\hat{R}} \sim\mathcal{D}} \mathbb{E}_{s \sim d_0} \left[ \hat{R}(s,\pi_e)^2 \right] \\
&= \mathbb{E}_{s \sim d_0} \mathbb{E}_{D_{\hat{R}} \sim\mathcal{D}} \left[ \hat{R}(s,\pi_e)^2 \right] \\
&= \mathbb{E}_{s \sim d_0} \left[ v^{\pi_e}(s)^2 + \mathbb{V}_{D_{\hat{R}}\sim\mathcal{D}}[\hat{R}(s,\pi_e)] \right] && \proofcomment{substitute corollary} \\
&= \mathbb{E}_{s \sim d_0} \left[ {v^{\pi_e}(s)}^2 \right] + \mathbb{E}_{s \sim d_0} \left[ \mathbb{V}_{D_{\hat{R}} \sim\mathcal{D}} [ \hat{R}(s,\pi_e) ] \right] && \proofcomment{linearity of expectation} \\
&= \mathbb{E}_{s \sim d_0} \left[ {v^{\pi_e}(s)} \right]^2 + \mathbb{V}_{s \sim d_0} \left[ {v^{\pi_e}(s)} \right] + \mathbb{E}_{s \sim d_0} \left[ \mathbb{V}_{D_{\hat{R}} \sim\mathcal{D}} [ \hat{R}(s,\pi_e) ] \right] && \proofcomment{definition of variance} \\
&= {v(\pi_e)}^2 + \mathbb{V}_{s \sim d_0} \left[ {v^{\pi_e}(s)} \right] + \mathbb{E}_{s \sim d_0} \left[ \mathbb{V}_{D_{\hat{R}} \sim\mathcal{D}} [ \hat{R}(s,\pi_e) ] \right] && \proofcomment{definition of value function} \\
\\
(2') &= \mathbb{E}_{D_{\hat{R}} \sim\mathcal{D}}\left[\mathbb{E}_{s \sim d_0} \left[ \hat{R}(s,\pi_e) \right]^2 \right] \\
&= \mathbb{E}_{D_{\hat{R}} \sim\mathcal{D}}\left[\hat{R}(d,\pi_e)^2 \right] \\
&= v(\pi_e)^2 + \mathbb{V}_{D_{\hat{R}}\sim\mathcal{D}}[\hat{R}(d,\pi_e)]
\end{align}

Thus,
\begin{align}
(2) - (2') &= \mathbb{V}_{s \sim d_0} \left[ {v^{\pi_e}(s)} \right] + \mathbb{E}_{s \sim d_0} \left[ \mathbb{V}_{D_{\hat{R}} \sim\mathcal{D}} [ \hat{R}(s,\pi_e) ] \right] - \mathbb{V}_{D_{\hat{R}} \sim\mathcal{D}}\left[\hat{R}(d,\pi_e) \right]
\end{align}

Putting everything together, we have:
\begin{align}
& \mathbb{V}_{D_{\hat{R}} \sim\mathcal{D}, D_{\hat{R}} \sim \mathcal{D}}[\hat{V}^{\text{DM}}] = (1) + (1') \\
&= \mathbb{V}_{D_{\hat{R}} \sim\mathcal{D}} \left[ \hat{R}(d,\pi_e) \right] \\
&+ \frac{1}{N} \left(\mathbb{V}_{s \sim d_0} \left[ {v^{\pi_e}(s)} \right] + \mathbb{E}_{s \sim d_0} \left[ \mathbb{V}_{D_{\hat{R}} \sim\mathcal{D}} [ \hat{R}(s,\pi_e) ] \right] - \mathbb{V}_{D_{\hat{R}} \sim\mathcal{D}}\left[\hat{R}(d,\pi_e) \right]\right) \\
&= \frac{1}{N}\mathbb{V}_{s \sim d_0} \left[ {v^{\pi_e}(s)} \right] + \frac{1}{N}\mathbb{E}_{s \sim d_0} \left[ \mathbb{V}_{D_{\hat{R}} \sim\mathcal{D}} [ \hat{R}(s,\pi_e) ] \right] + \big(1-\frac{1}{N}\big) \mathbb{V}_{D_{\hat{R}} \sim\mathcal{D}}\left[\hat{R}(d,\pi_e) \right] \\
&= \frac{1}{N}\mathbb{V}_{s \sim d_0} \left[ {v^{\pi_e}(s)} \right] 
+ \frac{1}{N}\mathbb{E}_{s \sim d_0} \bigg[ \mathbb{E}_{a \sim \pi_e(\cdot|s)} \Big[\pi_e(a|s) \, \sigma_R^2(s,a) \, \mathbb{E}_{D_{\hat{R}} \sim \mathcal{D}}\big[\frac{1}{N_{s,a}(D_{\hat{R}})}\big] \Big] \bigg] \\
& \qquad \qquad \qquad \qquad \;\; + \big(1-\frac{1}{N}\big) \mathbb{E}_{s \sim d_0}\mathbb{E}_{a \sim \pi_e(\cdot|s)} \left[d(s) \, \pi_e(a|s) \, \sigma_R^2(s,a) \, \mathbb{E}_{D_{\hat{R}} \sim \mathcal{D}}\Big[\frac{1}{N_{s,a}(D_{\hat{R}})}\Big] \right] \\
&= \frac{1}{N}\mathbb{V}_{s \sim d_0} \left[ {v^{\pi_e}(s)} \right] \\
&+ \mathbb{E}_{s \sim d_0}\mathbb{E}_{a \sim \pi_e(\cdot|s)} \left[\Big(\frac{1}{N} + \big(1-\frac{1}{N}\big)d(s) \Big) \, \pi_e(a|s) \, \sigma_R^2(s,a) \, \mathbb{E}_{D_{\hat{R}} \sim \mathcal{D}}\Big[\frac{1}{N_{s,a}(D_{\hat{R}})}\Big] \right]
\end{align}

\section{Expectation and variance of doubly robust estimators under different annotation conditions}
\label{apd:bias_variance_dr}
Now we derive the expectation and variance of the introduced doubly robust estimators under different annotation conditions. In the main text, we note that the annotations can be perfect (i.e. $\mathbb{E}[g_i] = \mathbb{E}[r_i], \mathbb{V}[g_i] = \mathbb{V}[r_i]$), biased (i.e. $\mathbb{E}[g_i] = \mathbb{E}[r_i] + \epsilon_G(s_i, a_i)$), and have higher variance (i.e. $\mathbb{V}[g_i] = \mathbb{V}[r_i] + \Delta_G(s_i, a_i)$). In our proofs, we first derive the expectation and variance under the imperfect annotations condition. Then, we show that the perfect annotation condition is a special case of these derivations. 

\subsection{Expectation and variance of \texorpdfstring{\CDMIS}{DM⁺-IS}}
The DR estimator is defined as
\begin{align*}
     \hat{V}^{DM^+-IS} &= \frac{1}{N} \sum_{i=1}^N \left(\hat{R}^+(s_i, \pi_e) + 
\frac{\pi_e(a_i|s_i)}{\pi_{b}(a_i|s_i)}(r_i - \hat{R}^+(s_i,a_i))\right)
\end{align*}
\subsubsection{Expectation}
We now prove \Cref{thm:CDM-IS-biased}, which is restated below. \\
\begin{proposition*}[name=Expectation of $\CDMIS$ under imperfect annotations]
Under \Cref{asm:biased_annot,asm:common-support},  $\mathbb{E}[\hat{V}^{\CDMIS}] = v(\pi_e)$.
\end{proposition*}

Proof:
The expectation is taken over the dataset $D_0$, which is used to fit the OPE estimate, and $D_{\hat{R}}$, which is used to learn the reward function estimate.  We first use the linearity of expectation. 
\begin{align}
    &\mathbb{E}_{D_0, D_{\hat{R}} \sim \mathcal{D}}[\hat{V}^{DM^+-IS}] \\
    &= \mathbb{E}_{\substack{D_{\hat{R}}, s \sim d_0 \\a \sim \pi_b(\cdot|s), r \sim R(s,a)}} \left[ \frac{1}{N} \sum_{i=1}^N \left(\hat{R}^+(s_i, \pi_e) + 
\frac{\pi_e(a_i|s_i)}{\pi_{b}(a_i|s_i)}(r_i - \hat{R}^+(s_i,a_i))\right) \right] \\
&=  \frac{1}{N} \sum_{i=1}^N \mathbb{E}_{\substack{D_{\hat{R}}, s \sim d_0\\ a \sim \pi_b(\cdot|s), r \sim R(s,a)}} \left[ \left(\hat{R}^+(s_i, \pi_e) + 
\frac{\pi_e(a_i|s_i)}{\pi_{b}(a_i|s_i)}(r_i - \hat{R}^+(s_i,a_i))\right) \right]
\end{align}
We now split the expectation into two terms:
\begin{align}
    &=  \frac{1}{N} \sum_{i=1}^N  \mathbb{E}_{D_{\hat{R}}, s \sim d_0}[\hat{R}^+(s_i, \pi_e)] + 
\mathbb{E}_{\substack{D_{\hat{R}}, s \sim d_0\\ a \sim \pi_b(\cdot|s), r \sim R(s,a)}} \left[\frac{\pi_e(a_i|s_i)}{\pi_{b}(a_i|s_i)}(r_i - \hat{R}^+(s_i,a_i)) \right] \\
&=  \frac{1}{N} \sum_{i=1}^N  \mathbb{E}_{D_{\hat{R}}, s \sim d_0}[\hat{R}^+(s_i, \pi_e)] + 
\mathbb{E}_{\substack{D_{\hat{R}}, s \sim d_0\\ r \sim R(s,a)}} \left[ \pi_b(a_i | s_i) \frac{\pi_e(a_i|s_i)}{\pi_{b}(a_i|s_i)}(r_i - \hat{R}^+(s_i,a_i)) \right] \\
&=  \frac{1}{N} \sum_{i=1}^N  \mathbb{E}_{s \sim d_0}[\hat{R}^+(s_i, \pi_e)] + 
\mathbb{E}_{s \sim d_0 a \sim \pi_e(\cdot|s), r \sim R(s,a)} \left[(r_i - \hat{R}^+(s_i,a_i)) \right] \\
&=  \frac{1}{N} \sum_{i=1}^N \mathbb{E}_{D_{\hat{R}}, s \sim d_0}[\hat{R}^+(s_i, \pi_e)] + 
\mathbb{E}_{D_{\hat{R}}, s \sim d_0 a \sim \pi_e(\cdot|s)} \left[(\bar{R}(s_i,a_i) - \hat{R}^+(s_i,a_i))\right] \\
&=  \frac{1}{N} \sum_{i=1}^N  \mathbb{E}_{s \sim d_0 a \sim \pi_e}[\bar{R}(s_i, a_i)] \\
&= v(\pi_e)
\end{align}
The expectation of $\CDMIS$ when the annotations are biased is the value of the target policy. The variance of the annotation has no effect on the expectation of this estimator.  
\begin{proposition}[name=Unbiasedness of $\CDMIS$] \label{thm:CDM-IS_unbiasedness}
If both \Cref{asm:common-support-cf,asm:perfect-annot} hold, the $\CDMIS$ estimator is unbiased, $\mathbb{E}[\hat{V}^{\CDMIS}] = v(\pi_e)$. 
\end{proposition}
Proof: Under imperfect annotations, the estimator is an unbiased estimator of the value of the target policy. If we have additional assumptions about perfect annotations, this estimator is also unbiased. 

\subsubsection{Variance}
We now prove \Cref{thm:CDM-IS-imperfect_variance}, which is restated below. 
\begin{theorem*}[name=Variance of $\CDMIS$ under annotations with higher variance]
Under \Cref{asm:variance_annot} and \Cref{asm:common-support},  
\begin{align*}
    &N \cdot \mathbb{V}[\hat{V}^{\CDMIS}] = \mathbb{V}_{s \sim d_0}[v^{\pi_e}(s)] + \mathbb{E}_{s \sim d_0}\mathbb{E}_{a \sim \pi_b(s)}[\rho_s(a)^2 \sigma_R^2(s,a)] \\
    &\hspace{-2em}+ \mathbb{E}_{s \sim d_0} \Big[ \mathbb{E}_{a \sim \pi_b}[\rho_s(a)^2 \varepsilon_{\hat{R}^+}(s,a)^2] - \varepsilon_{\hat{R}^+}^{\pi_e}(s)^2 \Big] + \mathbb{E}_{s \sim d_0} \mathbb{E}_{a \sim \pi_b} \Big[ \big(\rho_s(a)^2 - \tfrac{1}{\pi_b(a|s)} \big) \mathbb{V}_{D_{\hat{R}^+}}[\hat{R}^+(s,a)]] \Big]
\end{align*}
{\small \text{where} $\varepsilon_{\hat{R}}^+(s,a) = \mathbb{E}_{D_{\hat{R}^+}}[\hat{R}^+(s,a)] - \bar{R}(s,a)$ and $\varepsilon_{\hat{R}^+}^{\pi_e}(s) = \mathbb{E}_{a \sim \pi_e}[\varepsilon_{\hat{R}^+}(s,a)]$.} 
\end{theorem*}
We use the law of total variance, the earlier expectation term, independence of samples, and the law of total variance.
\begin{align}
    &\mathbb{V}_{D_0, D_{\hat{R}^+}}[\hat{V}^{\CDMIS}] = \mathbb{V}_{D_0, D_{\hat{R}^+}}[\frac{1}{N}\sum_{i=1}^N \hat{R}^+(s_i, \pi_e) + \rho_{s_i}(a_i)(r_i- \hat{R}^+(s_i, a_i)] \\
    &= \mathbb{E}_{D_{\hat{R}^+}}[\mathbb{V}_{D_0}[\frac{1}{N}\sum_{i=1}^N \hat{R}^+(s_i, \pi_e) + \rho_{s_i}(a_i)(r_i- \hat{R}^+(s_i, a_i)]] \\
    &+ \mathbb{V}_{D_{\hat{R}^+}}[\mathbb{E}_{D_0}[\frac{1}{N}\sum_{i=1}^N \hat{R}^+(s_i, \pi_e) + \rho_{s_i}(a_i)(r_i- \hat{R}^+(s_i, a_i)]] \\
    &= \mathbb{E}_{D_{\hat{R}^+}}[\frac{1}{N}\sum_{i=1}^N \mathbb{V}_{s_i \sim d_0 a_i \sim \pi_b, r_i \sim R(s_i,a_i)}[\hat{R}^+(s_i, \pi_e) + \rho_{s_i}(a_i)(r_i- \hat{R}^+(s_i, a_i)]] \\
    &= \mathbb{E}_{D_{\hat{R}^+}}[\frac{1}{N}\sum_{i=1}^N \mathbb{E}_{s_i \sim d_0}[\mathbb{V}_{a_i \sim \pi_b, r_i \sim R(s_i, a_i)}[\hat{R}^+(s_i, \pi_e) + \rho_{s_i}(a_i)(r_i- \hat{R}^+(s_i, a_i)]] \\
    &+ \mathbb{V}_{s_i \sim d_0}[\mathbb{E}_{a_i \sim \pi_b, r_i \sim R(s_i, a_i)}[\hat{R}^+(s_i, \pi_e) + \rho_{s_i}(a_i)(r_i- \hat{R}^+(s_i, a_i)]]] \\
    &= \mathbb{E}_{D_{\hat{R}^+}}[\frac{1}{N}\sum_{i=1}^N \underbrace{\mathbb{E}_{s_i \sim d_0}[\mathbb{V}_{a_i \sim \pi_b, r_i \sim R(s_i, a_i)}[\hat{R}^+(s_i, \pi_e) + \rho_{s_i}(a_i)(r_i- \hat{R}^+(s_i, a_i)]]}_1 \\
    &+ \mathbb{V}_{s_i \sim d_0}[v(\pi_e)]] \\
    \text{Term 1} &= \mathbb{E}_{D_{\hat{R}^+}}[\frac{1}{N}\sum_{i=1}^N \mathbb{E}_{s_i \sim d_0}[\mathbb{E}_{a_i \sim \pi_b}[\mathbb{V}_{r_i \sim R(s_i, a_i)}[\hat{R}^+(s_i, \pi_e) + \rho_{s_i}(a_i)(r_i- \hat{R}^+(s_i, a_i)]] \\
    &+ \mathbb{E}_{s_i \sim d_0}[\mathbb{V}_{a_i \sim \pi_b}[\mathbb{E}_{r_i \sim R(s_i, a_i)}[\hat{R}^+(s_i, \pi_e) + \rho_{s_i}(a_i)(r_i- \hat{R}^+(s_i, a_i)]] \\
    &= \mathbb{E}_{D_{\hat{R}^+}}[\frac{1}{N}\sum_{i=1}^N \mathbb{E}_{s_i \sim d_0}[\mathbb{E}_{a_i \sim \pi_b}[\rho_{s_i}(a_i)^2 \sigma_R^2]] \\
    &+ \mathbb{E}_{s_i \sim d_0}[\mathbb{V}_{a_i \sim \pi_b}[\rho_{s_i}(a_i)(\bar{R} - \hat{R}^+)]]
\end{align}
The total variance term is now:
\begin{align}
    &=  \frac{1}{N}\sum_{i=1}^N \mathbb{V}_{s_i \sim d_0}[v(\pi_e)]] + \mathbb{E}_{s_i \sim d_0}[\mathbb{E}_{a_i \sim \pi_b}[\rho_{s_i}(a_i)^2 \sigma_R^2]] \\
    &+  \mathbb{E}_{D_{\hat{R}^+}}[\mathbb{E}_{s_i \sim d_0}[\mathbb{V}_{a_i \sim \pi_b}[\rho_{s_i}(a_i)(\bar{R} - \hat{R}^+)]]]
\end{align}
We simplify the last term, the only term that involves $D_{\hat{R}^+}$. 
\begin{align}
    &  \mathbb{E}_{D_{\hat{R}^+}}[\mathbb{E}_{s_i \sim d_0}[\mathbb{V}_{a_i \sim \pi_b}[\rho_{s_i}(a_i)(\bar{R} - \hat{R}^+)]]] \\
    &= \mathbb{E}_{D_{\hat{R}^+}}[\mathbb{E}_{s_i \sim d_0}[\mathbb{E}_{a_i \sim \pi_b}[\rho_{s_i}(a_i)^2 (\bar{R} - \hat{R}^+)^2] - \mathbb{E}_{a_i \sim \pi_b}[\rho_{s_i}(a_i)(\bar{R} - \hat{R}^+)]^2]] \\
    &= \underbrace{\mathbb{E}_{D_{\hat{R}^+}}\mathbb{E}_{s_i \sim d_0}\mathbb{E}_{a_i \sim \pi_b}[\rho_{s_i}(a_i)^2 (\bar{R} - \hat{R}^+)^2]}_{1} - \underbrace{\mathbb{E}_{D_{\hat{R}^+}}\mathbb{E}_{s_i \sim d_0}[\mathbb{E}_{a_i \sim \pi_b}[\rho_{s_i}(a_i)(\bar{R} - \hat{R}^+)]^2]}_{2} \\
    &\text{Term 1} = \mathbb{E}_{s \sim d_0}\mathbb{E}_{a \sim \pi_b}[\rho_{s}(a)^2 \mathbb{E}_{D_{\hat{R}^+}}[(\bar{R} - \hat{R}^+)^2]] \\
    &= \mathbb{E}_{s \sim d_0}\mathbb{E}_{a \sim \pi_b}[\rho_{s}(a)^2 \mathbb{E}_{D_{\hat{R}^+}}[\bar{R}(s,a)^2 - 2 \bar{R}(s,a) \hat{R}^+(s,a) + \hat{R}(s,a)^2]] \\
    &= \mathbb{E}_{s \sim d_0}\mathbb{E}_{a \sim \pi_b}\Big[\rho_{s}(a)^2 \big(\bar{R}(s,a)^2 - 2 \bar{R}(s,a) \tilde{R}(s,a) + \tilde{R}(s,a)^2 + \mathbb{V}_{D_{\hat{R}^+}}[\hat{R}^+(s,a)] \big)\Big] \\
    &\text{where } \tilde{R}(s,a) = \mathbb{E}_{D_{\hat{R}^+}}[\hat{R}^+(s,a)] \\ 
    &\text{Term 2} = \mathbb{E}_{D_{\hat{R}^+}}\mathbb{E}_{s \sim d_0}[\mathbb{E}_{a \sim \pi_e}[\bar{R}(s,a) - \hat{R}^+(s,a)]^2] \\
    &= \mathbb{E}_{s \sim d_0} \mathbb{E}_{D_{\hat{R}^+}} \Big[ \big(V^{\pi_e}(s) - \hat{R}^+(s,\pi_e) \big)^2 \Big] \\
    &= \mathbb{E}_{s \sim d_0} \mathbb{E}_{D_{\hat{R}^+}} \Big[ V^{\pi_e}(s)^2 - 2 V^{\pi_e}(s)\hat{R}^+(s,\pi_e) + \hat{R}^+(s,\pi_e)^2 \Big] \\
    &= \mathbb{E}_{s \sim d_0} \Big[ V^{\pi_e}(s)^2 - 2 V^{\pi_e}(s) \tilde{R}(s,\pi_e) + \tilde{R}(s, \pi_e)^2 + \mathbb{V}_{D_{\hat{R}^+}}[\hat{R}^+(s,\pi_e)] \Big] \\
    &\text{where } \tilde{R}(s, \pi_e) = \mathbb{E}_{D_{\hat{R}^+}}[\hat{R}^+(s,\pi_e)] \\
    & \text{Term 1 - Term 2} = \\
    & \mathbb{E}_{s \sim d_0}\mathbb{E}_{a \sim \pi_b}\Big[\rho_{s}(a)^2 \big(\bar{R}(s,a)^2 - 2 \bar{R}(s,a) \tilde{R}(s,a) + \tilde{R}(s,a)^2 + \mathbb{V}_{D_{\hat{R}^+}}[\hat{R}^+(s,a)] \big)\Big] \\
    & - \mathbb{E}_{s \sim d_0} \Big[ V^{\pi_e}(s)^2 - 2 V^{\pi_e}(s) \tilde{R}(s,\pi_e) + \tilde{R}(s, \pi_e)^2 + \mathbb{V}_{D_{\hat{R}^+}}[\hat{R}^+(s,\pi_e)] \Big]
\end{align}

If the DM estimate is unbiased, we have $\tilde{R}(s,a) = \mathbb{E}_{D_{\hat{R}^+}}[\hat{R}^+(s,a)] = \bar{R}(s,a)$. Substituting into the expression above results in cancellations and the expression becomes:
\begin{align}
    & \mathbb{E}_{s \sim d_0} \Big[\mathbb{E}_{a \sim \pi_b}\big[\rho_{s}(a)^2 \mathbb{V}_{D_{\hat{R}^+}}[\hat{R}^+(s,a)] \big] - \mathbb{V}_{D_{\hat{R}^+}}[\hat{R}^+(s,\pi_e)] \Big] \\
    &= \mathbb{E}_{s \sim d_0} \Big[\sum_{a} \pi_e(a|s)^2 \frac{1}{\pi_b(a|s)} \mathbb{V}_{D_{\hat{R}^+}}[\hat{R}^+(s,a)] - \sum_{a} \pi_e(a|s)^2 \mathbb{V}_{D_{\hat{R}^+}}[\hat{R}^+(s,a)] \Big] \\
    &= \mathbb{E}_{s \sim d_0} \Big[\sum_{a} \pi_e(a|s)^2 \big( \frac{1}{\pi_b(a|s)} - 1 \big) \mathbb{V}_{D_{\hat{R}^+}}[\hat{R}^+(s,a)] \Big] \\
    &= \mathbb{E}_{s \sim d_0} \mathbb{E}_{a \sim \pi_b} \Big[ \big(\rho_{s}(a)^2 - \tfrac{1}{\pi_b(a|s)} \big) \mathbb{V}_{D_{\hat{R}^+}}[\hat{R}^+(s,a)] \Big]
\end{align}

If the DM estimate is biased, let's say $\tilde{R}(s,a) = \mathbb{E}_{D_{\hat{R}^+}}[\hat{R}^+(s,a)] = \bar{R}(s,a) + \varepsilon_R(s,a)$, those terms no longer cancel and we get additional terms as follows: 
\begin{align}
    & \mathbb{E}_{s \sim d_0} \mathbb{E}_{a \sim \pi_b} \Big[\rho_{s}(a)^2 \big(\bar{R}(s,a)^2 - 2 \bar{R}(s,a) \tilde{R}(s,a) + \tilde{R}(s,a)^2\big)\Big] \\
    -& \mathbb{E}_{s \sim d_0} \Big[ V^{\pi_e}(s)^2 + 2 V^{\pi_e}(s)^2 \tilde{R}(s,\pi_e) + \tilde{R}(s, \pi_e)^2 \Big] \\
    &= \mathbb{E}_{s \sim d_0} \Big[ \mathbb{E}_{a \sim \pi_b}[\rho_{s}(a)^2 \varepsilon(s,a)^2] - \varepsilon^{\pi_e}(s)^2 \Big] \\
    & \text{where } \varepsilon^{\pi_e}(s) = \mathbb{E}_{D_{\hat{R}^+}}[\hat{R}(s,\pi_e)] - V^{\pi_e}(s) \\
    &= \mathbb{E}_{s \sim d_0} \Big[ \sum_{a} \pi_e(a|s)^2 \frac{1}{\pi_b(a|s)} \varepsilon(s,a)^2 - \Big(\sum_{a} \pi_e(a|s) \varepsilon(s,a) \Big)^2 \Big]
\end{align}

If we assume the bias of DM is constant across all $(s,a)$ such that $\varepsilon(s,a) = \varepsilon$, then the above expression becomes
\begin{align}
    &= \varepsilon^2 \mathbb{E}_{s \sim d_0} \Big[ \big( \sum_{a} \pi_e(a|s)^2 \frac{1}{\pi_b(a|s)} \big) - 1 \Big] \\
    & \geq 0 \ \text{by Sedrakyan's lemma, a special case of Cauchy-Schwarz inequality}
\end{align}

The whole variance term is now:
\begin{align}
    N \cdot \mathbb{V}[\hat{V}^{DM^+-IS}] &= \mathbb{V}_{s \sim d_0}[v^{\pi_e}(s)] + \mathbb{E}_{s \sim d_0}\mathbb{E}_{a \sim \pi_b(s)}[\rho_{s}(a)^2 \sigma_R^2(s,a)] \\
    &+ \mathbb{E}_{s \sim d_0} \Big[ \mathbb{E}_{a \sim \pi_b}[\rho_{s}(a)^2 \varepsilon(s,a)^2] - \varepsilon^{\pi_e}(s)^2 \Big] \\
    &+ \mathbb{E}_{s \sim d_0} \mathbb{E}_{a \sim \pi_b} \Big[ \big(\rho_{s}(a)^2 - \tfrac{1}{\pi_b(a|s)} \big) \mathbb{V}_{D_{\hat{R}^+}}[\hat{R}^+(s,a)] \Big]
\end{align}

In the case that the annotations are perfect, the whole variance term is altered to reflect the fact that $\varepsilon = 0$. 
\begin{proposition}[Variance of $\CDMIS$]
\label{thm:CDM-IS_variance}
If \Cref{asm:common-support-cf,asm:perfect-annot} hold, and the reward model is well-specified,
\begin{align}
    N \cdot \mathbb{V}[\hat{V}^{DM^+-IS}] &= \mathbb{V}_{s \sim d_0}[v^{\pi_e}(s)] + \mathbb{E}_{s \sim d_0}\mathbb{E}_{a \sim \pi_b(s)}[\rho_{s}(a)^2 \sigma_R^2(s,a)] \\
    &+ \mathbb{E}_{s \sim d_0} \mathbb{E}_{a \sim \pi_b} \Big[ \big(\rho_{s}(a)^2 - \tfrac{1}{\pi_b(a|s)} \big) \mathbb{V}_{D_{\hat{R}^+}}[\hat{R}^+(s,a)] \Big]
\end{align}
\end{proposition}

\subsection{Expectation and variance of \texorpdfstring{\DMCIS}{DM-IS⁺}}
The DR estimator is defined as
\begin{align*}
     \hat{V}^{\DMCIS} &= \frac{1}{N} \sum_{i=1}^N  \left(\hat{R}(s_i, \pi_e) + \sum_{a \in A} w_{i}^{a} \frac{\pi_e(a|s_i)}{\pi_{b^+}(a|s_i)}(c_i^a - \hat{R}(s_i, a))\right)
\end{align*}
\subsubsection{Expectation}
The expectation of the estimator assuming biased annotations is summarized in \Cref{thm:DR_aug_biased_annot} which is restated below. 
\begin{theorem*}[name=Expectation of $\DMCIS$ and $\CDMCIS$ under imperfect annotations]
Under \Cref{asm:biased_annot,asm:common-support-cf}, the two estimators have the same expectation: 
$$\mathbb{E}[\hat{V}^{\DMCIS}] = \mathbb{E}[\hat{V}^{\CDMCIS}] = v(\pi_e) + \mathbb{E}_{s_i \sim d_0}[\mathbb{E}_{a \sim \pi_e(s_i)}[(1- \frac{\bar{W}(a|s_i,a) \pi_b(a|s_i)}{\pi_b^+(a|s_i)}) \epsilon_G(s_i, a)]]$$
\end{theorem*}
Proof: We use the linearity of expectation, and the definition of the expectation. 
\begin{align}
    &\mathbb{E}_{D_0, D_{\hat{R}} \sim \mathcal{D}}[\hat{V}^{\DMCIS}] \\
    &= \mathbb{E}_{\substack{D_{\hat{R}}, s_i \sim d_0 a_i \sim \pi_b(s_i),\\ r \sim R(s_i, a_i), w \sim W(s_i,a_i)}}\left[ \frac{1}{N} \sum_{i=1}^N \left(\hat{R}(s_i, \pi_e) + \sum_{a \in A} w_{i}^{a} \frac{\pi_e(a|s_i)}{\pi_{b^+}(a|s_i)}(c_i^a - \hat{R}(s_i, a))\right)\right] \\
    &= \mathbb{E}_{\substack{D_{\hat{R}}, s_i \sim d_0 a_i \sim \pi_b(s_i),\\ r \sim R(s_i, a_i), w \sim W(s_i,a_i),\\ g_i^a \sim G(s_i, a)}}[\frac{1}{N} \sum_{i=1}^N (\hat{R}(s_i, \pi_e) + w_i^{a_i} \frac{\pi_e(a_i|s_i)}{\pi_{b^+}(a_i|s_i)}(r_i - \hat{R}(s_i, a_i)) \\
    &+ \sum_{a \in A \setminus \{a_i\}} w_{i}^{a} \frac{\pi_e(a|s_i)}{\pi_{b^+}(a|s_i)}(g_i^a - \hat{R}(s_i, a)))] \\
    &= \frac{1}{N} \sum_{i=1}^N \mathbb{E}_{D_{\hat{R}}, s_i \sim d_0}[\hat{R}(s_i, \pi_e)] \\
    &+ \underbrace{\mathbb{E}_{D_{\hat{R}}, s_i \sim d_0 a_i \sim \pi_b(s_i), r_i \sim R(s_i, a_i), w \sim W(s_i,a_i)}[w_i^{a_i} \frac{\pi_e(a_i|s_i)}{\pi_{b^+}(a_i|s_i)}(r_i - \hat{R}(s_i, a_i)) ]}_2\\
    &+ \underbrace{\mathbb{E}_{D_{\hat{R}}, s_i \sim d_0 a_i \sim \pi_b(s_i), w \sim W(s_i,a_i), g_i^a \sim G(s_i, a)}[\sum_{a \in A \setminus \{a_i\}} w_{i}^{a} \frac{\pi_e(a|s_i)}{\pi_{b^+}(a|s_i)}(g_i^{a} - \hat{R}(s_i, a))]}_3
\end{align}
Now we simplify term 2 using the expectation of the weights and the expectation of scalar reward, and the definition of expectation:
\begin{align}
    \text{Term 2} &= \mathbb{E}_{D_{\hat{R}}, s \sim d_0 a_i \sim \pi_b, r \sim R, w \sim W}[w_i^{a_i} \frac{\pi_e(a_i|s_i)}{\pi_{b^+}(a_i|s_i)}(r_i - \hat{R}(s_i, a_i)) ] \\
    &= \mathbb{E}_{D_{\hat{R}}, s \sim d_0 a_i \sim \pi_b, r \sim R}[\mathbb{E}_{w \sim W}[w_i^{a_i}] \frac{\pi_e(a_i|s_i)}{\pi_{b^+}(a_i|s_i)}(\mathbb{E}_{r \sim R}[r_i] - \hat{R}(s_i, a_i)) ] \\
    &= \mathbb{E}_{D_{\hat{R}}, s \sim d_0 a_i \sim \pi_b, r \sim R}[\bar{W}(a_i | s_i, a_i) \frac{\pi_e(a_i|s_i)}{\pi_{b^+}(a_i|s_i)}(\bar{R}(s_i, a_i) - \hat{R}(s_i, a_i)) ] \\
\end{align}
Now we simplify term 3 using a similar procedure:
\begin{align}
    \text{Term 3} &= \mathbb{E}_{D_{\hat{R}}, s \sim d_0 a \sim \pi_b, w \sim W, g \sim G}[\sum_{a \in A \setminus \{a_i\}} w_{i}^{a} \frac{\pi_e(a|s_i)}{\pi_{b^+}(a|s_i)}(g_i^{a} - \hat{R}(s_i, a))] \\
    &= \mathbb{E}_{D_{\hat{R}}, s \sim d_0 a \sim \pi_b}[\sum_{a \in A \setminus \{a_i\}} \mathbb{E}_{w \sim W}[w_{i}^{a}] \frac{\pi_e(a|s_i)}{\pi_{b^+}(a|s_i)}(\mathbb{E}_{g \sim G}[g_i^{a}] - \hat{R}(s_i, a))] \\
    &= \mathbb{E}_{D_{\hat{R}}, s \sim d_0 a \sim \pi_b}[\sum_{a \in A \setminus \{a_i\}} \bar{W}(a|s_i, a) \frac{\pi_e(a|s_i)}{\pi_{b^+}(a|s_i)}(\bar{R}(s_i, a) + \epsilon_G - \hat{R}(s_i, a))] \\
\end{align}
Putting together the terms:
\begin{align}
    \mathbb{E}_{s \sim d_0}[\hat{R}(s_i, \pi_e)] 
    &+ \mathbb{E}_{s \sim d_0 a_i \sim \pi_b, r \sim R}[\bar{W}(a_i | s_i, a_i) \frac{\pi_e(a_i|s_i)}{\pi_{b^+}(a_i|s_i)}(\bar{R}(s_i, a_i) - \hat{R}(s_i, a_i)) ] \\
    &+ \mathbb{E}_{s \sim d_0 a \sim \pi_b}[\sum_{a \in A \setminus \{a_i\}} \bar{W}(a|s_i, a) \frac{\pi_e(a|s_i)}{\pi_{b^+}(a|s_i)}(\bar{R}(s_i, a) + \epsilon_G - \hat{R}(s_i, a))] \\
\end{align}
Note that we can simplify the terms:
\begin{align}
     &= \mathbb{E}_{D_{\hat{R}}, s \sim d_0}[\hat{R}(s_i, \pi_e)] \\
     &+ \mathbb{E}_{D_{\hat{R}}, s \sim d_0 a_i \sim \pi_b, r \sim R}[\sum_{a \in A} \bar{W}(a_i | s_i, a_i) \frac{\pi_e(a_i|s_i)}{\pi_{b^+}(a_i|s_i)}(\bar{R}(s_i, a_i) - \hat{R}(s_i, a_i)) ] \\
     &+ \mathbb{E}_{D_{\hat{R}}, s \sim d_0 a \sim \pi_b}[\sum_{a \in A \setminus \{a_i\}} \bar{W}(a|s_i, a) \frac{\pi_e(a|s_i)}{\pi_{b^+}(a|s_i)}(\epsilon_G)]
\end{align}
The sum of the first two terms becomes the value of the target policy:
\begin{align}
    &\mathbb{E}_{D_{\hat{R}}, s \sim d_0}[\hat{R}(s_i, \pi_e)] + \mathbb{E}_{D_{\hat{R}}, s \sim d_0 a_i \sim \pi_b, r \sim R}[\sum_{a \in A} \bar{W}(a_i | s_i, a_i) \frac{\pi_e(a_i|s_i)}{\pi_{b^+}(a_i|s_i)}(\bar{R}(s_i, a_i) - \hat{R}(s_i, a_i)) ] \\
    &= \mathbb{E}_{D_{\hat{R}}, s \sim d_0}[\hat{R}(s_i, \pi_e)] \\
    &+ \mathbb{E}_{D_{\hat{R}}, s \sim d_0 r \sim R}[\sum_{a_i \in A} \pi_b(a_i|s_i) \sum_{a \in A} \bar{W}(a_i | s_i, a_i) \frac{\pi_e(a_i|s_i)}{\pi_{b^+}(a_i|s_i)}(\bar{R}(s_i, a_i) - \hat{R}(s_i, a_i)) ] \\
    &= \mathbb{E}_{D_{\hat{R}}, s \sim d_0}[\hat{R}(s_i, \pi_e)] + \mathbb{E}_{D_{\hat{R}}, s \sim d_0 r \sim R}[\pi_e(a_i|s_i)(\bar{R}(s_i, a_i) - \hat{R}(s_i, a_i)) ] \\
    &= \mathbb{E}_{D_{\hat{R}}, s \sim d_0}[\hat{R}(s_i, \pi_e)] + \mathbb{E}_{s \sim d_0 a_i \sim \pi_e, r \sim R}[(\bar{R}(s_i, a_i) - \hat{R}(s_i, a_i))] \\
    &= \mathbb{E}_{s \sim d_0 a_i \sim \pi_e}[(\bar{R}(s_i, a_i)] \\
    &= v(\pi_e)
\end{align}
The final term can be simplified as follows:
\begin{align}
    &\mathbb{E}_{D_{\hat{R}}, s \sim d_0 a \sim \pi_b}[\sum_{\tilde{a} \in A \setminus \{a_i\}} \bar{W}(\tilde{a}|s_i, a) \frac{\pi_e(\tilde{a}|s_i)}{\pi_{b^+}(\tilde{a}|s_i)} \epsilon_G] \\
    &=\mathbb{E}_{D_{\hat{R}}, s \sim d_0}[\sum_{a \in A} \pi_b(a | s_i) \left(\sum_{\tilde{a} \in A \setminus \{a_i\}} \bar{W}(\tilde{a}|s_i, a) \frac{\pi_e(\tilde{a}|s_i)}{\pi_{b^+}(\tilde{a}|s_i)} \epsilon_G\right)] \\
    &= \mathbb{E}_{D_{\hat{R}}, s \sim d_0}[\sum_{\tilde{a} \in A}((\sum_{a \in A} \pi_b(a|s) \bar{W}(\tilde{a}|s, a)) \frac{\pi_e(\tilde{a}|s)}{\pi_b^+(\tilde{a}|s} \epsilon_G))] \\
    &- \mathbb{E}_{D_{\hat{R}}, s \sim d_0}[(\sum_{a \in A} \pi_b(a|s) \bar{W}(a|s,a) \frac{\pi_e(a|s)}{\pi_b^+(a|s)} \epsilon_G)] \\
    &= \mathbb{E}_{D_{\hat{R}}, s \sim d_0}[\sum_{\tilde{a} \in A} \pi_e(\tilde{a}|s) \epsilon_G))] \\
    &- \mathbb{E}_{D_{\hat{R}}, s \sim d_0}[(\sum_{a \in A} \pi_e(a|s) \bar{W}(a|s,a) \frac{\pi_b(a|s)}{\pi_b^+(a|s)} \epsilon_G)] \\
    &= \mathbb{E}_{s \sim d_0}[\mathbb{E}_{a \sim \pi_e}[\left(1- \frac{\bar{W}(a|s,a) \pi_b(a|s)}{\pi_b^+(a|s)}\right) \epsilon_G]]
\end{align}
Thus, the final term for the expectation is:
\begin{align}
    &v(\pi_e) + \mathbb{E}_{s \sim d_0}\left[\mathbb{E}_{a \sim \pi_e}\left[\left(1- \frac{\bar{W}(a|s,a) \pi_b(a|s)}{\pi_b^+(a|s)}\right) \epsilon_G\right]\right]
\end{align}
\begin{proposition}[name=Unbiasedness of $\DMCIS$] \label{thm:DM+C-IS_unbiasedness}
If both \Cref{asm:common-support-cf,asm:perfect-annot} hold, the $\DMCIS$ estimator is unbiased, $\mathbb{E}[\hat{V}^{\DMCIS}] = v(\pi_e)$. 
\end{proposition}
Proof: If $\epsilon_G = 0$, then, the expectation term for the imperfect annotations case reduces to the value of the expert policy $v(\pi_e)$. 

\subsubsection{Variance}

We first study the variance of the $\DMCIS$ estimator under imperfect annotations. 

\begin{proposition}[name=Variance of $\DMCIS$ under imperfect annotations]
\label{prop:dmcis_imperfect_variance}
In the case that \Cref{asm:common-support-cf,asm:biased_annot,asm:variance_annot} hold, 
\begin{align}
    &\mathbb{V}_{D \sim \mathcal{D}}[\hat{V}^{\DMCIS}] =\frac{1}{N^2}\sum_{i=1}^N\mathbb{V}_{s_i \sim d_0} \Bigg[ v(\pi_e) + \mathbb{E}_{\substack{s_i \sim d_0 \\ a \sim \pi_e(\cdot| s_i)}}\Big[(1- \frac{\bar{W}(a|s_i,a) \pi_b(a|s_i)}{\pi_b^+(a|s_i)}) \epsilon_G(s_i, a)\Big] \Bigg] \\
    &+ \mathbb{E}_{s_i \sim d_0}\Bigg[\mathbb{V}_{a_i \sim \pi_b(\cdot|s_i)}\Big[
    \frac{\pi_e(s_i| a_i)}{\pi_b^+(s_i| a_i)} \bar{W}(a_i|s_i, a_i)(\bar{R}(s_i, a_i) - \hat{R}(s_i, a_i)) \\
    &+ \sum_{a \in A \setminus \{a_i\}}  \frac{\pi_e(a|s_i)}{\pi_b^+(a|s_i)} \bar{W}(a|s_i, a_i) (\bar{R}(s_i, a) + \epsilon_R(s_i, a) - \hat{R}(s_i,a)) \Big| a_i \Big]\Bigg| s_i \Bigg] \\
    &+ \mathbb{E}_{s_i \sim d_0}\Bigg[\mathbb{E}_{a_i \sim \pi_b}\Big[\sum_{a \in A} \sum_{a \in A} \bar{W}(a | s_i, a_i)^2 \frac{\pi_e(a|s_i)}{\pi_b^+(a|s_i)}^2 \sigma_R^2(s_i,a) \\
    &+ \sum_{a \in A \setminus \{a_i\}}\bar{W}(a | s_i, a_i)^2 \frac{\pi_e(a|s_i)}{\pi_b^+(a|s_i)}^2 \Delta_G(s_i, a) \Big| s_i\Big]\Bigg] \\
    &+ \mathbb{E}_{s_i \sim d_0}\Bigg[\mathbb{E}_{a_i \sim \pi_b(\cdot|s_i)}\Big[\frac{\pi_e(a_i|s_i)}{\pi_b^+(a_i|s_i)}^2 \mathbb{V}_{w_{i}^{a} \sim W(s_i, a)}[w_{i}^{a_i}] \\
     & \times \Bigg(\sigma_R^2(s_i, a) + (\bar{R}(s_i, a) - \hat{R}(s_i,a))^2\Bigg) \\
     &+ \sum_{a \in A \setminus \{a_i\}} \frac{\pi_e(a|s_i)}{\pi_b^+(a|s_i)}^2 \mathbb{V}_{w_{i}^{a} \sim W(s_i, a)}[w_{i}^{a}] \\
     &\times \Bigg(\sigma_R^2(s_i, a) + \Delta_G(s_i, a) + (\bar{R}(s_i, a) + \epsilon_R(s_i, a) - \hat{R}(s_i,a))^2\Bigg) \\
     &+ 2 \bar{\psi}_R(s_i, a_i)\sum_{\tilde{a}_j \neq \tilde{a}_i} \bar{\psi}_G(s_i, \tilde{a}_j)  \times Cov_{w_{i}^{a} \sim W(s_i,a)}\Big( w_i^{\tilde{a}_j}, w_i^{a_i}\Big)\\ 
     &+ \sum_{\tilde{a}_j \neq a_i} \sum_{\tilde{a}_k \neq a_i, \tilde{a}_k \neq \tilde{a}_j} \bar{\psi}_G(s_i, \tilde{a}_j)\bar{\psi}_G(s_i, \tilde{a}_k)\\
     &\times  Cov_{w_{i}^{a} \sim W(s_i,a)}\Bigg( w_i^{\tilde{a}_j}, w_i^{\tilde{a}_k}\Bigg) \Bigg| a_i\Bigg] \Bigg| s_i\Bigg]
\end{align}
\end{proposition}

Proof:
We start by moving the variance inside the summation because the variance of each term that the summation is over is independent. This is a similar strategy proposed in the  Let $c_i^{a} \sim C(s_i, a_i)$ be $r_i^a \sim R(s_i, a)$ if $c_i^a$ is a factual reward and $g_i^a \sim G(s_i, a)$ if $c_i^a$ is a counterfactual annotation. 
\begin{align}
    &\mathbb{V}_{D \sim \mathcal{D}}[\hat{V}^{\DMCIS}] \\
    &= \mathbb{V}_{\substack{s_i \sim d_0 a_i \sim \pi_b(\cdot|s_i),\\c_i^a \sim C(s_i, a), w_{i}^{a} \sim W(s_i, a)}}\left[\frac{1}{N} \sum_{i=1}^N
    \left(\hat{R}(s_i,\pi_e) + \sum_{a \in A} w_{i}^{a} 
    \frac{\pi_e(a|s_i)}{\pi_{b^+}(a|s_i)}(c_i^{a} - \hat{R}(s_i,a))\right)\right] \\
    &= \frac{1}{N^2} \sum_{i=1}^N \mathbb{V}_{\substack{s_i \sim d_0 a_i \sim \pi_b(\cdot|s_i),\\ c_i^a \sim C(s_i, a), w_{i}^{a} \sim W(s_i, a)}}\left[
    \hat{R}(s_i,\pi_e) + \sum_{a \in A} w_{i}^{a} 
    \frac{\pi_e(a|s_i)}{\pi_{b^+}(a|s_i)}(c_i^{a} - \hat{R}(s_i,a))\right]
\end{align}
Now, we use the law of total variance to separate the joint distribution that the variance is over:
\begin{align}
    &= \frac{1}{N^2} \sum_{i=1}^N \underbrace{\mathbb{E}_{s_i \sim d_0} \left[\mathbb{V}_{\substack{a_i \sim \pi_b(\cdot|s_i),\\ c_i^a \sim C(s_i, a),\\ w_{i}^{a}\sim W(s_i, a)}}\left[ \hat{R}(s_i, \pi_e) + \sum_{a \in A} w_{i}^{a} \frac{\pi_e(a|s_i)}{\pi_b^+(a|s_i)} (c_i^a - \hat{R}(s_i,a)) \middle| s_i\right]\right]}_2 \\
    &+ \mathbb{V}_{s_i \sim d_0} \left[\underbrace{\mathbb{E}_{\substack{a_i \sim \pi_b(\cdot|s_i),\\ c_i^a \sim C(s_i, a),\\ w_{i}^{a} \sim W(s_i, a)}}\left[ \hat{R}(s_i, \pi_e) + \sum_{a \in A} w_{i}^{a} \frac{\pi_e(a|s_i)}{\pi_b^+(a|s_i)} (c_i^a- \hat{R}(s_i,a)) \middle| s_i\right]}_1\right]
\end{align}
Note that Term 1 is just the value of the target policy conditioned on a specific $s_i$. We have already proved the bias of the estimator under imperfect annotations. This makes the whole variance now:
\begin{align}
    &= \frac{1}{N^2} \sum_{i=1}^N \underbrace{\mathbb{E}_{s_i \sim d_0} \left[\mathbb{V}_{\substack{a_i \sim \pi_b(\cdot|s_i),\\ c_i^a \sim C(s_i, a),\\ w_{i}^{a} \sim W(s_i, a)}}\left[ \hat{R}(s_i, \pi_e) + \sum_{a \in A} w_{i}^{a} \frac{\pi_e(a|s_i)}{\pi_b^+(a|s_i)} (c_i^a - \hat{R}(s_i,a)) \middle| s_i\right]\right]}_2 \\
    &+ \mathbb{V}_{s_i \sim d_0} \left[ v(\pi_e) + \mathbb{E}_{\substack{s_i \sim d_0 \\ a \sim \pi_e(\cdot|s_i)}}\left[\left(1- \frac{\bar{W}(a|s_i,a) \pi_b(a|s_i)}{\pi_b^+(a|s_i)}\right) \epsilon_G(s_i, a)\right]\right]
\end{align}
Now we decompose term 2 using the law of total variance:
\begin{align}
    &\text{Term 2} = \mathbb{E}_{s_i \sim d_0} \left[\mathbb{V}_{\substack{a_i \sim \pi_b(\cdot|s_i),\\c_i^a \sim C(s_i, a),\\ w_{i}^{a} \sim W(s_i, a)}}\left[ \hat{R}(s_i, \pi_e) + \sum_{a \in A} w_{i}^{a} \frac{\pi_e(a|s_i)}{\pi_b^+(a|s_i)} (c_i^a - \hat{R}(s_i,a)) \middle| s_i\right]\right] \\
    &= \mathbb{E}_{s_i \sim d_0} \Bigg[\underbrace{\mathbb{E}_{a_i \sim \pi_b(\cdot|s_i)}\left[ \mathbb{V}_{\substack{c_i^a \sim C(s_i, a),\\ w_{i}^{a} \sim W(s_i, a)}}\left[\hat{R}(s_i, \pi_e) + \sum_{a \in A} w_{i}^{a} \frac{\pi_e(a|s_i)}{\pi_b^+(a|s_i)} (c_i^a - \hat{R}(s_i,a)) \Bigg| a_i \right] \Bigg| s_i\right]}_3 \\
    &+ \underbrace{\mathbb{V}_{a_i \sim \pi_b(\cdot|s_i)}\left[ \mathbb{E}_{\substack{c_i^a \sim C(s_i, a),\\ w_{i}^{a} \sim W(s_i, a)}}\left[\hat{R}(s_i, \pi_e) + \sum_{a \in A} w_{i}^{a} \frac{\pi_e(a|s_i)}{\pi_b^+(a|s_i)} (c_i^a - \hat{R}(s_i,a)) \middle| a_i \right] \Bigg| s_i\right]}_4\Bigg]
\end{align}
Now we decompose term 4 by distributing the inner-most expectation. Note that the inner-most expectation only depends on the distribution over $c_i$ and $w_i$, which is not in many of the terms:
\begin{align}
    \text{Term 4} &= \mathbb{E}_{s_i \sim d_0}\left[\mathbb{V}_{a_i \sim \pi_b(\cdot|s_i)}\left[ \mathbb{E}_{\substack{c_i^a \sim C(s_i, a),\\ w_{i}^{a} \sim W(s_i, a)}}\left[\hat{R}(s_i, \pi_e) + \sum_{a \in A} w_{i}^{a} \frac{\pi_e(a|s_i)}{\pi_b^+(a|s_i)} (c_i^a - \hat{R}(s_i,a)) \middle| a_i \right] \middle| s_i\right]\right] \\
    &= \mathbb{E}_{s_i \sim d_0}\left[\mathbb{V}_{a_i \sim \pi_b(\cdot|s_i)}\left[ \hat{R}(s_i, \pi_e) +\mathbb{E}_{\substack{c_i^a \sim C(s_i, a),\\ w_{i}^{a} \sim W(s_i, a)}}\left[\sum_{a \in A} w_{i}^{a} \frac{\pi_e(a|s_i)}{\pi_b^+(a|s_i)} (c_i^a - \hat{R}(s_i,a)) \middle| a_i \right] \middle| s_i\right]\right]
\end{align}
Applying linearity of expectation and the fact that $c, w$ are conditionally independent given $s, a$:
\begin{align}
    &= \mathbb{E}_{s_i \sim d_0}\left[\mathbb{V}_{a_i \sim \pi_b(\cdot|s_i)}\left[\hat{R}(s_i, \pi_e) + \sum_{a \in A}  \frac{\pi_e(a|s_i)}{\pi_b^+(a|s_i)}\mathbb{E}_{\substack{c_i^a \sim C(s_i, a),\\ w_{i}^{a} \sim W(s_i, a)}}\left[ w_{i}^{a} (c_i^a - \hat{R}(s_i,a)) \middle| a_i \right] \middle| s_i\right]\right] \\
    &= \mathbb{E}_{s_i \sim d_0}\left[\mathbb{V}_{a_i \sim \pi_b(\cdot|s_i)}\left[\sum_{a \in A}  \frac{\pi_e(a|s_i)}{\pi_b^+(a|s_i)}\mathbb{E}_{\substack{c_i^a \sim C(s_i, a),\\ w_{i}^{a} \sim W(s_i, a)}}\left[ w_{i}^{a} (c_i^a - \hat{R}(s_i,a)) \middle| a_i \right] \middle| s_i\right]\right] \\
\end{align}
Now, separating into factual and counterfactual:
\begin{align}
    &= \mathbb{E}_{s_i \sim d_0}\Bigg[\mathbb{V}_{a_i \sim \pi_b(\cdot|s_i)}\Big[
    \frac{\pi_e(s_i| a_i)}{\pi_b^+(s_i| a_i)} \bar{W}(a_i|s_i, a_i)(\bar{R}(s_i, a_i) - \hat{R}(s_i, a_i)) \\
    &+ \sum_{a \in A \setminus \{a_i\}}  \frac{\pi_e(a|s_i)}{\pi_b^+(a|s_i)} \bar{W}(a|s_i, a_i) (\bar{R}(s_i, a) + \epsilon_R(s_i, a) - \hat{R}(s_i,a)) \Big| a_i, s_i \Big]\Bigg] \\
\end{align}
We cannot decompose this term any further because each weight is dependent on $a_i$, and the variance is over the distribution of $a_i$. Now we decompose term 3 using the law of total variance:
\begin{align}
    &\text{Term 3} = \mathbb{E}_{s_i \sim d_0}\left[\mathbb{E}_{a_i \sim \pi_b(\cdot|s_i)}\left[ \mathbb{V}_{\substack{c_i^a \sim C(s_i, a),\\ w_{i}^{a} \sim W(s_i, a)}}\left[\hat{R}(s_i, \pi_e) + \sum_{a \in A} w_{i}^{a} \frac{\pi_e(a|s_i)}{\pi_b^+(a|s_i)} (c_i^a - \hat{R}(s_i,a)) \middle| a_i \right] \middle| s_i\right]\right] \\
    &= \mathbb{E}_{s_i \sim d_0}\Bigg[\mathbb{E}_{a_i \sim \pi_b(\cdot|s_i)}\Bigg[\underbrace{\mathbb{E}_{c_i^a \sim C(s_i, a)}\Big[\mathbb{V}_{w_{i}^{a} \sim W(s_i, a)}\Big[ \hat{R}(s_i, \pi_e) + \sum_{a \in A} w_{i}^{a} \frac{\pi_e(a|s_i)}{\pi_b^+(a|s_i)} (c_i^a - \hat{R}(s_i,a))\Big| c_i\Big] \Bigg|a_i\Bigg]}_5 \\
    &+ \underbrace{\mathbb{V}_{c_i^a \sim C(s_i, a)}\Bigg[\mathbb{E}_{w_{i}^{a} \sim W(s_i, a)}\Big[ \hat{R}(s_i, \pi_e) + \sum_{a \in A} w_{i}^{a} \frac{\pi_e(a|s_i)}{\pi_b^+(a|s_i)} (c_i^a - \hat{R}(s_i,a))\Big| c_i\Big] \Big|a_i\Big]}_6\Bigg| s_i\Bigg]\Bigg]
\end{align}
Now we decompose term 6 by distributing the inner-most expectation. Most terms do not have a weight in them:
\begin{align}
    &\text{Term 6} = \\
    &\mathbb{E}_{s_i \sim d_0}\left[\mathbb{E}_{a_i \sim \pi_b(\cdot|s_i)}\left[\mathbb{V}_{c_i^a \sim C(s_i, a)}\left[\mathbb{E}_{w_{i}^{a} \sim W(s_i, a)}\left[ \hat{R}(s_i, \pi_e) + \sum_{a \in A} w_{i}^{a} \frac{\pi_e(a|s_i)}{\pi_b^+(a|s_i)} (c_i^a - \hat{R}(s_i,a))\middle| c_i\right] \middle|a_i\right]\middle| s_i\right]\right] \\
    &= \mathbb{E}_{s_i \sim d_0}\left[\mathbb{E}_{a_i \sim \pi_b(\cdot|s_i)}\left[\mathbb{V}_{c_i^a \sim R(s_i, a_i)}\left[\hat{R}(s_i, \pi_e) + \sum_{a \in A} \mathbb{E}_{w_{i}^{a} \sim W(s_i, a)}\left[w_{i}^{a}\right] \frac{\pi_e(a|s_i)}{\pi_b^+(a|s_i)} (c_i^a - \hat{R}(s_i,a))\middle| a_i\right] \middle| s_i\right]\right] \\
    &= \mathbb{E}_{s_i \sim d_0}\left[\mathbb{E}_{a_i \sim \pi_b(\cdot|s_i)}\left[\mathbb{V}_{c_i^a \sim C(s_i, a)}\left[\hat{R}(s_i, \pi_e) + \sum_{a \in A} \bar{W}(a | s_i, a_i) \frac{\pi_e(a|s_i)}{\pi_b^+(a|s_i)} (c_i^a - \hat{R}(s_i,a))\middle| a_i\right] \middle| s_i\right]\right]
\end{align}
Notice that the inner-most variance term now is over $c_i^a$, so all terms that do not consider $c_i^a$ are considered constants. 
\begin{align}
    &= \mathbb{E}_{s_i \sim d_0}\left[\mathbb{E}_{a_i \sim \pi_b(\cdot|s_i)}\left[\sum_{a \in A} \bar{W}(a | s_i, a_i)^2 \frac{\pi_e(a|s_i)}{\pi_b^+(a|s_i)}^2 (\mathbb{V}_{c_i^a \sim C(s_i, a)}\left[c_i| a_i\right]) | s_i\right]\right] 
\end{align}
Now, because the annotations are assumed to be imperfect, we must split the variance term into a factual sample and several possible counterfactual annotations. 
\begin{align}
    &= \mathbb{E}_{s_i \sim d_0}\Bigg[\mathbb{E}_{a_i \sim \pi_b(\cdot|s_i)}\Bigg[\bar{W}(a_i | s_i, a_i)^2 \frac{\pi_e(a_i|s_i)}{\pi_b^+(a_i|s_i)}^2 \sigma_R^2(s_i,a_i)\\
    &+ \sum_{a \in A \setminus \{a_i\}}\bar{W}(a | s_i, a_i)^2 \frac{\pi_e(a|s_i)}{\pi_b^+(a|s_i)}^2 \left(\sigma_R^2(s_i,a) + \Delta_G(s_i, a) \right)\Bigg| s_i\Bigg]\Bigg] \\
    &= \mathbb{E}_{s_i \sim d_0}\Bigg[\mathbb{E}_{a_i \sim \pi_b(\cdot|s_i)}\Bigg[\sum_{a \in A} \bar{W}(a | s_i, a_i)^2 \frac{\pi_e(a|s_i)}{\pi_b^+(a|s_i)}^2 \sigma_R^2(s_i,a) \\
    &+ \sum_{a \in A \setminus \{a_i\}}\bar{W}(a | s_i, a_i)^2 \frac{\pi_e(a|s_i)}{\pi_b^+(a|s_i)}^2 \Delta_G(s_i, a) \Bigg| s_i\Bigg]\Bigg] \\
\end{align}
Now we decompose term 5. Because the weights are not independent (they need to sum to 1 across all the actions/annotations for a given context), we need to consider the covariance between them.  First, we move the variance term outside the summation because $\hat{R}(s_i, \pi_e)$ is considered a constant.
\begin{align}
    &\text{Term 5} = \mathbb{E}_{s_i \sim d_0}\Bigg[\mathbb{E}_{a_i \sim \pi_b(\cdot|s_i)(\cdot|s_i)}\Bigg[\mathbb{E}_{c_i^a \sim C(s_i, a)}\Bigg[\mathbb{V}_{w_{i}^{a} \sim W(s_i, a)}\Bigg[ \hat{R}(s_i, \pi_e) \\
    &+ \sum_{a \in A} w_{i}^{a} \frac{\pi_e(a|s_i)}{\pi_b^+(a|s_i)} (c_i^a - \hat{R}(s_i,a))\Bigg| c_i\Bigg] \Bigg|a_i\Bigg] \Bigg|s_i\Bigg]\Bigg] \\
    &= \mathbb{E}_{s_i \sim d_0}\left[\mathbb{E}_{a_i \sim \pi_b(\cdot|s_i)(\cdot|s_i)}\left[\mathbb{E}_{c_i^a \sim C(s_i, a)}\left[\mathbb{V}_{w_{i}^{a} \sim W(s_i, a)}\left[\sum_{a \in A} w_{i}^{a} \frac{\pi_e(a|s_i)}{\pi_b^+(a|s_i)} (c_i^a - \hat{R}(s_i,a))\middle| c_i\right] \middle|a_i\right] \middle|s_i\right]\right]
\end{align}
When we move the variance inside the summation if we also consider the covariance of each term in the summation. Then, we simplify and consider all terms that don't contain a weight as constant.
\begin{align}
    &= \mathbb{E}_{s_i \sim d_0}\Bigg[\mathbb{E}_{a_i \sim \pi_b(\cdot|s_i)}\Bigg[\mathbb{E}_{c_i^a \sim C(s_i, a)}\Bigg[\sum_{a \in A} \mathbb{V}_{w_{i}^{a} \sim W(s_i, a)}\Bigg[w_{i}^{a} \frac{\pi_e(a|s)}{\pi_b^+(a|s)} (c_i^a - \hat{R}(s_i,a))\Bigg| c_i\Bigg] \\
    &+ 2 \sum_{\tilde{a}_j, \tilde{a}_k}^{\tilde{a}_j \neq \tilde{a}_k} Cov_{w_{i}^{a} \sim W(s_i,a)}\Bigg( w_i^{\tilde{a}_j} \frac{\pi_e(\tilde{a}_j|s_i)}{\pi_b^+(\tilde{a}_j|s_i)} (c_i^{\tilde{a}_j} - \hat{R}(s_i,\tilde{a}_j)), w_i^{ \tilde{a}_k}\frac{\pi_e(\tilde{a}_k|s_i)}{\pi_b^+(\tilde{a}_k|s_i)} \Bigg(c_i^{\tilde{a}_k} - \hat{R}(s_i,\tilde{a}_k))\Bigg) \Bigg| a_i\Bigg] \Bigg| s_i \Bigg]\Bigg] \\
    &= \mathbb{E}_{s_i \sim d_0}\Bigg[\mathbb{E}_{a_i \sim \pi_b(\cdot|s_i)}\Bigg[\mathbb{E}_{c_i^a \sim C(s_i, a)}\Bigg[\sum_{a \in A} \frac{\pi_e(a|s_i)}{\pi_b^+(a|s_i)}^2 (c_i^a - \hat{R}(s_i,a))^2 \mathbb{V}_{w_{i}^{a} \sim W(s_i, a)}\Big[w_{i}^{a}\Big] \\
    &+ 2 \sum_{\tilde{a}_j, \tilde{a}_k}^{\tilde{a}_j \neq \tilde{a}_k} \frac{\pi_e(\tilde{a}_j|s_i)}{\pi_b^+(\tilde{a}_j|s_i)} (c_i^{\tilde{a}_j} - \hat{R}(s_i,\tilde{a}_j)) \frac{\pi_e(\tilde{a}_k|s_i)}{\pi_b^+(\tilde{a}_k|s_i)} (c_i^{\tilde{a}_k} - \hat{R}(s_i,\tilde{a}_k)) \\
    &\times Cov_{w_{i}^{a} \sim W(s_i,a)}\Bigg( w_i^{\tilde{a}_j}, w_i^{\tilde{a}_k}\Bigg) \Bigg| a_i\Bigg] \Bigg| s_i\Bigg]\Bigg]
\end{align}
Now, we separate the terms to account for different distributions of the factual rewards and counterfactual annotations. Let $\psi(s_i, a) = \frac{\pi_e(a|s_i)}{\pi_b^+(a|s_i)} (c_i^{a} - \hat{R}(s_i,a))$. Then, we can re-write the expression as:
\begin{align}
     &\mathbb{E}_{s_i \sim d_0}\Bigg[\mathbb{E}_{a_i \sim \pi_b(\cdot|s_i)}\Bigg[\mathbb{E}_{c_i^a \sim C(s_i, a)}\Bigg[\sum_{a \in A} \frac{\pi_e(a|s_i)}{\pi_b^+(a|s_i)}^2 (c_i^a - \hat{R}(s_i,a))^2 \mathbb{V}_{w_{i}^{a} \sim W(s_i, a)}\Big[w_{i}^{a}\Big] \\
     &+ 2 \sum_{\tilde{a}_j \neq \tilde{a}_k} \psi(s_i, \tilde{a}_j) \psi(s_i, \tilde{a}_k) \\
     &\times  Cov_{w_{i}^{a} \sim W(s_i,a)}\Bigg( w_i^{\tilde{a}_j}, w_i^{\tilde{a}_k}\Bigg) \Bigg|a_i\Bigg] |s_i\Bigg]\Bigg]
\end{align}
Now, splitting into factual and counterfactual:
\begin{align}
     &\mathbb{E}_{s_i \sim d_0}\Bigg[\mathbb{E}_{a_i \sim \pi_b(\cdot|s_i)}\Bigg[\frac{\pi_e(a_i|s_i)}{\pi_b^+(a_i|s_i)}^2 \mathbb{E}_{r_i^{a_i} \sim R(s_i, a_i)}\Bigg[(r_i^a - \hat{R}(s_i,a))^2]\mathbb{V}_{w_{i}^{a} \sim W(s_i, a)}\Big[w_{i}^{a_i}\Big] \\
     &+ \sum_{a \in A \setminus \{a_i\}} \frac{\pi_e(a|s_i)}{\pi_b^+(a|s_i)}^2 \mathbb{E}_{g_i^a \sim G(s_i, a)}\Big[(g_i^a - \hat{R}(s_i,a))^2\Big] \mathbb{V}_{w_{i}^{a} \sim W(s_i, a)}\Big[w_{i}^{a}\Big] \\
     &+ 2 \mathbb{E}_{r_i^{a_i} \sim R(s_i, a_i)}\Big[\psi(s_i, a_i)\Big]\sum_{\tilde{a}_j \neq \tilde{a}_i} \mathbb{E}_{g_i^a \sim G(s_i, a)}\Big[\psi(s_i, \tilde{a}_j)\Big]  \times Cov_{w_{i}^{a} \sim W(s_i,a)}\Big( w_i^{\tilde{a}_j}, w_i^{a_i}\Big)\\ 
     &+ \sum_{\tilde{a}_j \neq a_i} \sum_{\tilde{a}_k \neq a_i, \tilde{a}_k \neq \tilde{a}_j} \mathbb{E}_{g_i^a \sim G(s_i, a)}\Big[\psi(s_i, \tilde{a}_j) \psi(s_i, \tilde{a}_k)\Big] \\
     &\times  Cov_{w_{i}^{a} \sim W(s_i,a)}\Bigg( w_i^{\tilde{a}_j}, w_i^{\tilde{a}_k}\Bigg) \Bigg|a_i\Bigg] \Bigg|s_i\Bigg]\Bigg]
\end{align}
 Recall that $\mathbb{E}_{r_i^{a_i} \sim R(s_i, a_i)}[\psi(s_i, a)] = \bar{\psi}_R(s_i, a) =  \frac{\pi_e(a|s_i)}{\pi_b^+(a|s_i)} (\bar{R}(s_i, a) - \hat{R}(s_i,a))$  and $\mathbb{E}_{g_i^a \sim G(s_i, a)}[\psi(s_i, a)] = \bar{\psi}_G(s_i, a) = \frac{\pi_e(a|s_i)}{\pi_b^+(a|s_i)} (\bar{R}(s_i, a) + \epsilon_R(s_i, a) - \hat{R}(s_i,a))$. Also recall that the product of expectation of two independent terms is the expectation of the product. Thus: 
\begin{align}
     &\mathbb{E}_{s_i \sim d_0}\Bigg[\mathbb{E}_{a_i \sim \pi_b(\cdot|s_i)}\Bigg[\frac{\pi_e(a_i|s_i)}{\pi_b^+(a_i|s_i)}^2 \mathbb{E}_{r_i^{a_i} \sim R(s_i, a_i)}\Big[(r_i^a - \hat{R}(s_i,a))^2\Big]\mathbb{V}_{w_{i}^{a} \sim W(s_i, a)}\Big[w_{i}^{a_i}\Big] \\
     &+ \sum_{a \in A \setminus \{a_i\}} \frac{\pi_e(a|s_i)}{\pi_b^+(a|s_i)}^2 \mathbb{E}_{g_i^a \sim G(s_i, a)}\Big[(g_i^a - \hat{R}(s_i,a))^2\Big] \mathbb{V}_{w_{i}^{a} \sim W(s_i, a)}\Big[w_{i}^{a}\Big] \\
     &+ 2 \bar{\psi}_R(s_i, a_i)\sum_{\tilde{a}_j \neq \tilde{a}_i} \bar{\psi}_G(s_i, \tilde{a}_j)  \times Cov_{w_{i}^{a} \sim W(s_i,a)}\Big( w_i^{\tilde{a}_j}, w_i^{a_i}\Big)\\ 
     &+ \sum_{\tilde{a}_j \neq a_i} \sum_{\tilde{a}_k \neq a_i, \tilde{a}_k \neq \tilde{a}_j} \bar{\psi}_G(s_i, \tilde{a}_j)\bar{\psi}_G(s_i, \tilde{a}_k)\\
     &\times  Cov_{w_{i}^{a} \sim W(s_i,a)}\Big( w_i^{\tilde{a}_j}, w_i^{\tilde{a}_k}\Big) \Bigg|a_i\Bigg] \Bigg|s_i\Bigg]
\end{align}
Now we use the definition of variance to simplify the first two expectation squared terms.
\begin{align}
     &\mathbb{E}_{s_i \sim d_0}\Bigg[\mathbb{E}_{a_i \sim \pi_b(\cdot|s_i)}\Bigg[\frac{\pi_e(a_i|s_i)}{\pi_b^+(a_i|s_i)}^2 \mathbb{V}_{w_{i}^{a} \sim W(s_i, a)}\Big[w_{i}^{a_i}\Big] \\
     & \times \Bigg(\mathbb{V}_{r_i^{a_i} \sim R(s_i, a_i)}\Big[(r_i^a - \hat{R}(s_i,a))\Big] + \mathbb{E}_{r_i^a \sim R(s_i, a}\Big[(r_i^a - \hat{R}(s_i,a))\Big]^2\Bigg) \\
     &+ \sum_{a \in A \setminus \{a_i\}} \frac{\pi_e(a|s_i)}{\pi_b^+(a|s_i)}^2 \mathbb{V}_{w_{i}^{a} \sim W(s_i, a)}\Big[w_{i}^{a}\Big] \\
     &\times \Bigg(\mathbb{V}_{g_i^a \sim G(s_i, a)}\Big[(g_i^a - \hat{R}(s_i,a))\Big] + \mathbb{E}_{g_i^a \sim G(s_i, a)}\Big[(g_i^a - \hat{R}(s_i,a))\Big]^2\Bigg) \\
     &+ 2 \bar{\psi}_R(s_i, a_i)\sum_{\tilde{a}_j \neq \tilde{a}_i} \bar{\psi}_G(s_i, \tilde{a}_j)  \times Cov_{w_{i}^{a} \sim W(s_i,a)}\Big( w_i^{\tilde{a}_j}, w_i^{a_i}\Big)\\ 
     &+ \sum_{\tilde{a}_j \neq a_i} \sum_{\tilde{a}_k \neq a_i, \tilde{a}_k \neq \tilde{a}_j} \bar{\psi}_G(s_i, \tilde{a}_j)\bar{\psi}_G(s_i, \tilde{a}_k)\\
     &\times  Cov_{w_{i}^{a} \sim W(s_i,a)}\Big( w_i^{\tilde{a}_j}, w_i^{\tilde{a}_k}\Big) \Bigg|a_i\Bigg] \Bigg|s_i\Bigg]
\end{align}
Simplifying:
\begin{align}
     &\mathbb{E}_{s_i \sim d_0}\Bigg[\mathbb{E}_{a_i \sim \pi_b(\cdot|s_i)}\Bigg[\frac{\pi_e(a_i|s_i)}{\pi_b^+(a_i|s_i)}^2 \mathbb{V}_{w_{i}^{a} \sim W(s_i, a)}\Big[w_{i}^{a_i}\Big] \\
     & \times \Bigg(\sigma_R^2(s_i, a) + (\bar{R}(s_i, a) - \hat{R}(s_i,a))^2\Bigg) \\
     &+ \sum_{a \in A \setminus \{a_i\}} \frac{\pi_e(a|s_i)}{\pi_b^+(a|s_i)}^2 \mathbb{V}_{w_{i}^{a} \sim W(s_i, a)}\Big[w_{i}^{a}\Big] \\
     &\times \Bigg(\sigma_R^2(s_i, a) + \Delta_G(s_i, a) + (\bar{R}(s_i, a) + \epsilon_R(s_i, a) - \hat{R}(s_i,a))^2\Bigg) \\
     &+ 2 \bar{\psi}_R(s_i, a_i)\sum_{\tilde{a}_j \neq \tilde{a}_i} \bar{\psi}_G(s_i, \tilde{a}_j)  \times Cov_{w_{i}^{a} \sim W(s_i,a)}\Big( w_i^{\tilde{a}_j}, w_i^{a_i}\Big)\\ 
     &+ \sum_{\tilde{a}_j \neq a_i} \sum_{\tilde{a}_k \neq a_i, \tilde{a}_k \neq \tilde{a}_j} \bar{\psi}_G(s_i, \tilde{a}_j)\bar{\psi}_G(s_i, \tilde{a}_k)\\
     &\times  Cov_{w_{i}^{a} \sim W(s_i,a)}\Big( w_i^{\tilde{a}_j}, w_i^{\tilde{a}_k}\Big) \Bigg|a_i\Bigg] \Bigg|s_i\Bigg]
\end{align}
Putting together the variance term:
\begin{align}
    &\mathbb{V}_{D \sim \mathcal{D}}[\hat{V}^{\DMCIS}] = \\
    &\frac{1}{N^2}\sum_{i=1}^N\mathbb{V}_{s_i \sim d_0} \Bigg[ v(\pi_e) + \mathbb{E}_{\substack{s_i \sim d_0 \\ a \sim \pi_e(\cdot| s_i)}}\Big[(1- \frac{\bar{W}(a|s_i,a) \pi_b(a|s_i)}{\pi_b^+(a|s_i)}) \epsilon_G(s_i, a)\Big] \Bigg] \\
    &+ \mathbb{E}_{s_i \sim d_0}\Bigg[\mathbb{V}_{a_i \sim \pi_b(\cdot|s_i)}\Big[
    \frac{\pi_e(s_i| a_i)}{\pi_b^+(s_i| a_i)} \bar{W}(a_i|s_i, a_i)(\bar{R}(s_i, a_i) - \hat{R}(s_i, a_i)) \\
    &+ \sum_{a \in A \setminus \{a_i\}}  \frac{\pi_e(a|s_i)}{\pi_b^+(a|s_i)} \bar{W}(a|s_i, a_i) (\bar{R}(s_i, a) + \epsilon_R(s_i, a) - \hat{R}(s_i,a)) \Big| a_i \Big]\Bigg| s_i \Bigg] \\
    &+ \mathbb{E}_{s_i \sim d_0}\Bigg[\mathbb{E}_{a_i \sim \pi_b}\Big[\sum_{a \in A} \sum_{a \in A} \bar{W}(a | s_i, a_i)^2 \frac{\pi_e(a|s_i)}{\pi_b^+(a|s_i)}^2 \sigma_R^2(s_i,a) \\
    &+ \sum_{a \in A \setminus \{a_i\}}\bar{W}(a | s_i, a_i)^2 \frac{\pi_e(a|s_i)}{\pi_b^+(a|s_i)}^2 \Delta_G(s_i, a) \Big| s_i\Big]\Bigg] \\
    &+ \mathbb{E}_{s_i \sim d_0}\Bigg[\mathbb{E}_{a_i \sim \pi_b(\cdot|s_i)}\Big[\frac{\pi_e(a_i|s_i)}{\pi_b^+(a_i|s_i)}^2 \mathbb{V}_{w_{i}^{a} \sim W(s_i, a)}[w_{i}^{a_i}] \\
     & \times \Bigg(\sigma_R^2(s_i, a) + (\bar{R}(s_i, a) - \hat{R}(s_i,a))^2\Bigg) \\
     &+ \sum_{a \in A \setminus \{a_i\}} \frac{\pi_e(a|s_i)}{\pi_b^+(a|s_i)}^2 \mathbb{V}_{w_{i}^{a} \sim W(s_i, a)}[w_{i}^{a}] \\
     &\times \Bigg(\sigma_R^2(s_i, a) + \Delta_G(s_i, a) + (\bar{R}(s_i, a) + \epsilon_R(s_i, a) - \hat{R}(s_i,a))^2\Bigg) \\
     &+ 2 \bar{\psi}_R(s_i, a_i)\sum_{\tilde{a}_j \neq \tilde{a}_i} \bar{\psi}_G(s_i, \tilde{a}_j)  \times Cov_{w_{i}^{a} \sim W(s_i,a)}\Big( w_i^{\tilde{a}_j}, w_i^{a_i}\Big)\\ 
     &+ \sum_{\tilde{a}_j \neq a_i} \sum_{\tilde{a}_k \neq a_i, \tilde{a}_k \neq \tilde{a}_j} \bar{\psi}_G(s_i, \tilde{a}_j)\bar{\psi}_G(s_i, \tilde{a}_k)\\
     &\times  Cov_{w_{i}^{a} \sim W(s_i,a)}\Bigg( w_i^{\tilde{a}_j}, w_i^{\tilde{a}_k}\Bigg) \Bigg| a_i\Bigg] \Bigg| s_i\Bigg]
\end{align}

The variance of $\DMCIS$ under perfect annotations is a slight variant to the expression derived under imperfect annotations. 
\begin{proposition}[Variance of $\DMCIS$ under perfect annotations]
\label{thm:DM+C-IS_variance}
If \Cref{asm:common-support-cf,asm:perfect-annot} hold, 
\begin{align}
    &\mathbb{V}_{D \sim \mathcal{D}}[\hat{V}^{\DMCIS}] = \frac{1}{N^2}\sum_{i=1}^N\mathbb{V}_{s_i \sim d_0}[\hat{V}^{\pi_e}(s_i)] \\
    &+ \mathbb{E}_{s_i \sim d_0}\left[\mathbb{V}_{a_i \sim \pi_b(\cdot|s_i)}\left[\sum_{a \in A}  \frac{\pi_e(a|s)}{\pi_b^+(a|s)} \bar{W}(a|s_i, a_i) (\bar{R}(s_i, a)- \hat{R}(s_i,a)) \middle| s_i\right]\right] \\
    &+ \mathbb{E}_{s_i \sim d_0}\left[\mathbb{E}_{a_i \sim \pi_b}\left[\sum_{a \in A} \bar{W}(a | s_i, a_i)^2 \frac{\pi_e(a|s)}{\pi_b^+(a|s)}^2 \sigma_R(s_i, a)^2 \middle| s_i\right]\right] \\
    &+ \mathbb{E}_{s_i \sim d_0}[\mathbb{E}_{a_i \sim \pi_b(\cdot|s_i)}[\sum_{a \in A} \frac{\pi_e(a|s_i)}{\pi_b^+(a|s_i)}^2 \mathbb{V}_{w_{i}^{a} \sim W(s_i, a_i)}[w_{i}^{a}] \\
    &\times \left(\sigma_R(s_i, a)^2 + (\bar{R}(s_i, a) - \hat{R}(s_i, a)^2\right) \\
    &+ 2 \sum_{\tilde{a}_j, \tilde{a}_k}^{\tilde{a}_j \neq \tilde{a}_k} \frac{\pi_e(\tilde{a}_j|s_i)}{\pi_b^+(\tilde{a}_j|s_i)}  \frac{\pi_e(\tilde{a}_k|s_i)}{\pi_b^+(\tilde{a}_k|s_i)} \\
    &\times Cov_{w_{i}^{a} \sim W(s_i,a)}( w_i^{\tilde{a}_j}, w_i^{\tilde{a}_k})(\bar{R}(s_i, \tilde{a}_j)- \hat{R}(s_i,\tilde{a}_j)) (\bar{R}(s_i, \tilde{a}_k) - \hat{R}(s_i,\tilde{a}_k))|a_i] |s_i]]
\end{align}
\end{proposition}

\begin{proof}
    The expression follows from the prior proof for $\DMCIS$ under imperfect annotations, except with $\Delta_G, \epsilon_G = 0$. 
\end{proof}

\subsection{Expectation and variance of \texorpdfstring{\CDMCIS}{DM⁺-IS⁺}}
The DR estimator is defined as
\begin{align*}
     \hat{V}^{\CDMCIS} &= \frac{1}{N} \sum_{i=1}^N  \left(\hat{R}^+(s_i, \pi_e) + \sum_{a \in A} w_{i}^{a} \frac{\pi_e(a|s_i)}{\pi_{b^+}(a|s_i)}(c_i^a - \hat{R}^+(s_i, a))\right)
\end{align*}
\subsubsection{Expectation}
The expectation without making any assumptions about the quality of the counterfactual annotations is summarized in  \Cref{thm:DR_aug_biased_annot} which is restated below. 
\begin{theorem*}[name=Expectation of $\DMCIS$ and $\CDMCIS$ under imperfect annotations]
Under \Cref{asm:biased_annot,asm:common-support-cf}, the two estimators have the same expectation: 
$$\mathbb{E}[\hat{V}^{\DMCIS}] = \mathbb{E}[\hat{V}^{\CDMCIS}] = v(\pi_e) + \mathbb{E}_{s_i \sim d_0}[\mathbb{E}_{a \sim \pi_e(s_i)}[(1- \frac{\bar{W}(a|s_i,a) \pi_b(a|s_i)}{\pi_b^+(a|s_i)}) \epsilon_G(s_i, a)]]$$
\end{theorem*}
The proof is nearly identical to the one for the $\DMCIS$ estimator, except that the reward function $\hat{R}$ is replaced by $\hat{R}^+$. 
\begin{proposition}[name=Unbiasedness of $\CDMCIS$] \label{thm:C-DM+C-IS_unbiasedness}
If both \Cref{asm:common-support-cf,asm:perfect-annot} hold, the $\CDMCIS$ estimator is unbiased, $\mathbb{E}[\hat{V}^{\CDMCIS}] = v(\pi_e)$. 
\end{proposition}
Proof: The proof is identical to the unbiasedness proof for the $\DMCIS$ estimator. 

\subsubsection{Variance}
We first discuss the variance under imperfect annotations. 
\begin{proposition}[Variance of $\CDMCIS$ under imperfect annotations]
\label{cdmcis_variance_imperfect_annot}
    \begin{align}
    &\mathbb{V}_{D \sim \mathcal{D}}\Big[\hat{V}^{\CDMCIS}\Big] = \\
    &\frac{1}{N^2}\sum_{i=1}^N\mathbb{V}_{s_i \sim d_0} \Bigg[ v(\pi_e) + \mathbb{E}_{\substack{s_i \sim d_0 \\ a \sim \pi_e(\cdot| s_i)}}\Big[(1- \frac{\bar{W}(a|s_i,a) \pi_b(a|s_i)}{\pi_b^+(a|s_i)}) \epsilon_G(s_i, a)\Big] \Bigg] \\
    &+ \mathbb{E}_{s_i \sim d_0}\Bigg[\mathbb{V}_{a_i \sim \pi_b(\cdot|s_i)}\Big[
    \frac{\pi_e(s_i| a_i)}{\pi_b^+(s_i| a_i)} \bar{W}(a_i|s_i, a_i)(\bar{R}(s_i, a_i) - \hat{R}^+(s_i, a_i)) \\
    &+ \sum_{a \in A \setminus \{a_i\}}  \frac{\pi_e(a|s_i)}{\pi_b^+(a|s_i)} \bar{W}(a|s_i, a_i) (\bar{R}(s_i, a) + \epsilon_R(s_i, a) - \hat{R}^+(s_i,a)) \Big| a_i \Big]\Bigg| s_i \Bigg] \\
    &+ \mathbb{E}_{s_i \sim d_0}\Bigg[\mathbb{E}_{a_i \sim \pi_b}\Big[\sum_{a \in A} \sum_{a \in A} \bar{W}(a | s_i, a_i)^2 \frac{\pi_e(a|s_i)}{\pi_b^+(a|s_i)}^2 \sigma_R^2(s_i,a) \\
    &+ \sum_{a \in A \setminus \{a_i\}}\bar{W}(a | s_i, a_i)^2 \frac{\pi_e(a|s_i)}{\pi_b^+(a|s_i)}^2 \Delta_G(s_i, a) \Big| s_i\Big]\Bigg] \\
    &+ \mathbb{E}_{s_i \sim d_0}\Bigg[\mathbb{E}_{a_i \sim \pi_b(\cdot|s_i)}\Big[\frac{\pi_e(a_i|s_i)}{\pi_b^+(a_i|s_i)}^2 \mathbb{V}_{w_{i}^{a} \sim W(s_i, a)}[w_{i}^{a_i}] \\
     & \times \Bigg(\sigma_R^2(s_i, a) + (\bar{R}(s_i, a) - \hat{R}^+(s_i,a))^2\Bigg) \\
     &+ \sum_{a \in A \setminus \{a_i\}} \frac{\pi_e(a|s_i)}{\pi_b^+(a|s_i)}^2 \mathbb{V}_{w_{i}^{a} \sim W(s_i, a)}[w_{i}^{a}] \\
     &\times \Bigg(\sigma_R^2(s_i, a) + \Delta_G(s_i, a) + (\bar{R}(s_i, a) + \epsilon_R(s_i, a) - \hat{R}^+(s_i,a))^2\Bigg) \\
     &+ 2 \bar{\psi}_R(s_i, a_i)\sum_{\tilde{a}_j \neq \tilde{a}_i} \bar{\psi}_G(s_i, \tilde{a}_j)  \times Cov_{w_{i}^{a} \sim W(s_i,a)}\Big( w_i^{\tilde{a}_j}, w_i^{a_i}\Big)\\ 
     &+ \sum_{\tilde{a}_j \neq a_i} \sum_{\tilde{a}_k \neq a_i, \tilde{a}_k \neq \tilde{a}_j} \bar{\psi}_G(s_i, \tilde{a}_j)\bar{\psi}_G(s_i, \tilde{a}_k)\\
     &\times  Cov_{w_{i}^{a} \sim W(s_i,a)}\Bigg( w_i^{\tilde{a}_j}, w_i^{\tilde{a}_k}\Bigg) \Bigg| a_i\Bigg] \Bigg| s_i\Bigg]
\end{align}
\end{proposition}
The proof is identical to that of \Cref{prop:dmcis_imperfect_variance}, except the reward model $\hat{R}$ is replaced with $\hat{R}^+$. 

Now, we derive the variance under the perfect annotation setting.  
\begin{proposition}[Variance of $\CDMCIS$ under perfect annotations]
\label{thm:C-DM-C-IS_variance}
If \Cref{asm:common-support-cf,asm:perfect-annot} holds, 
\begin{align}
    &\mathbb{V}_{D \sim \mathcal{D}}[\hat{V}^{\CDMCIS}] = \frac{1}{N^2}\sum_{i=1}^N\mathbb{V}_{s_i \sim d_0}[\hat{V}^{\pi_e}(s_i)] \\
    &+ \mathbb{E}_{s_i \sim d_0}\left[\mathbb{V}_{a_i \sim \pi_b(\cdot|s_i)}\left[\sum_{a \in A}  \frac{\pi_e(a|s)}{\pi_b^+(a|s)} \bar{W}(a|s_i, a_i) (\bar{R}(s_i, a)- \hat{R}^+(s_i,a)) \middle| s_i\right]\right] \\
    &+ \mathbb{E}_{s_i \sim d_0}\left[\mathbb{E}_{a_i \sim \pi_b}\left[\sum_{a \in A} \bar{W}(a | s_i, a_i)^2 \frac{\pi_e(a|s)}{\pi_b^+(a|s)}^2 \sigma_R^2(s_i,a) \middle| s_i\right]\right] \\
    &+ \mathbb{E}_{s_i \sim d_0}[\mathbb{E}_{a_i \sim \pi_b(\cdot|s_i)}[\sum_{a \in A} \frac{\pi_e(a|s_i)}{\pi_b^+(a|s_i)}^2 \mathbb{V}_{w_{i}^{a} \sim W(s_i, a_i)}[w_{i}^{a}] \\
    &\times \left(\sigma_R^2(s_i,a) + (\bar{R}(s_i, a) - \hat{R}^+(s_i, a)^2\right) \\
    &+ 2 \sum_{\tilde{a}_j, \tilde{a}_k}^{\tilde{a}_j \neq \tilde{a}_k} \frac{\pi_e(\tilde{a}_j|s_i)}{\pi_b^+(\tilde{a}_j|s_i)}  \frac{\pi_e(\tilde{a}_k|s_i)}{\pi_b^+(\tilde{a}_k|s_i)} \\
    &\times Cov_{w_{i}^{a} \sim W(s_i,a)}( w_i^{\tilde{a}_j}, w_i^{\tilde{a}_k})(\bar{R}(s_i, \tilde{a}_j)- \hat{R}^+(s_i,\tilde{a}_j)) (\bar{R}(s_i, \tilde{a}_k) - \hat{R}^+(s_i,\tilde{a}_k))|a_i] |s_i]] 
\end{align}
\end{proposition}
The proof is identical to the one used to derive variance under perfect conditions for the $\DMCIS$ estimator, except with $\hat{R}^+$ replacing $\hat{R}$. 
\section{Equivalence of \texorpdfstring{\CIS}{IS⁺}, \texorpdfstring{\DMCIS}{DM-IS⁺}, \texorpdfstring{\CDMCIS}{DM⁺-IS⁺} under equal weights}
\label{apd:equivalence}
\begin{corollary}[name=Equivalence of $\CIS\text{, } \CDMIS \text{, }\CDMCIS$ under equal weights] \label{corr:equiv_equal_weights}
If we assume that all the weights are equal, and that all OPE methods have access to the same set of counterfactual annotations, $\hat{V}^{\CIS}$, $\hat{V}^{\DMCIS}$, $\hat{V}^{\CDMCIS}$ are equivalent by definition. 
\end{corollary}
Proof: 
First we re-state the definition of the three methods for a single sample $s_i, a_i, r_i$, and all counterfactual annotations $c_i^a$, under the assumption that all weights are equal. Under this assumption, each weight is $\frac{1}{|A|}$:
\begin{enumerate}
    \item 
    \begin{align}
        \hat{V}^{\CIS} &= \sum_{a \in \mathcal{A}} w_i^a \frac{\pi_e(a|s_i)}{\pi_{b^+}(a|s_i)} c_i^a  \\
        \hat{V}^{\CIS*} &= \sum_{a \in A} \pi_e(a|s_i) c_i^a \\
        \hat{V}^{\DMCIS} &= \hat{R}(s_i, \pi_e) + \sum_{a \in A} \pi_e(a|s_i) (c_i^a - \hat{R}(s_i, a)) \\
        \hat{V}^{\CDMCIS*} &= \hat{R}^+(s_i, \pi_e) + \sum_{a \in A} \pi_e(a|s_i) (c_i^a - \hat{R}^+(s_i, a))
    \end{align}
    $\hat{V}^{\CIS*}$ is $\hat{V}^{\CIS}$ under equal weights. The following is a derivation of how we can reach $\hat{V}^{\CIS*}$ under equal weights using the definition of the augmented behavior policy:
    \begin{align}
        \hat{V}^{\CIS} &= \sum_{a \in \mathcal{A}} w_i^a\frac{\pi_e(a|s_i)}{\pi_{b^+}(a|s_i)} c_i^a \\
        &= \sum_{a \in \mathcal{A}} \bar{W}(a|s_i, a_i) \frac{\pi_e(a|s_i)}{\pi_{b^+}(a|s_i)} c_i^a \\
        &= \sum_{a \in \mathcal{A}} \pi_e(a|s_i)c_i^a \\
        &= \hat{V}^{C*-IS}
    \end{align}
    \item Now, we write the definition for $\hat{R}(s_i, \pi_e)$ and $\hat{R}^+(s_i, \pi_e)$. 
    \begin{align}
        \hat{R}(s_i, \pi_e) &= \sum_{a \in A} \pi_e(a|s_i) \hat{R}(s_i, a) \\
        \hat{R}^+(s_i, \pi_e) &= \sum_{a \in A} \pi_e(a|s_i) \hat{R}^+(s_i, a)
    \end{align}
    \item Let us start by re-writing the definition of $\hat{V}^{\DMCIS}$ using the definition of $\hat{R}(s_i, \pi_e)$:
    \begin{align}
        \hat{V}^{\DMCIS} &= \sum_{a \in A} \pi_e(a|s_i) \hat{R}(s_i, a) + \sum_{a \in A} \pi_e(a|s_i) (c_i^a - \hat{R}(s_i, a)) \\ 
        &= \sum_{a \in A} \pi_e(a|s_i) \hat{R}(s_i, a) + \sum_{a \in A} \pi_e(a|s_i) c_i^a  - \sum_{a \in A}\pi_e(a|s_i)\hat{R}(s_i, a)\\ 
    \end{align}
    \item Note that the second term is the definition of $AugIS$ for a single sample. Also note that subtracting term 3 from term 1 is equal to 0 by definition. 

\end{enumerate}
This shows that $\CIS$ is equivalent to $\DMCIS$. Note that we can replace the estimate of the reward function with $\hat{R}^+$ and produce the same derivation for $\CDMCIS$. Thus, we have shown that the three estimators $\CIS$, $\DMCIS$ $\CDMCIS$ are equivalent when the weights are equal and all three methods have access to the same counterfactual annotations. 
\begin{corollary}[name=Equivalence of Variance of \texorpdfstring{\CIS}{IS⁺}, \texorpdfstring{\DMCIS}{DM-IS⁺}, \texorpdfstring{\CDMCIS}{DM⁺-IS⁺} under equal weights] \label{corr:equiv_variance_equal_weights}
\end{corollary}
Now, we want to verify that our variance decomposition for $\hat{V}^{\CIS}, \hat{V}^{\DMCIS*}, \hat{V}^{\CDMCIS*}$ are equivalent to that of $\CIS*$. Recall that we are in the setting where the weights are equal. Here, we also assume that $\hat{R}, \hat{R}^+$ are constant and have been learned already. 
\begin{enumerate}
    \item First we note that the variance decomposition for $\DMCIS$ and $\CDMCIS$ each have 7 terms. The last 4 terms reason about the covariance of the weights. Because our weights are constant and equal, the last four terms become 0. 
    \item Next, we note that the first term is identical in all of the variance decompositions. Namely, this is $\mathbb{V}_{s_i \sim d_0}[V^{\pi_e}(s_i)]$. 
    \item Now, we note that if $\hat{R}, \hat{R}^+$ are constants, the second term in the variance of $\DMCIS$ and $\CDMCIS$ becomes 0. 
    \item All we need to do is establish that that the third term is equivalent to the third term for the variance decomposition. This is true by definition. 
\end{enumerate}
Thus, all three variance decompositions are the same under this setting. 

\end{document}